\documentclass{article}

\usepackage[main, final]{neurips_2026}

\usepackage[utf8]{inputenc}
\usepackage[T1]{fontenc}

\usepackage{microtype}
\usepackage{graphicx}
\usepackage{subcaption}
\usepackage{booktabs}
\usepackage{url}
\usepackage{xurl}
\usepackage{xcolor}
\usepackage{nicefrac}
\usepackage{wrapfig}

\usepackage{amsmath}
\usepackage{amssymb}
\usepackage{amsfonts}
\usepackage{mathtools}
\usepackage{amsthm}
\DeclareMathOperator*{\softmax}{softmax}
\usepackage{xcolor}
\usepackage{makecell}
\usepackage{multirow}
\usepackage{rotating}
\usepackage{enumitem}
\usepackage{xspace}
\usepackage{algorithm}
\usepackage{algorithmic}
\usepackage{relsize}

\usepackage[breaklinks=true]{hyperref}
\usepackage[capitalize,noabbrev]{cleveref}

\newcommand{\super}{SuperActivator}
\newcommand{\supers}{SuperActivators}

\newcommand{\tabletextsize}{\small}
\newcommand{\tablenumbersize}{\small}

\theoremstyle{plain}
\newtheorem{theorem}{Theorem}[section]

\newtheorem{lemma}[theorem]{Lemma}
\newtheorem{corollary}[theorem]{Corollary}
\theoremstyle{definition}

\theoremstyle{remark}

\usepackage{amsmath,amssymb,amsthm}
\usepackage{geometry}
\usepackage{hyperref}

\newcommand{\Var}{\mathrm{Var}}
\newcommand{\E}{\mathbb{E}}

\usepackage[textsize=tiny]{todonotes}

\title{The SuperActivator Mechanism: Transformers Concentrate Reliable Concept Signals in the Tail}

\author{%
  Cassandra Goldberg \\
  University of Pennsylvania \\
  \texttt{cgoldber@seas.upenn.edu} \\
  \And
  Chaehyeon Kim \\
  University of Pennsylvania \\
  \texttt{chaenyk@engineering.upenn.edu}
  \And
  Adam Stein\thanks{Work conducted while at the University of Pennsylvania}\\
Reflection AI\\
\texttt{adam.stein@reflection.ai}
  \And
  Eric Wong \\
  University of Pennsylvania \\
  \texttt{exwong@engineering.upenn.edu}
}

\begin{document}

\maketitle
\setcounter{footnote}{0}

% === Main sections ===
\begin{abstract}
Concept vectors aim to enhance model interpretability by linking learned internal representations with human-understandable semantics, but their practical utility is often limited by noisy and inconsistent activations.
In this work, we discover the \textbf{\super{} Mechanism}: a transformer dynamic that amplifies concept activation gaps, concentrating the most reliable concept evidence into a small set of highly activated tokens, which we call \supers{}. We theoretically characterize this mechanism, proving that concept-aligned attention produces two competing effects: a \textit{tail amplification effect}, where gaps among already-high activations grow disproportionately fast, and an \textit{equalization effect} that eventually dominates as attention concentrates on extreme tokens. We validate the predicted tail amplification empirically, finding that in-concept activation distributions develop well-separated upper tails consistently across models, image and text modalities, and concept types.
Moreover, we demonstrate the practical implications of \supers{} for concept detection and localization. First, using only high-activation tokens as concept indicators improves detection by up to 14 F1 points over competing baselines. Second, restricting detection to the extreme activation tail is not merely sufficient but essential: including activations beyond the top 5–10\% of the in-concept distribution introduces noise that degrades performance. Finally, using \supers{} as instance-specific attribution targets yields more accurate and faithful concept localization heatmaps than using global concept vectors. \footnote {Code available at \url{https://github.com/BrachioLab/SuperActivators}}
\end{abstract}
\section{Introduction}
\label{sec:introduction}

% 1. High order problem the world is currently facing !!

% 2. What is the area that we are working in that is going to solve this problem

% 3. What are the challenges

% 4. How do we address this challenges

% Summarizes the main contribution (Bulletpoints)

Modern transformer-based models, while increasingly powerful and ubiquitous~\citep{Minaee2024LargeLM}, remain opaque and can behave in ways that are unpredictable or harmful~\citep{Liu2025TheSO,Greenblatt2024AlignmentFI}. This opacity hinders our ability to identify and debug undesirable model behaviors, such as relying on spurious correlations~\citep{Zhou2023ExploreSC}, biases~\citep{Yang2024UnmaskingAQ}, or fragile reasoning~\citep{Berglund2023TheRC}, or to intervene when models produce problematic outputs.

Concept vectors~\citep{Kim2017InterpretabilityBF}, or semantically meaningful directions in a model’s latent space, provide a lightweight tool for examining and influencing internal representations. They have been used to uncover hidden model failures~\citep{Abid2021MeaningfullyDM, Pfau2021RobustSI}, and to steer model behavior away from hallucinations~\citep{Rimsky2023SteeringL2, Suresh2025FromNT}, unsafe responses~\citep{assogba2026sparse, Ghosh2025SafeSteerIS}, and toxic language~\citep{Turner2023SteeringLM, Nejadgholi2022TowardsPF}. Moreover, unsupervised concept extraction can reveal previously unknown knowledge embedded within model representations~\citep{Donhauser2024TowardsSD, ZednikBoelsen2022ScientificXAI} without requiring costly labels. 

To analyze the presence of concepts within a sample (i.e., an image or text instance), we typically rely on \textit{concept activation scores}, which measure alignment between an input token's embedding and a concept vector. However, these scores are often inconsistent and noisy, and as a result misrepresent true concept presence. Prior work has shown that concept vectors frequently activate on unintended semantics~\citep{Olah2020ZoomIA, bricken2023monosemanticity}, produce overlapping signals for correlated concepts~\citep{goh2021multimodal, Olah2020ZoomIA}, and exhibit unstable activation patterns across different model layers~\citep{Nicolson2024ExplainingER}. 

Figure~\ref{fig:superpatch-example} illustrates this ambiguity using \emph{Llama-3.2-11B-Vision-Instruct} linear separator activations on a \emph{COCO} image of a dog reflected in a car mirror. Both the \textit{Animal} and \textit{Person} activation heatmaps highlight the same region, despite only the former being present. The fairly high \textit{Person} activations could easily mislead someone into thinking the image contains a person. Conversely, even when a concept is present, not all corresponding tokens activate strongly. For example, many tokens in the \textit{Car} region have activations indistinguishable from the background. Yet, amid this noise, we identify a small subset of tokens that reliably indicate concept presence. These extremely high activations, which we call \textit{\supers{}} (green boxes), appear only in the heatmaps for present concepts.

To understand why these \supers{} emerge, we analyze how concept activations evolve across transformer layers, both theoretically and empirically. We prove that concept-aligned attention produces two opposing effects: a \textit{tail amplification effect}, where gaps among high activations grow disproportionately fast, and an eventual \textit{equalization effect}, which erases relative differences between token activations. The \textbf{\super{} Mechanism} is the early regime where tail amplification overcomes equalization. Since equalization develops gradually across layers and is further delayed by noise, \supers{} emerge before activations can equalize in real, finite-depth models. We then validate this theory empirically, observing that in-concept activation distributions develop well-separated upper tails across models, modalities, datasets, and concept types. Crucially, the \supers{} in these tails reliably indicate concept presence despite broader activation noise.

Our key contributions are summarized as follows:

\begin{figure}[!t]
  \centering
    \includegraphics[width=\textwidth]{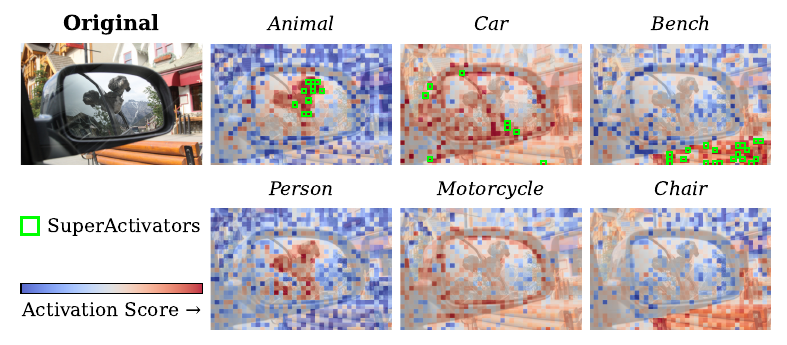}
    \vspace{-0.5cm}
    \caption{\textbf{The \super{} Mechanism concentrates the most informative concept signals into a sparse set of in-concept activations.} Heatmaps show token-level concept activation scores, with red indicating high concept alignment and blue indicating low alignment. \supers{} are extremely high-activation tokens, marked by green boxes, that reliably distinguish true concept occurrences even when concept activation heatmaps appear to highlight absent concepts or fail to fully capture present ones. Additional image and text examples are provided in Appendix~\ref{app:superactivator-examples}.}
    \vspace{-1em}
  \label{fig:superpatch-example}
\end{figure}

\begin{itemize}

\item \textbf{Theoretically Grounded \super{} Mechanism:} We prove that concept-aligned attention produces a \textit{tail amplification effect}, in which gaps among already-high activations grow disproportionately fast, and an \textit{equalization effect} that eventually dominates, washing out relative activation differences.

\item \textbf{High-Tail Activations Indicate Concept Presence:}
We validate our theoretical results and show that using only the most extreme activations for concept detection improves F1 by up to 14 points above baselines, while including activations beyond the top 5--10\% of the in-concept distribution degrades performance.

\item \textbf{Generalization and Robustness:} We demonstrate consistent results and robustness across modalities, models, layers, concept types, and fixed tail-sparsity settings.

\item \textbf{\supers{} Improve Concept Localization:} 
Compared with global concept vector targets, \super{}-derived targets yield attribution maps that better align with ground-truth annotations and achieve better faithfulness scores.

\end{itemize}

\begin{figure*}[!t]
    \centering
    % Left subplot
    \begin{subfigure}[t]{0.59\textwidth}
      \centering
      \includegraphics[width=\textwidth]{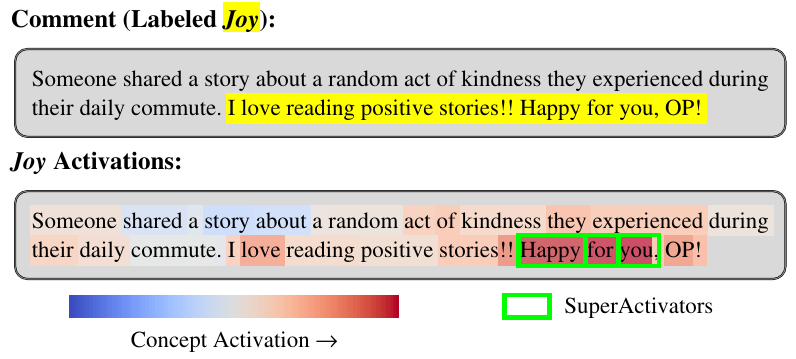}
    \end{subfigure}
    \hfill
    \begin{subfigure}[t]{0.4\textwidth}
      \centering
      \includegraphics[width=\textwidth]{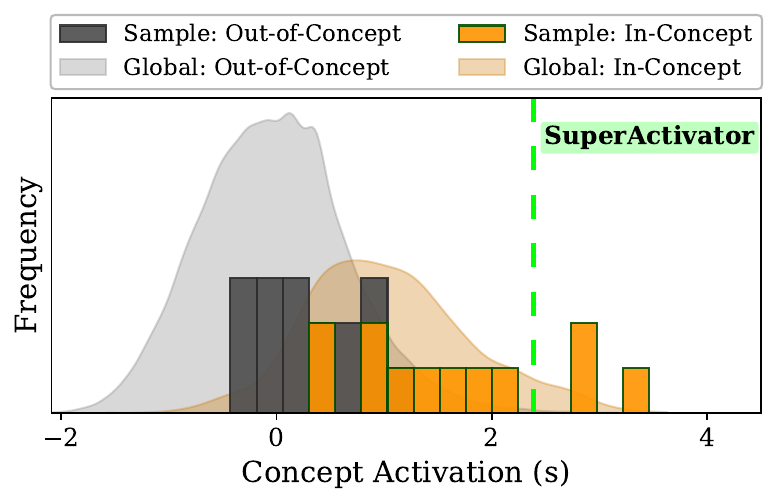}
    \end{subfigure}
    \caption{Concept activations are noisy and non-uniform, making it difficult to distinguish in-concept tokens from out-of-concept ones. In this example, the ground-truth span for \textit{Joy} is highlighted, with token-level activations shown both as a heatmap over the text (left) and as distributions (right). While a few in-concept tokens exhibit extremely high activations, many remain indistinguishable from out-of-concept token activations within the sample and across $D_c^{\text{out}}$.}
    \label{fig:noisy-acts}
\end{figure*}

\section{Preliminaries}
\label{sec:concept_background}
This section defines basic notation for representing inputs, concept vectors, and activation scores; more detailed formal definitions are provided in Appendix~\ref{app:detection-formalisms}.

Let $f$ be a pretrained transformer model that processes an input sample $x \in \mathcal{X}$ (an image or a text sequence) through its layers. From any given layer of $f$, we can extract token-level embeddings $(z_1^{\text{tok}}(x), \dots, z_{n(x)}^{\text{tok}}(x)) \in (\mathbb{R}^d)^{n(x)}$ and a sample-level embedding $z^{\text{cls}}(x) \in \mathbb{R}^d$. The  number of tokens, $n(x)$, is sample size dependent. For any semantic concept $c$, we associate a concept vector $v_c \in \mathbb{R}^d$, which represents a direction in the embedding space. The concept activation score of an embedding $z$ with respect to concept $c$ is defined as the dot product of the embedding with the concept vector, $s_c(z) = \langle z, v_c \rangle$, where positive scores indicate alignment with the concept.

We characterize, for each concept $c$, the distribution of activation scores across many samples. Let $\mathcal{D}_c^{\text{in}}$ and $\mathcal{D}_c^{\text{out}}$ denote the population-level distributions of activation scores for \textit{in-concept tokens} (those labeled concept-positive for $c$) and \textit{out-of-concept tokens} (those labeled concept-negative for $c$), respectively. We estimate them using finite datasets $D_c^{\text{in}}$ and $D_c^{\text{out}}$ constructed from observed activations. 
Let $Z_c^{\text{in}}, Z_c^{\text{out}} \subseteq Z$ denote the in-concept and out-of-concept tokens, with $Z_c^{\text{out}}$ restricted to samples not containing $c$ to prevent self-attention leakage. Then
\[
D_c^{\text{in}} = \{\, s_c(z) : z \in Z_c^{\text{in}} \,\}, 
\;
D_c^{\text{out}} = \{\, s_c(z) : z \in Z_c^{\text{out}} \,\},
\]
which serve as empirical samples from $\mathcal{D}_c^{\text{in}}$ and $\mathcal{D}_c^{\text{out}}$.

\begin{figure*}[!t]
    \centering
    \includegraphics[width=\linewidth]{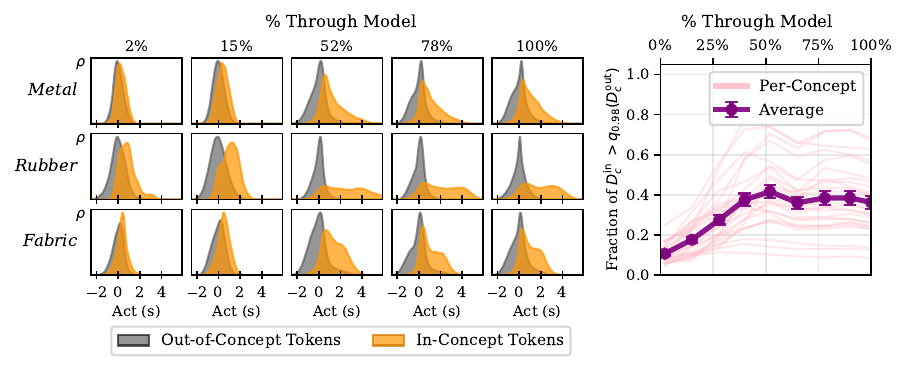}
    
    \caption{The in-concept ($D_c^{\text{in}}$) and out-of-concept ($D_c^{\text{out}}$) distributions become more distinct with model depth, though the separation is concentrated in a small subset of tokens in the high-activation tail of $D_c^{\text{in}}$. Shown are activation distributions across model layers for three \emph{OpenSurfaces} concepts (left), along with the proportion of $D_c^{\text{in}}$ activations exceeding $q_{0.98}(D_c^{\text{out}})$ at each layer (right).}

    \label{fig:acts-across-layers}
\end{figure*}
% \section{The \super{} Mechanism Yields Clear Concept Signals Amid Noisy Activations}
\section{The \super{} Mechanism Amplifies High-Activation Tails That Cut Through the Noise}

% \subsection{Concept Activations are Inconsistent and Poorly Separated}
\subsection{Concept Activations are Noisy and Poorly Separated}
Concept vectors promise interpretability, but they often deliver noisy activations that are difficult to extract meaningful insights from. It is well-documented that concept representations can encode spurious correlations and blur important context-specific distinctions~\citep{Abid2021MeaningfullyDM, Zhou2021FrequencybasedDI}. Other studies highlight issues of entanglement, where related features co-activate, and polysemanticity, where a single vector represents multiple unrelated concepts~\citep{goh2021multimodal, Olah2020ZoomIA, bricken2023monosemanticity}.

To study these limitations empirically, we analyze the separability of concept activations. In doing so, we identify a key challenge: many tokens labeled as in-concept exhibit activation scores that are not meaningfully different from out-of-concept tokens. Figure~\ref{fig:noisy-acts} illustrates this problem for the \textit{Joy} concept using linear separator activations from \emph{Llama-3.2-11B-Vision-Instruct} on a sample from the \emph{Augmented GoEmotions} test set. While a subset of in-concept tokens exhibit strong \textit{Joy} activations, a substantial portion fall well within the range of out-of-concept activations, both within the given text sample and across the broader dataset. Thus, no single activation threshold can partition true \textit{Joy} tokens from non-\textit{Joy} ones.

We analyze this behavior more broadly by characterizing the empirical activation distributions $D_c^{\text{in}}$ and $D_c^{\text{out}}$, and examining how their separability evolves across model depth. Activations extracted from the \emph{OpenSurfaces} dataset are visualized in Figure~\ref{fig:acts-across-layers}, and additional examples from other datasets are given in Appendix~\ref{app:distributions}. At each layer, $D_c^{\text{out}}$ appears roughly normal and centered at zero. In early layers, $D_c^{\text{in}}$ overlaps considerably with $D_c^{\text{out}}$. 
As depth increases, the proportion of $D_c^{\text{in}}$ exceeding the $98th$ percentile of $D_c^{\text{out}}$, $q_{0.98}(D_c^{\text{out}})$, grows steadily before plateauing in middle layers. This trend aligns with prior findings that concept representations become more separable at intermediate depths and can collapse again in the final layers due to task-specific compression~\citep{Salam2025LargeLM, Yu2024LatentCE, Dalvi2022DiscoveringLC}.
However, even at its peak, the average proportion is only around $0.4$, indicating that most in-concept activations remain largely indistinguishable from $D_c^{\text{out}}$.

\subsection{Introducing the \super{} Mechanism}
\label{sec:introducing_supers}
\begin{wrapfigure}{r}{0.43\textwidth}
  \vspace{-3em}
  \centering
  \setlength{\columnsep}{40pt}
  \includegraphics[width=\linewidth]{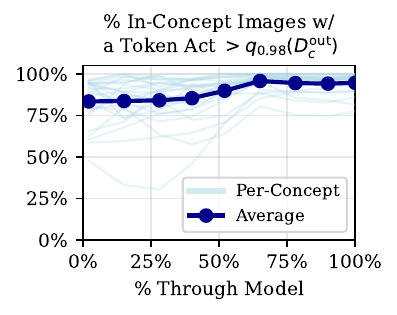}
  \vspace{-2em}
  \caption{Most in-concept \emph{OpenSurfaces} images have at least one activation in the high-activation tail of $D_c^{\text{in}}$, well-separated from $q_{0.98}(D_c^{\text{out}})$.}
  \vspace{-2em}
  \label{fig:coverage-example}
\end{wrapfigure}
The growing separation between $D_c^{\text{in}}$ and $D_c^{\text{out}}$ with model depth does not result from a uniform shift of all in-concept activations. 
Instead, $D_c^{\text{in}}$ develops a heavy positive tail as a 
small subset of extreme activations become increasingly separable with depth, standing out clearly from 
out-of-concept noise.

However, a \textit{reliable} concept signal
must be not only \textit{clear}, but also \textit{consistent}, with broad coverage across in-concept samples. Notably, as shown in Figure~\ref{fig:coverage-example} for \emph{Llama-3.2-11B-Vision-Instruct} linear separator concepts on \emph{OpenSurfaces}, we find that the high-activation tail of $D_c^{\text{in}}$ exhibits good coverage: most in-concept images contain at least one activation above the 98th percentile of $D_c^{\text{out}}$.

We define the \textbf{\super{} Mechanism} as the transformer dynamic that non-uniformly separates concept activations across model depth: while much of $D_c^{\text{in}}$ remains overlapping with $D_c^{\text{out}}$, it develops a positive tail that contains a small set of highly reliable concept signals.

Empirically, we observe this phenomenon consistently across image and text modalities, model architectures, and concept vector types (see Appendix \ref{app:super-motivation}). Next, we theoretically characterize the attention dynamics underlying this mechanism.

% (See Appendix ~\ref{app:optimal-sparsity-across-layers} and ~\ref{app:sparsity-ablation}). 

% We find that \super{} detection is most effective at very low $\delta$, showing that the most reliable concept information is concentrated in a small high-activation tail of $D_c^{\text{in}}$. For final evaluation, we calibrate $\delta$ per-concept on the validation set to maximize detection $F_1$.  

\subsection{Theoretical Analysis Underlying the SuperActivator Mechanism}
\label{sec:theory}

\begin{figure*}[!t]
    \centering
    \includegraphics[width=.91\linewidth]{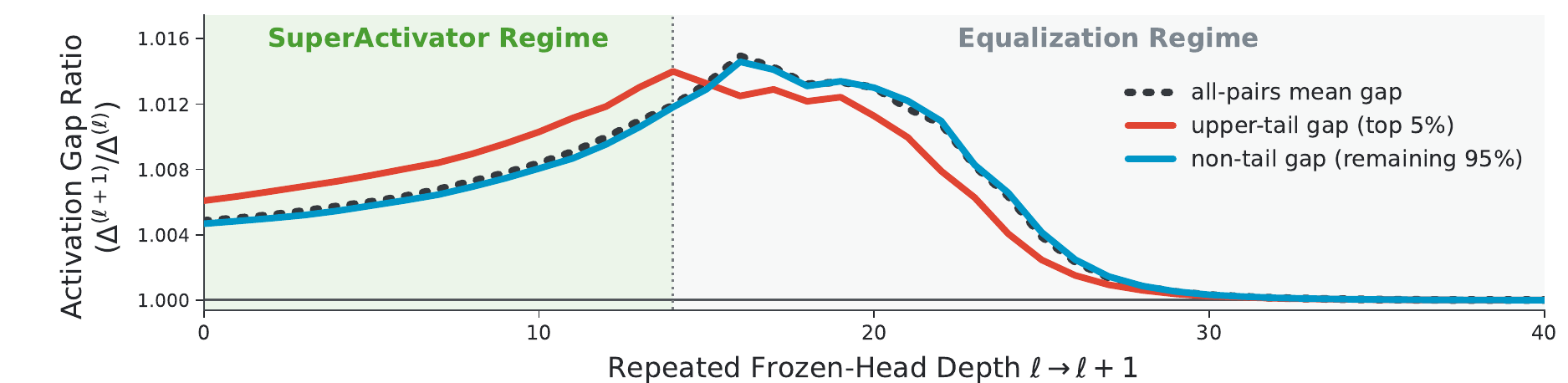}
    
    \caption{The \super{} Mechanism is the tail amplification regime that occurs before equalization takes over. When a concept-aligned head is applied repeatedly, upper-tail gaps (top $5\%$, red) initially amplify faster than non-tail gaps (blue), before the activations equalize and their gap ratios approach 1. The setup that produced this plot is detailed in Appendix~\ref{sec:constructed-heads}.}
    % \vspace{-0.em}
    \label{fig:toy-dynamics}
\end{figure*}

Here, we derive a theoretical account of the transformer dynamics that give rise to the \super{} Mechanism. The mechanism is governed by two competing effects. First, the \textit{tail amplification effect}: concept-aligned attention disproportionately amplifies gaps among already-high concept activations, causing the upper tail to separate from the rest of the distribution (Theorems~\ref{thm:gap-growth} and \ref{thm:main-skew}). However, as these activations grow, attention increasingly concentrates on the highest-activation token, producing the competing \textit{equalization effect}, in which relative differences between token activations become negligible (Corollaries \ref{cor:attention-concentration} and \ref{cor:within-tail-equalization}). The \super{} Mechanism is the initial regime where tail amplification dominates, before equalization eventually overtakes it at sufficient depth. Figure~\ref{fig:toy-dynamics} illustrates this transition under repeated application of a fixed concept-aligned attention head.

\paragraph{Setup.} 

Building on the read-write decomposition of attention heads \citep{elhage2021mathematical} and prior work identifying semantically specialized heads \citep{Gandelsman2023InterpretingCI}, we formalize a concept-aligned head as one that reads a concept direction $v$ through its QK pathway and writes back along $v$ through its OV pathway. For such a head, with query, key, value, and output matrices $W_Q,W_K,W_V,W_O$, head dimension $d_k$, and OV circuit $W_{OV} := W_O W_V$, we assume:
\vspace{-1.5em}

\begin{align*}
    \text{(A1)}\quad
    \tfrac{1}{\sqrt{d_k}}\, W_Q^\top W_K &= \gamma\, vv^\top + \varepsilon_{QK},
    \qquad
    \text{(A2)}\quad
    W_{OV} = \lambda\, vv^\top + \varepsilon_{OV},
\end{align*}
where $\gamma,\lambda>0$ measure the strength of the concept-aligned QK and OV components, and $\varepsilon_{QK}$ and $\varepsilon_{OV}$ capture the remaining non-$v$-aligned structure in the head, with $v^\top \varepsilon_{QK}v=v^\top \varepsilon_{OV}v=0$. 

We verify in Appendix~\ref{app:head-validation} that heads with positive alignment along $v$ in both the QK and OV pathways exist for every concept in every model we study. We further validate their functional role by showing that ablating these heads reduces tail-bulk activation gaps more than ablating randomly selected heads (Appendix~\ref{app:aligned-head-ablation}).

In this section, we analyze the dynamics induced by such a head and, for brevity, present the noiseless case with $\varepsilon_{QK}=\varepsilon_{OV}=0$. Appendix~\ref{sec:noisy} extends the analysis to a bounded-noise setting, showing that these dynamics persist when the concept signal is sufficiently strong relative to the noise. For input token representation $x_i^{(l)}$ at layer $l$, projecting the standard transformer residual attention update from this head onto $v$ gives the concept activation update:
\begin{equation}
    s_i^{(l+1)} = s_i^{(l)} + \lambda \sum_{j=1}^{N} \alpha_{ij}^{(l)}\, s_j^{(l)},
    \qquad
    \alpha_{ij}^{(l)} = \softmax_j\!\left(\gamma\, s_i^{(l)} s_j^{(l)}\right).
    \label{eq:scalar-recursion}
\end{equation}
In the noisy setting, we also assume a non-concept residual $\eta_i^{(l)}$ in the full layer update (A3), capturing contributions from other heads, MLP updates, and residual-stream effects.

\begin{theorem}[Activation gap amplification]
\label{thm:gap-growth}
For any pair of tokens $a,b$ with $s_a^{(l)} \neq s_b^{(l)}$, the corresponding activation gap
$\Delta_{ab}^{(l)} := s_a^{(l)} - s_b^{(l)}$ evolves as:
\begin{equation*}
    \Delta_{ab}^{(l+1)}
    =
    \bigl(1+\lambda\gamma \overline{V}_{ab}^{(l)}\bigr)
    \Delta_{ab}^{(l)},
    \qquad
    \overline{V}_{ab}^{(l)}
    :=
    \int_0^1
    \operatorname{Var}_{\alpha^{(l)}(\gamma(s_b^{(l)}+u\Delta_{ab}^{(l)}))}
    \!\left[s^{(l)}\right]\,du .
\end{equation*}
\vspace{-1em}

Here, $\overline{V}_{ab}^{(l)}$ is the attention-weighted variance of the layer-$l$ activations, averaged as the query activation moves from $s_b^{(l)}$ to $s_a^{(l)}$. Since $s_a^{(l)} \neq s_b^{(l)}$ and softmax gives every token positive weight, $\overline{V}_{ab}^{(l)}>0$, so the multiplicative factor is strictly larger than one.

\end{theorem}

This theorem gives the basic amplification engine: existing concept activation differences expand multiplicatively from one layer to the next, with larger attention-weighted variance producing faster gap growth. Two corollaries follow that describe the asymptotic consequences of this amplification:

\begin{corollary}[Attention concentration]
\label{cor:attention-concentration}
As the margins between the maximal and minimal activations and the rest of the sequence grow, attention eventually concentrates on these two extreme tokens (or poles).
\end{corollary}

\begin{corollary}[Within-tail concept activation equalization]
\label{cor:within-tail-equalization}
Once attention has concentrated on the poles, same-tail activations receive nearly the same update and therefore equalize in relative scale.
\end{corollary}

Together, these corollaries show that continued gap amplification eventually gives way to within-tail equalization as attention concentrates on the extreme tokens.

\begin{theorem}[Tail-asymmetric amplification]
\label{thm:main-skew}
Under the noiseless scalar dynamics in Eq.~\eqref{eq:scalar-recursion}, define the one-layer update map for a token with activation $s$ as
\[
    T^{(l)}(s)
    :=
    s
    +
    \lambda
    \mathbb{E}_{\alpha^{(l)}(\gamma s)}[s^{(l)}],
\]
where $\alpha^{(l)}(\gamma s)$ is the softmax attention distribution over layer-$l$ activations induced by query scale $\gamma s$. Then the curvature of this update map satisfies
\begin{equation*}
    \bigl(T^{(l)}\bigr)''(s)
    =
    \lambda \gamma^{2}
    \cdot
    \mathbb{E}_{\alpha^{(l)}(\gamma s)}
    \!\Bigl[
    \bigl(
    s^{(l)}
    -
    \mathbb{E}_{\alpha^{(l)}(\gamma s)}[s^{(l)}]
    \bigr)^{3}
    \Bigr].
\end{equation*}
Thus, when the attention-weighted third central moment (skew) is positive, the local gap-amplification rate increases with $s$, so upper-tail gaps amplify faster than bulk gaps.
\end{theorem}

This asymmetric amplification drives the \super{} Mechanism: before equalization dominates, upper-tail gaps grow faster than non-tail ones, causing the strongest concept activations to separate from the rest of the distribution. Reliable concept signals are thus concentrated in this extreme tail rather than broadly distributed across in-concept activations.

% \subsection{\supers{} Identify Where Concept-Presence Signals Are Most Reliable}
% Requires \usepackage{booktabs} and \usepackage{natbib} (for \citep).
\begin{table*}[t]
\centering
\small
\setlength{\tabcolsep}{6pt}
\caption{\textbf{As predicted by the \super{} Mechanism, \super{}-based detection outperforms standard aggregation and prompting baselines.} Results are shown for \emph{Llama-3.2-11B-Vision-Instruct} linear separators, comparing against RandTok, LastTok, MeanTok, CLSTok, and Prompting. Appendix~\ref{app:detection-results} reports results across models and concept types. Reported uncertainties are 95\% bootstrap confidence intervals. \textbf{Bold} is best on average; \underline{underline} is second best.}
\vspace{-0.5em}
\label{tab:detection_f1_summary}
\begin{tabular}{lcccccc}
    \toprule
    & \multicolumn{6}{c}{Concept Detection Methods} \\
    \cmidrule(lr){2-7}
    & RandTok & LastTok {\scriptsize\cite{Chen2025PersonaVM}} & MeanTok {\scriptsize\citep{McKenzie2025DetectingHI}} & CLSTok {\scriptsize\citep{Yu2024LatentCE}} & Prompt {\scriptsize\citep{Wu2025AxBenchSL}} & SuperAct \\
    \midrule
    COCO         & 0.41 {\scriptsize$\pm 0.01$} & 0.36 {\scriptsize$\pm 0.01$} & 0.67 {\scriptsize$\pm 0.01$} & 0.57 {\scriptsize$\pm 0.01$} & \underline{0.69} {\scriptsize$\pm 0.01$} & \textbf{0.83} {\scriptsize$\pm 0.01$} \\
    OpenSurfaces & 0.38 {\scriptsize$\pm 0.01$} & 0.37 {\scriptsize$\pm 0.01$} & \underline{0.51} {\scriptsize$\pm 0.01$} & 0.46 {\scriptsize$\pm 0.01$} & 0.49 {\scriptsize$\pm 0.01$} & \textbf{0.57} {\scriptsize$\pm 0.02$} \\
    Pascal       & 0.45 {\scriptsize$\pm 0.01$} & 0.46 {\scriptsize$\pm 0.01$} & \underline{0.71} {\scriptsize$\pm 0.01$} & 0.64 {\scriptsize$\pm 0.01$} & 0.68 {\scriptsize$\pm 0.01$} & \textbf{0.82} {\scriptsize$\pm 0.01$} \\
    \midrule
    PERSUADE     & 0.42 {\scriptsize$\pm 0.02$} & 0.42 {\scriptsize$\pm 0.02$} & 0.51 {\scriptsize$\pm 0.03$} & \underline{0.55} {\scriptsize$\pm 0.03$} & 0.29 {\scriptsize$\pm 0.03$} & \textbf{0.64} {\scriptsize$\pm 0.03$} \\
    El Capitan   & 0.47 {\scriptsize$\pm 0.03$} & 0.45 {\scriptsize$\pm 0.02$} & 0.52 {\scriptsize$\pm 0.03$} & \underline{0.58} {\scriptsize$\pm 0.03$} & 0.51 {\scriptsize$\pm 0.03$} & \textbf{0.70} {\scriptsize$\pm 0.02$} \\
    GoEmotions   & 0.19 {\scriptsize$\pm 0.03$} & 0.21 {\scriptsize$\pm 0.02$} & 0.28 {\scriptsize$\pm 0.02$} & \underline{0.35} {\scriptsize$\pm 0.03$} & 0.25 {\scriptsize$\pm 0.03$} & \textbf{0.45} {\scriptsize$\pm 0.03$} \\
    \bottomrule
\end{tabular}
% \vspace{-1em}
\end{table*}

\section{\supers{} are Reliable and Semantically Grounded Concept Signals}
\label{sec:experiments}

We now validate the theoretical insights from Section~\ref{sec:theory}, testing whether the high-activation tokens amplified by the \super{} Mechanism are reliable and semantically meaningful concept signals. 

Recall that  $Z$ denotes the collection of token embeddings across samples. For each concept $c$, we define \supers{} as the set of tokens whose activation scores exceed a concept-specific threshold $\tau_c^{\text{super}}$, shared across samples:
% \vspace{-2em}
\[
\mathcal{T}_{c}^{\text{super}}
=
\{\, z \in Z : s_c(z) \geq \tau_{c}^{\text{super}} \,\}.
\]
\vspace{-1.5em}

The threshold $\tau_c^{\text{super}}$ can be tuned or selected using a fixed tail sparsity level. 

We evaluate \supers{} through two tasks. (1) \textit{Concept detection} tests whether \supers{} consistently provide sample-level evidence of concept presence, while sweeping the \super{} threshold, $\tau_c^{\mathrm{super}}$, lets us investigate how narrowly this reliable evidence is concentrated in the high-activation tail. (2) \textit{Concept localization} tests whether local \super{} targets align with human-labeled concept regions rather than arbitrary extreme activations.

\textbf{Experimental Setup.}\hspace{0.5em}
\emph{Datasets:} \textsc{COCO}~\citep{Lin2014MicrosoftCC}, \textsc{Pascal}~\citep{everingham2010pascalvoc}, and \textsc{OpenSurfaces}~\citep{bell13opensurfaces} (via \textsc{Broden}~\citep{bau2017network}) for vision; \textsc{PERSUADE}~\citep{CROSSLEY2024100865}, \textsc{El Capitan}~\citep{ElCapitanDataset}, and augmented \textsc{GoEmotions}~\citep{Demszky2020GoEmotionsAD} for text; synthetic \textsc{CLEVR}~\citep{Johnson2016CLEVRAD} and augmented \textsc{iSarcasm}~\citep{Oprea2019iSarcasmAD} in the appendix.
\emph{Models:} \emph{CLIP ViT-L/14}~\citep{Radford2021LearningTV} (vision), \emph{Qwen3-Embedding-4B}~\citep{Zhang2025Qwen3EA} (text), and \emph{Llama-3.2-11B-Vision-Instruct}~\citep{Dubey2024TheL3} (both).
\emph{Concept vectors:} mean prototypes~\citep{Zou2023RepresentationEA}, linear separators~\citep{Kim2017InterpretabilityBF}, $k$-means prototypes~\citep{Ghorbani2019AutomatingID} and separators, and sparse autoencoder features~\citep{bricken2023monosemanticity}.
Full experiment details are in Appendix~\ref{app:experiment-configs}.

\subsection{\supers{} are Reliable Indicators of Concept Presence}
\label{sec:detection-results}

\textit{Concept detection} aims to determine whether a concept is present anywhere in a sample $x \in \mathcal{X}$~\citep{Wu2025AxBenchSL, Koh2020ConceptBM}. Because token activations vary within a sample, standard concept-vector approaches apply an aggregation operator $G: \mathbb{R}^{n(x)+1} \to \mathbb{R}$ to obtain a per-sample concept activation score:
\[
s^{\mathrm{agg}}_c(x) =
G\big(s_c(z_1^{\mathrm{tok}}(x)), \dots, s_c(z_{n(x)}^{\mathrm{tok}}(x)), s_c(z^{\mathrm{cls}}(x))\big).
\]
\vspace{-1.3em}

The concept is predicted to be present if $s^{\mathrm{agg}}_c(x)$ 
exceeds a decision threshold. 

\textbf{\super{}-Based Detector.}\hspace{0.5em}
Our initial theoretical (Sec~\ref{sec:theory}) and empirical (Sec ~\ref{sec:introducing_supers}) findings suggest that the high-activation tail contains the most reliable concept signals. This implies a detection rule: predict the concept present if the sample contains a SuperActivator, i.e. if some token activation lies in the high-activation tail. With a sparse tail, this is equivalent to maximum aggregation~\citep{Tillman2025InvestigatingTP, Wu2025AxBenchSL}:
% \vspace{-0.1em}
\[
G_{\max}(s_c(z^{\text{tok}}_1(x)), \ldots, s_c(z^{\text{tok}}_{n(x)}(x))) \;\geq\; \tau_c^{\text{super}}.
\]
For each concept, we compute the maximum token activation for every validation sample and select   $\tau_c^{\text{super}}$ to maximize $F_1$ in distinguishing concept-positive from concept-negative samples.

\begin{wrapfigure}{r}{0.49\textwidth}
    \centering
    % \vspace{-1em}
    \includegraphics[width=\linewidth]{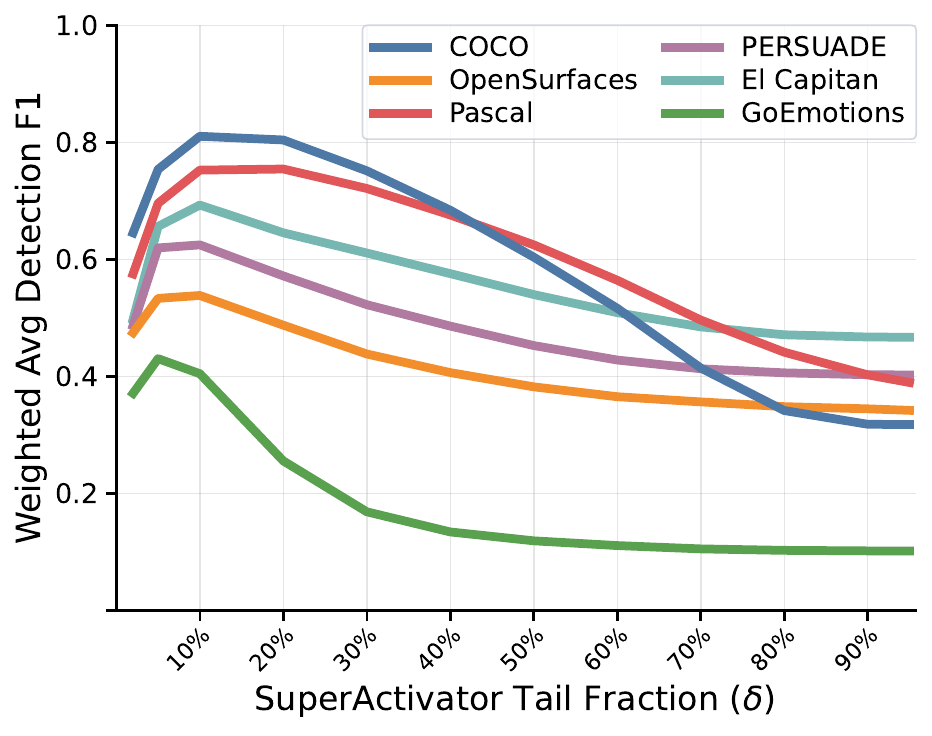}
    \caption{\textbf{Reliable concept detection requires isolating the extreme activation tail.} Detection peaks when only the top $5$--$10\%$ of in-concept activations exceed the threshold and declines as more of the distribution is included.}
    \vspace{-1em}
    \label{fig:percentile-hist}

\end{wrapfigure}

\textbf{Baselines and Tuning.}\hspace{0.5em}
We compare against several common choices of the aggregation operator $G$:
CLSTok, which selects the [CLS] token activation~\citep{Yu2024LatentCE}; MeanTok, which averages token activations~\citep{McKenzie2025DetectingHI}; LastTok, which selects the final token activation~\citep{Chen2025PersonaVM}; and RandTok, which selects a random token activation. We also include a prompting baseline for compatible models, directly querying whether the target concept is present~\citep{Wu2025AxBenchSL}; see Appendix~\ref{app:concept-extraction-techniques} for details.
For all vector-based methods, we select the model layer and decision threshold (over the sample-level aggregated scores) that maximize each concept's validation $F_1$, and report frequency-weighted test $F_1$ across concepts.

\textbf{Detection Results.}\hspace{0.5em} 
As predicted by the \super{} Mechanism, using only the most extreme token activations for concept detection consistently outperforms all other evaluated strategies on \emph{Llama-3.2-11B-Vision-Instruct} linear separator concepts (Table~\ref{tab:detection_f1_summary}). Appendix~\ref{app:detection-results} shows that these gains hold consistently across various models and concept vector types. The strongest alternative typically varies among Prompt, CLSTok, and MeanTok.

\textbf{Detection Peaks at High Sparsity.}\hspace{0.5em}
Strong max aggregation performance alone does not establish that reliable concept evidence is confined to a sparse tail: the maximum could remain predictive even if a much broader portion of $D_c^{\mathrm{in}}$ was well-separated from $D_c^{\mathrm{out}}$. To measure how concentrated the reliable concept evidence really is, we retain the max aggregation rule and sweep a sparsity fraction $\delta_c$ over a fixed grid. For each $\delta_c$, we set $\tau_c^{\text{super}} = Q_{1-\delta_c}(D_c^{\mathrm{in}})$, so that only the top $\delta_c$ fraction of in-concept activations in the validation set exceed the threshold.

Figure~\ref{fig:percentile-hist} shows that detection $F_1$ for most \emph{Llama-3.2-11B-Vision-Instruct} linear separator concepts peaks when only the top $5$--$10\%$ of in-concept activations exceed the threshold. Although this excludes the vast majority of genuinely in-concept 
activations, lowering the threshold to capture more of them also admits too  much out-of-concept noise, 
ultimately degrading detection performance. Similar trends hold across models and concept types (Appendix~\ref{app:sparsity-ablation}), and we show that a fixed $\delta=10\%$ detector outperforms the strongest baseline on seven of eight datasets (with overlapping confidence intervals on the eighth) and remains competitive with the fully tuned detector (Appendix~\ref{app:fixed-delta-detection}). Thus, max aggregation is effective \textit{not} because a large portion of the in-concept distribution is broadly separated, but because a small high-activation tail is.

\textbf{Appendix Experiments Validate Robustness and Scope of \supers{}.}\hspace{0.5em}
We further investigate the robustness of \supers{} across tail sparsity levels (Appendices~\ref{app:sparsity-ablation}, \ref{app:superdetector-cdfs}, \ref{app:optimal-sparsity-across-layers}), model depths (Appendices~\ref{app:detection-results-across-layers}, \ref{app:best-layer-per-concept}, \ref{app:qual-super-over-layers}), and token positions (Appendix~\ref{app:pos-dependencies}). \supers{} also generalize across concept vector types, with one exception: for sparse autoencoder (SAE) concepts, detection remains competitive but performance is similar across tail fractions $\delta$, likely because the SAE training objective already
encourages sparsity (Appendix~\ref{app:sae-results}). Together, these analyses demonstrate that \supers{} are generally robust across settings.

\subsection{\supers{} Improve Attributions for Concepts}
% \subsection{\textcolor{blue}{\supers{} Reflect Localized Concept Semantics}}
\label{sec:inversion}

\begin{figure*}[!t]
    \centering
    \setlength{\fboxsep}{0pt}
    \begin{subfigure}[t]{0.31\textwidth}
    \includegraphics[width=\linewidth]{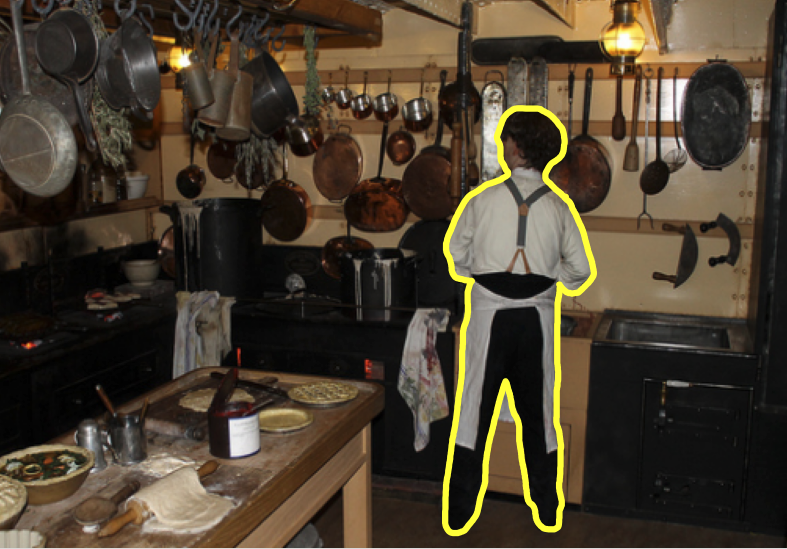}
        \caption{Image with \textit{Person} labeled}
        \label{fig:lime-orig}
    \end{subfigure}% 
    \hspace{0.01\textwidth}
    \begin{subfigure}[t]{0.31\textwidth}
    \includegraphics[width=\linewidth]{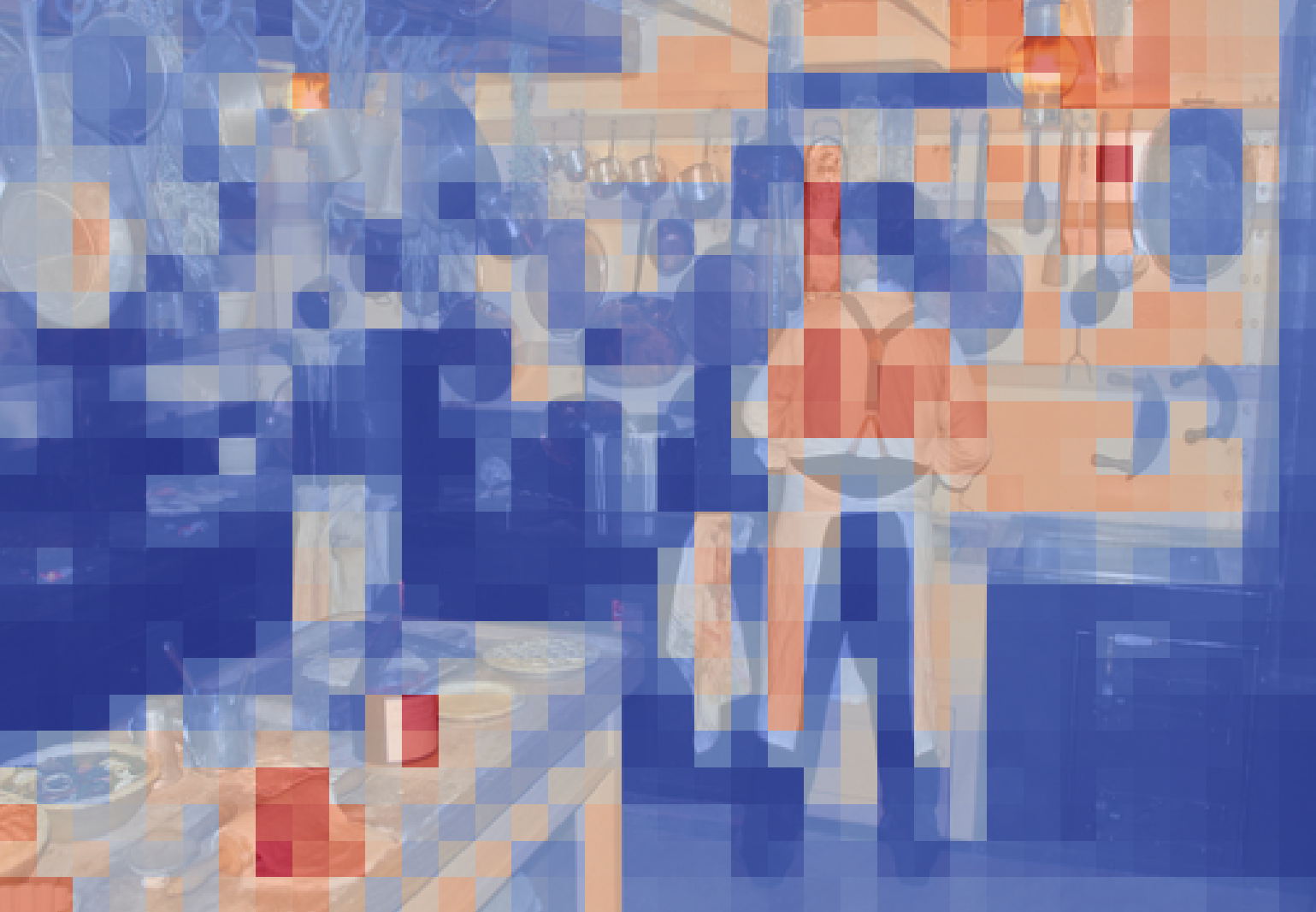}
        \caption{Global Concept Vector Target}
        \label{fig:lime-global}
    \end{subfigure}% 
    \hspace{0.01\textwidth}
    \begin{subfigure}[t]{0.31\textwidth}
        \includegraphics[width=\linewidth]{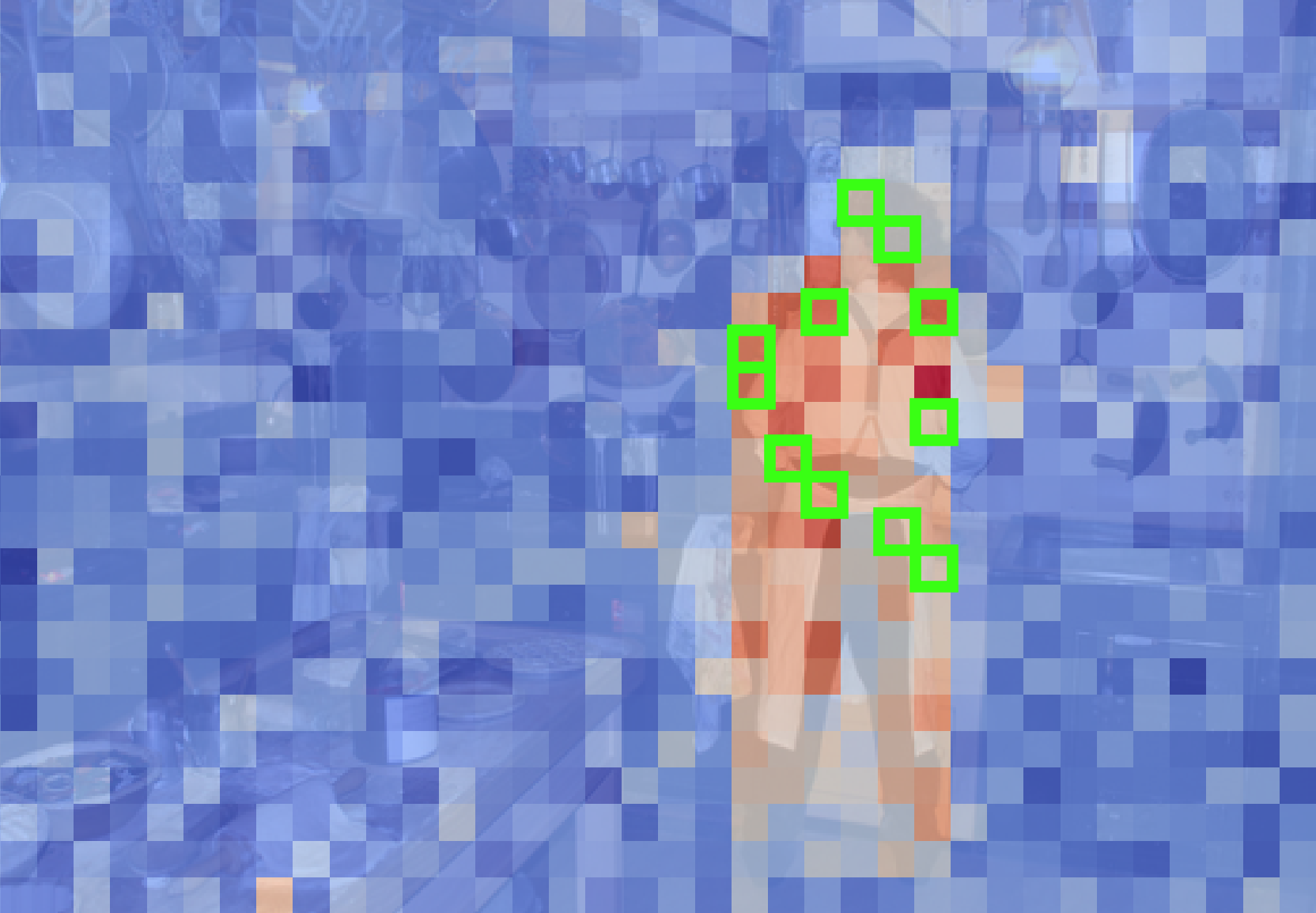}
        \caption{Local \supers{} Target}
        \label{fig:lime-super}
    \end{subfigure}% 
    \caption{\textbf{\supers{} yield attribution maps that better align with ground-truth concept regions.} LIME attribution maps for the \textit{Person} concept using (b) the global concept vector $v_c$ and (c) the local \super{}-derived target $\mu_c(x)$, computed by averaging the \supers{} marked in green. Red indicates high concept alignment and blue indicates low alignment.}
    \vspace{-1em}

    \label{fig:lime-example}
\end{figure*}

\begin{table}[t]
\caption{\textbf{Attribution maps explaining alignment to local \supers{} achieve higher accuracy and faithfulness than those explaining alignment to global concept vectors.} Average $F_1$ against ground-truth masks ($\uparrow$), insertion ($\uparrow$), and deletion ($\downarrow$) across nine attribution methods for \emph{Llama} separators on \emph{COCO} and \emph{PERSUADE}. Reported uncertainties are standard errors of the mean across localization runs. See Appendix~\ref{app:full-inversion} for additional datasets, models, and concept types.}

\label{tab:filtered}
\centering
\small
\setlength{\tabcolsep}{3pt}
\resizebox{\textwidth}{!}{
\begin{tabular}{@{}llcccc@{\hspace{1.0em}}cc@{}}
\toprule
\begin{tabular}[t]{@{}c@{}}Attribution\\Method\end{tabular} & \begin{tabular}[t]{@{}c@{}}Dataset\end{tabular}
& \multicolumn{2}{c}{\begin{tabular}[t]{@{}c@{}}Attribution $F_1$ \\ {\scriptsize ($\uparrow$ is better)}\end{tabular}}
& \multicolumn{2}{c}{\begin{tabular}[t]{@{}c@{}}Insertion Score \\ {\scriptsize ($\uparrow$ is better)}\end{tabular}}
& \multicolumn{2}{c}{\begin{tabular}[t]{@{}c@{}}Deletion Score \\ {\scriptsize ($\downarrow$ is better)}\end{tabular}} \\
\cmidrule(lr){3-4} \cmidrule(lr){5-6} \cmidrule(lr){7-8}
& & Concept & SuperActs & Concept & SuperActs & Concept & SuperActs \\
\midrule
\multirow{2}{*}{\shortstack[l]{LIME\\[-0.25ex]{\scriptsize \citep{ribeiro2016should}}}} & COCO & $0.49\,{\scriptstyle \pm 0.02}$ & \textbf{\boldmath$0.50\,{\scriptstyle \pm 0.03}$} & $0.397\,{\scriptstyle \pm 0.008}$ & \textbf{\boldmath$0.438\,{\scriptstyle \pm 0.009}$} & $0.007\,{\scriptstyle \pm 0.001}$ & \textbf{\boldmath$0.005\,{\scriptstyle \pm 0.001}$} \\
& PERSUADE & $0.32\,{\scriptstyle \pm 0.05}$ & \textbf{\boldmath$0.40\,{\scriptstyle \pm 0.06}$} & $0.128\,{\scriptstyle \pm 0.003}$ & \textbf{\boldmath$0.131\,{\scriptstyle \pm 0.003}$} & $0.018\,{\scriptstyle \pm 0.003}$ & \textbf{\boldmath$0.012\,{\scriptstyle \pm 0.003}$} \\
\midrule
\multirow{2}{*}{\shortstack[l]{SHAP\\[-0.25ex]{\scriptsize \citep{lundberg2017unified}}}} & COCO & $0.49\,{\scriptstyle \pm 0.02}$ & \textbf{\boldmath$0.55\,{\scriptstyle \pm 0.04}$} & $0.403\,{\scriptstyle \pm 0.008}$ & \textbf{\boldmath$0.439\,{\scriptstyle \pm 0.003}$} & $0.006\,{\scriptstyle \pm 0.001}$ & \textbf{\boldmath$0.005\,{\scriptstyle \pm 0.002}$} \\
& PERSUADE & $0.40\,{\scriptstyle \pm 0.07}$ & \textbf{\boldmath$0.43\,{\scriptstyle \pm 0.06}$} & $0.109\,{\scriptstyle \pm 0.002}$ & \textbf{\boldmath$0.122\,{\scriptstyle \pm 0.001}$} & $0.020\,{\scriptstyle \pm 0.003}$ & \textbf{\boldmath$0.017\,{\scriptstyle \pm 0.002}$} \\
\midrule
\multirow{2}{*}{\shortstack[l]{RISE\\[-0.25ex]{\scriptsize \citep{petsiuk2018rise}}}} & COCO & $0.35\,{\scriptstyle \pm 0.03}$ & \textbf{\boldmath$0.40\,{\scriptstyle \pm 0.02}$} & $0.394\,{\scriptstyle \pm 0.005}$ & \textbf{\boldmath$0.439\,{\scriptstyle \pm 0.009}$} & $0.007\,{\scriptstyle \pm 0.002}$ & \textbf{\boldmath$0.005\,{\scriptstyle \pm 0.000}$} \\
& PERSUADE & $0.30\,{\scriptstyle \pm 0.05}$ & \textbf{\boldmath$0.35\,{\scriptstyle \pm 0.06}$} & $0.195\,{\scriptstyle \pm 0.004}$ & \textbf{\boldmath$0.221\,{\scriptstyle \pm 0.005}$} & $0.022\,{\scriptstyle \pm 0.003}$ & \textbf{\boldmath$0.020\,{\scriptstyle \pm 0.002}$} \\
\midrule
\multirow{2}{*}{\shortstack[l]{SHAP IQ\\[-0.25ex]{\scriptsize \citep{fel2023shap}}}} & COCO & $0.36\,{\scriptstyle \pm 0.01}$ & \textbf{\boldmath$0.38\,{\scriptstyle \pm 0.03}$} & $0.397\,{\scriptstyle \pm 0.004}$ & \textbf{\boldmath$0.433\,{\scriptstyle \pm 0.004}$} & $0.007\,{\scriptstyle \pm 0.002}$ & \textbf{\boldmath$0.006\,{\scriptstyle \pm 0.001}$} \\
& PERSUADE & \textbf{\boldmath$0.39\,{\scriptstyle \pm 0.06}$} & \textbf{\boldmath$0.39\,{\scriptstyle \pm 0.04}$} & $0.193\,{\scriptstyle \pm 0.005}$ & \textbf{\boldmath$0.208\,{\scriptstyle \pm 0.004}$} & $0.013\,{\scriptstyle \pm 0.002}$ & \textbf{\boldmath$0.010\,{\scriptstyle \pm 0.002}$} \\ 
\midrule
\multirow{2}{*}{\shortstack[l]{IntGrad\\[-0.25ex]{\scriptsize \citep{sundararajan2017axiomatic}}}} & COCO & $0.43\,{\scriptstyle \pm 0.02}$ & \textbf{\boldmath$0.48\,{\scriptstyle \pm 0.01}$} & $0.397\,{\scriptstyle \pm 0.004}$ & \textbf{\boldmath$0.437\,{\scriptstyle \pm 0.009}$} & $0.009\,{\scriptstyle \pm 0.003}$ & \textbf{\boldmath$0.006\,{\scriptstyle \pm 0.003}$} \\
& PERSUADE & $0.40\,{\scriptstyle \pm 0.07}$ & \textbf{\boldmath$0.43\,{\scriptstyle \pm 0.07}$} & $0.112\,{\scriptstyle \pm 0.003}$ & \textbf{\boldmath$0.124\,{\scriptstyle \pm 0.002}$} & $0.022\,{\scriptstyle \pm 0.001}$ & \textbf{\boldmath$0.019\,{\scriptstyle \pm 0.001}$} \\ 
\midrule
\multirow{2}{*}{\shortstack[l]{GradCAM\\[-0.25ex]{\scriptsize \citep{selvaraju2017grad}}}} & COCO & \textbf{\boldmath$0.37\,{\scriptstyle \pm 0.01}$} & \textbf{\boldmath$0.37\,{\scriptstyle \pm 0.02}$} & $0.398\,{\scriptstyle \pm 0.004}$ & \textbf{\boldmath$0.434\,{\scriptstyle \pm 0.008}$} & $0.007\,{\scriptstyle \pm 0.003}$ & \textbf{\boldmath$0.006\,{\scriptstyle \pm 0.001}$} \\
& PERSUADE & $0.39\,{\scriptstyle \pm 0.07}$ & \textbf{\boldmath$0.45\,{\scriptstyle \pm 0.08}$} & $0.163\,{\scriptstyle \pm 0.005}$ & \textbf{\boldmath$0.167\,{\scriptstyle \pm 0.005}$} & $0.012\,{\scriptstyle \pm 0.003}$ & \textbf{\boldmath$0.011\,{\scriptstyle \pm 0.003}$} \\  
\midrule
\multirow{2}{*}{\shortstack[l]{FullGrad\\[-0.25ex]{\scriptsize \citep{srinivas2019full}}}} & COCO & \textbf{\boldmath$0.39\,{\scriptstyle \pm 0.01}$} & \textbf{\boldmath$0.39\,{\scriptstyle \pm 0.03}$} & $0.393\,{\scriptstyle \pm 0.005}$ & \textbf{\boldmath$0.430\,{\scriptstyle \pm 0.010}$} & $0.007\,{\scriptstyle \pm 0.001}$ & \textbf{\boldmath$0.006\,{\scriptstyle \pm 0.002}$} \\
& PERSUADE & $0.25\,{\scriptstyle \pm 0.03}$ & \textbf{\boldmath$0.27\,{\scriptstyle \pm 0.02}$} & $0.199\,{\scriptstyle \pm 0.005}$ & \textbf{\boldmath$0.215\,{\scriptstyle \pm 0.005}$} & $0.017\,{\scriptstyle \pm 0.003}$ & \textbf{\boldmath$0.015\,{\scriptstyle \pm 0.002}$} \\ 
\midrule
\multirow{2}{*}{\shortstack[l]{CALM\\[-0.25ex]{\scriptsize \citep{kim2021calm}}}} & COCO & $0.38\,{\scriptstyle \pm 0.02}$ & \textbf{\boldmath$0.41\,{\scriptstyle \pm 0.01}$} & $0.394\,{\scriptstyle \pm 0.007}$ & \textbf{\boldmath$0.432\,{\scriptstyle \pm 0.004}$} & $0.007\,{\scriptstyle \pm 0.002}$ & \textbf{\boldmath$0.005\,{\scriptstyle \pm 0.000}$} \\
& PERSUADE & \textbf{\boldmath$0.31\,{\scriptstyle \pm 0.04}$} & \textbf{\boldmath$0.31\,{\scriptstyle \pm 0.03}$} & $0.153\,{\scriptstyle \pm 0.005}$ & \textbf{\boldmath$0.162\,{\scriptstyle \pm 0.005}$} & $0.016\,{\scriptstyle \pm 0.003}$ & \textbf{\boldmath$0.013\,{\scriptstyle \pm 0.003}$} \\
\midrule
\multirow{2}{*}{\shortstack[l]{MFABA\\[-0.25ex]{\scriptsize \citep{zhu2024mfaba}}}} & COCO & $0.35\,{\scriptstyle \pm 0.02}$ & \textbf{\boldmath$0.39\,{\scriptstyle \pm 0.01}$} & $0.401\,{\scriptstyle \pm 0.002}$ & \textbf{\boldmath$0.443\,{\scriptstyle \pm 0.009}$} & $0.004\,{\scriptstyle \pm 0.001}$ & \textbf{\boldmath$0.003\,{\scriptstyle \pm 0.001}$} \\
& PERSUADE & $0.26\,{\scriptstyle \pm 0.02}$ & \textbf{\boldmath$0.30\,{\scriptstyle \pm 0.02}$} & $0.132\,{\scriptstyle \pm 0.002}$ & \textbf{\boldmath$0.143\,{\scriptstyle \pm 0.002}$} & $0.016\,{\scriptstyle \pm 0.002}$ & \textbf{\boldmath$0.013\,{\scriptstyle \pm 0.001}$} \\
\bottomrule
\end{tabular}}
\vspace{-0.5em}
\end{table}

Concept localization asks \textit{where} a concept appears within a sample~\citep{desantis2025visualtcav}. A common approach is to generate attribution maps, which estimate how strongly each token provides evidence for the concept. These maps should be both \textit{accurate}, aligning with ground-truth annotations, and \textit{faithful}~\citep{Jacovi2020TowardsFI}, identifying features that drive the model's concept alignment rather than spurious correlations. We measure faithfulness with perturbation-based \textit{insertion} and \textit{deletion} metrics~\citep{petsiuk2018rise}, which track how the target objective changes as highly attributed tokens are added or removed.

% To localize a concept within an input, a common approach is to generate an attribution map that estimates each token’s contribution to the target concept. 

We generate attribution maps using nine techniques described in Appendix~\ref{app:inversion_methods}, such as SHAP-IQ~\citep{fel2023shap}, GradCAM~\citep{selvaraju2017grad}, and MFABA~\citep{zhu2024mfaba}. All methods explain the same scalar objective: the average cosine similarity between the token embeddings and a target embedding $t \in \mathbb{R}^d$. We formalize these methods with a generic attribution operator $\mathcal{E}(x,t)$, which produces token-level scores $\{\mathcal{E}_i(x,t)\}_{i=1}^{n(x)}$ for an input $x$ with $n(x)$ tokens, measuring each token’s contribution to alignment with the target.

\textbf{\super{}-Based Localization.}\hspace{0.5em}
Prior work typically uses the global concept vector $v_c$ as the attribution target~\citep{desantis2025visualtcav, lucieri2020explaining}. Because this vector is aggregated across many samples rather than instance-specific, it can be less precise and obscure localized evidence for concept presence. In contrast, \supers{} provide reliable, context-specific concept signals; therefore, we replace the global target with a local target derived from the \supers{} present in each sample.

Recall that $z_i^{\text{tok}}(x) \in \mathbb{R}^d$ is the $i$-th token embedding in input $x$ from layer $l$, and that $\mathcal{T}_{c}^{\text{super}}$ is the set of \supers{} for concept $c$, defined using the detection-optimized threshold $\tau_c^{\text{super}}$. For a given input $x$, we define the set of \supers{} present in $x$ as
$\mathcal{Z}_c^{\text{super}}(x)=\{\, z_i^{\text{tok}}(x) : z_i^{\text{tok}}(x) \in \mathcal{T}_{c}^{\text{super}} \,\}$.
We then define the \super{}-derived attribution target as the mean embedding of the local \supers{}:
\vspace{-0.5 em}
\[
\mu_c(x)
=
|\mathcal{Z}_c^{\text{super}}(x)|^{-1}
\sum_{z \in \mathcal{Z}_c^{\text{super}}(x)} z \in \mathbb{R}^d .
\]
\vspace{-1.25 em}

For each attribution method, we compare $\mathcal{E}_i(x,\mu_c(x))$ to the standard $\mathcal{E}_i(x,v_c)$ on the same set of in-concept test samples that contain at least one \super{}.

\textbf{Attribution Accuracy Results.}\hspace{0.5em}
Attribution maps that explain alignment to the local \super{}-derived target $\mu_c(x)$ align more closely with ground-truth concept regions than those explaining alignment to the global concept vector $v_c$. For \emph{Llama-3.2-11B-Vision-Instruct} linear separators on \textsc{COCO} and \textsc{PERSUADE}, the local target matches or improves attribution $F_1$ for all nine attribution methods (Table~\ref{tab:filtered}). This generally holds across datasets, models, and concept types (Tables~\ref{tab:inversion_clevr}--\ref{tab:inversion_isarcasm}).

We illustrate these gains in Figure~\ref{fig:lime-example}, using LIME to localize the \textit{Person} concept in a \emph{COCO} image with a \emph{Llama-3.2-11B-Vision-Instruct} linear separator concept. The $v_c$-derived map assigns high attribution to broad, semantically irrelevant regions, including likely correlated objects such as jars and pans. In contrast, the \super{}-derived  map concentrates more precisely on the labeled \textit{Person} region. Additional image and text visualizations are provided in Appendix~\ref{app:inversion-qual-examples}.

\textbf{Attribution Faithfulness Results.}\hspace{0.5em}
\super{}-based attributions achieve higher insertion and lower deletion scores than those obtained using global concept vectors across all attribution methods and datasets (Table~\ref{tab:filtered}, Table~\ref{tab:inversion_llama_insdel}). This suggests that the local \super{}-derived target $\mu_c(x)$ better identifies tokens that influence the model's concept alignment.

\section{Related Work}

\textbf{Concept-Based Interpretability.}\hspace{0.5em}  
Concept-based interpretability seeks to link model internals to human-understandable features, often by representing concepts as linear separators (e.g., TCAV~\citep{Kim2017InterpretabilityBF}) or labeled centroid embeddings~\citep{Zou2023RepresentationEA}. Unsupervised methods instead discover concepts through ACE~\citep{Ghorbani2019AutomatingID}, hierarchical clustering~\citep{Dalvi2022DiscoveringLC}, matrix factorization~\citep{zhang2021invertible, Fel2022CRAFTCR}, or sparse autoencoders~\citep{bricken2023monosemanticity, Cunningham2023SparseAF, Gao2024ScalingAE}. Across these approaches, concepts are treated as recoverable structures in representation space.

\textbf{Challenges in Concept Representations.}\hspace{0.5em}  
Many open questions remain about the structure of concept representations. Concept activations are often noisy: they can be \emph{entangled} with unintended semantics \citep{goh2021multimodal, Olah2020ZoomIA}, \emph{polysemantic} across multiple features \citep{bricken2023monosemanticity, OMahony2023DisentanglingNR}, and \emph{unstable} across layers, locations, exemplar sets, and random seeds \citep{Wu2025AxBenchSL, Mahinpei2021PromisesAP, Nicolson2024ExplainingER, Mikriukov2023EvaluatingTS}. These issues can amplify spurious correlations \citep{Zhou2023ExploreSC} and concept leakage \citep{Parisini2025LeakageAI}. While prior work modifies training or extraction to encourage interpretable, disentangled, or compositional concepts \citep{Chen2020ConceptWF, Wang2022DisentangledRL, Stein2024TowardsCI}, we instead identify a sparse, reliable signal already present within noisy activation distributions.

\textbf{Theoretical Analyses of Self-Attention Dynamics.}\hspace{0.5em}
Motivated by transformer-circuits work that decomposes attention heads into query-key routing and value/output writing components~\citep{elhage2021mathematical}, we analyze how self-attention dynamics evolve when this read-write structure is aligned with a semantic concept direction. Another line of work studies self-attention as a dynamical system, showing that pure attention can drive token representations toward rank collapse or token consensus~\citep{pmlr-v139-dong21a, noci2022signal, pmlr-v267-abella25a}. Complementary dynamical-systems perspectives view self-attention as an interacting particle system, proving the emergence of clusters, leaders, and metastable configurations before eventual collapse~\citep{geshkovski2023emergence, Geshkovski2024DynamicMetastability}. Related attention-sink and activation-outlier work studies extreme-token phenomena that arise from attention dynamics~\citep{Guo2024ActiveDormant, an2025systematic}. In contrast to these works, which analyze generic token dynamics, we study a concept-conditioned scalar projection of self-attention dynamics.

\textbf{Concept Detection and Localization.}\hspace{0.5em}
Concept detection asks \textit{whether} a concept is present in a sample and is a central task for concept-based interpretability~\citep{Wu2025AxBenchSL}. 
Most approaches score samples by concept activation, using either global representations such as [CLS] tokens or pooled embeddings~\citep{Choi2021EvaluationOB,Tang2024PoolingAA}, or token-level activations aggregated into a sample-level score via [CLS]-based scoring~\citep{Nejadgholi2022TowardsPF,Yu2024LatentCE,Behrendt2025MaxPoolBERTEB}, mean pooling~\citep{McKenzie2025DetectingHI,Suresh2025FromNT}, max pooling~\citep{Tillman2025InvestigatingTP,Wu2025AxBenchSL}, or last-token scoring~\citep{Chen2025PersonaVM,Tillman2025InvestigatingTP,Tang2024PoolingAA}. 
Recent vision--language models also enable zero-shot prompting approaches that bypass explicit concept vectors~\citep{Wu2025AxBenchSL,Robicheaux2025Roboflow100VLAM,Tillman2025InvestigatingTP}. 
Complementarily, concept localization asks \textit{where} a concept appears within a sample~\citep{desantis2025visualtcav}. 
Attribution methods such as Integrated Gradients~\citep{sundararajan2017axiomatic} and Grad-CAM~\citep{selvaraju2017grad}, along with concept-based adaptations~\citep{desantis2025visualtcav,Fel2022CRAFTCR,kuroki2025fam,lucieri2020explaining}, highlight input tokens aligned with the concept; other work localizes concepts using raw concept activation scores~\citep{Lim2024SparseAR,Zhou2023ELVITPV} or attention values~\citep{Gandelsman2023InterpretingCI}.

\section{Discussion and Limitations}
\label{sec:limitations}
We introduced the \super{} Mechanism, showing that transformers push a small subset of strong concept signals into a high-activation tail. Leveraging this structure helps filter noisy concept activations and yields stronger signals for both concept detection and localization.

Several limitations and open questions remain. Although our theory establishes that transformers amplify tail asymmetry, it does not explain how this initial asymmetry arises, and extending the analysis to richer transformer interactions remains an open problem. Our detection approach also requires validation data to estimate \super{} thresholds, as do other threshold-based methods, and further analysis is needed to explain why optimal sparsity varies across concepts.
\section*{Acknowledgments}
This research was partially supported by a gift from AWS AI to Penn Engineering's ASSET Center for Trustworthy AI, by ARPA-H program on Safe and Explainable AI under the award D24AC00253-00, by NSF award SLES 2331783, and by an NSF Graduate Research Fellowship under award DGE-2236662. The authors declare no competing interests.

\bibliography{references}
\bibliographystyle{plainnat}

%%%%%%%%%%%%%%%%%%%%%%%%%%%%%%%%%%%%%%%%%%%%%%%%%%%%%%%%%%%%%%%%%%%%%%%%%%%%%%%
% APPENDIX
%%%%%%%%%%%%%%%%%%%%%%%%%%%%%%%%%%%%%%%%%%%%%%%%%%%%%%%%%%%%%%%%%%%%%%%%%%%%%%%
\newpage
\appendix
\clearpage
\section{\super{} Visual Examples}

\label{app:superactivator-examples}
This section presents additional examples of \supers{} across image and text datasets. Heatmaps show signed activation scores between token embeddings and labeled concept vectors, with red indicating positive alignment, blue negative alignment, and green boxes marking \supers{}. For each concept, we use a fixed color scale across all in-concept test samples, with \texttt{vmin} and \texttt{vmax} computed globally and a zero-centered \texttt{TwoSlopeNorm(vcenter=0)}, allowing activation magnitudes to be compared directly across examples.

For image heatmaps, token-level scores are upsampled with nearest-neighbor interpolation and lightly Gaussian-smoothed ($\sigma=0.6$) for visualization only; \supers{} are selected from the original unsmoothed scores. Concept vectors are linear separators trained on \emph{Llama-3.2-11B-Vision-Instruct} embeddings at the depth with highest validation performance, and \supers{} are defined using the sparsity level~$\delta$ yielding the highest validation~$F_1$ for each concept.

\begin{figure}[H]
  \centering
    \includegraphics[width=\textwidth]{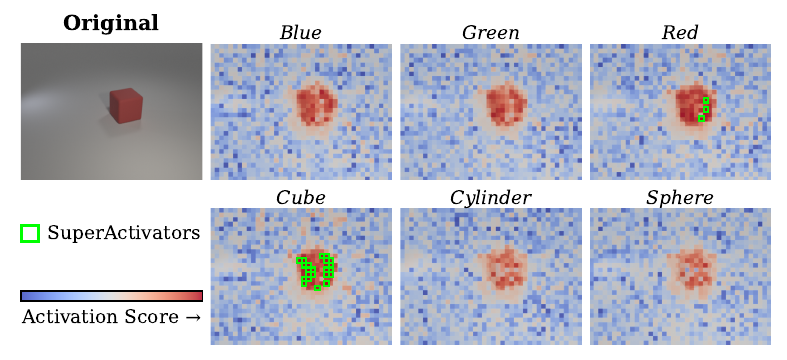}
    \vspace{-0.5cm}
    \caption{\emph{CLEVR} -- Visualization of Concept Activations and \supers{}}
  \label{fig:superpatch-example-clevr}
\end{figure}

% \begin{figure}[H]
%   \centering
%     \includegraphics[width=\textwidth]{Figures/Llama_Coco_superpatch_example_806_linsep_concepts_BD_True_BN_False_Llama_patch_embeddings_percentthrumodel_100.pt.pdf}
%     \vspace{-0.5cm}
%     \caption{\emph{COCO} -- Visualization of Concept Activations and \supers{}}
%   \label{fig:superpatch-example-coco}
% \end{figure}

\begin{figure}[H]
  \centering
    \includegraphics[width=\textwidth]{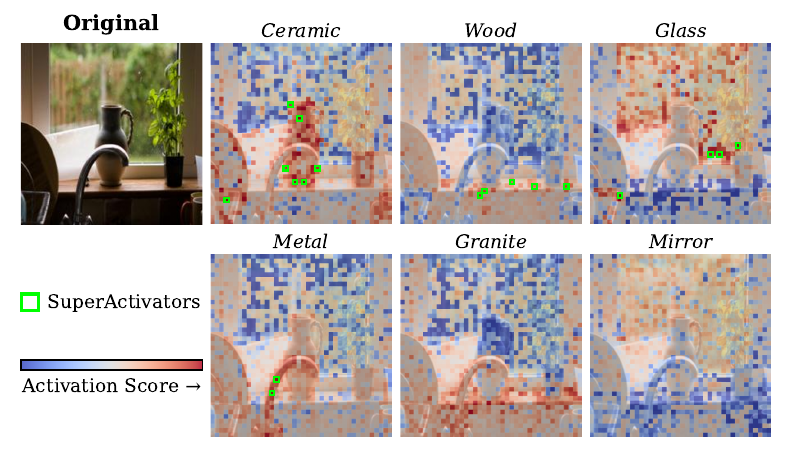}
    \vspace{-0.5cm}
    \caption{\emph{OpenSurfaces} -- Visualization of Concept Activations and \supers{}}
  \label{fig:superpatch-example-opensurfaces}
\end{figure}

\begin{figure}[H]
  \centering
    \includegraphics[width=\textwidth]{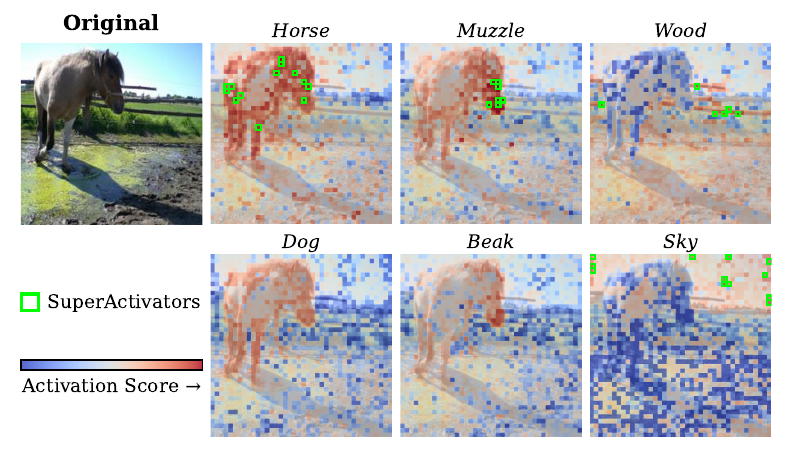}
    \vspace{-0.5cm}
    \caption{\emph{Pascal} -- Visualization of Concept Activations and \supers{}}
  \label{fig:superpatch-example-pascal}
\end{figure}

\begin{figure}[H]
  \centering
    \includegraphics[width=\textwidth]{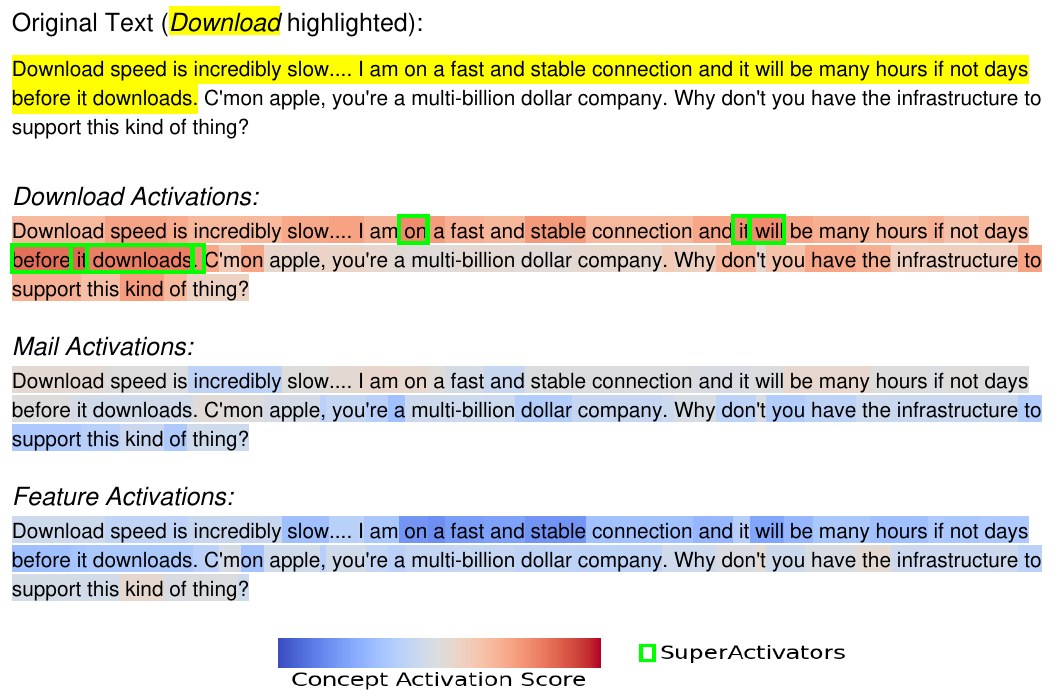}
    \vspace{-0.5cm}
    \caption{\emph{El Capitan} -- Visualization of Concept Activations and \supers{}}
  \label{fig:superpatch-example-elcapitan}
\end{figure}

\begin{figure}[H]
  \centering
    \includegraphics[width=\textwidth]{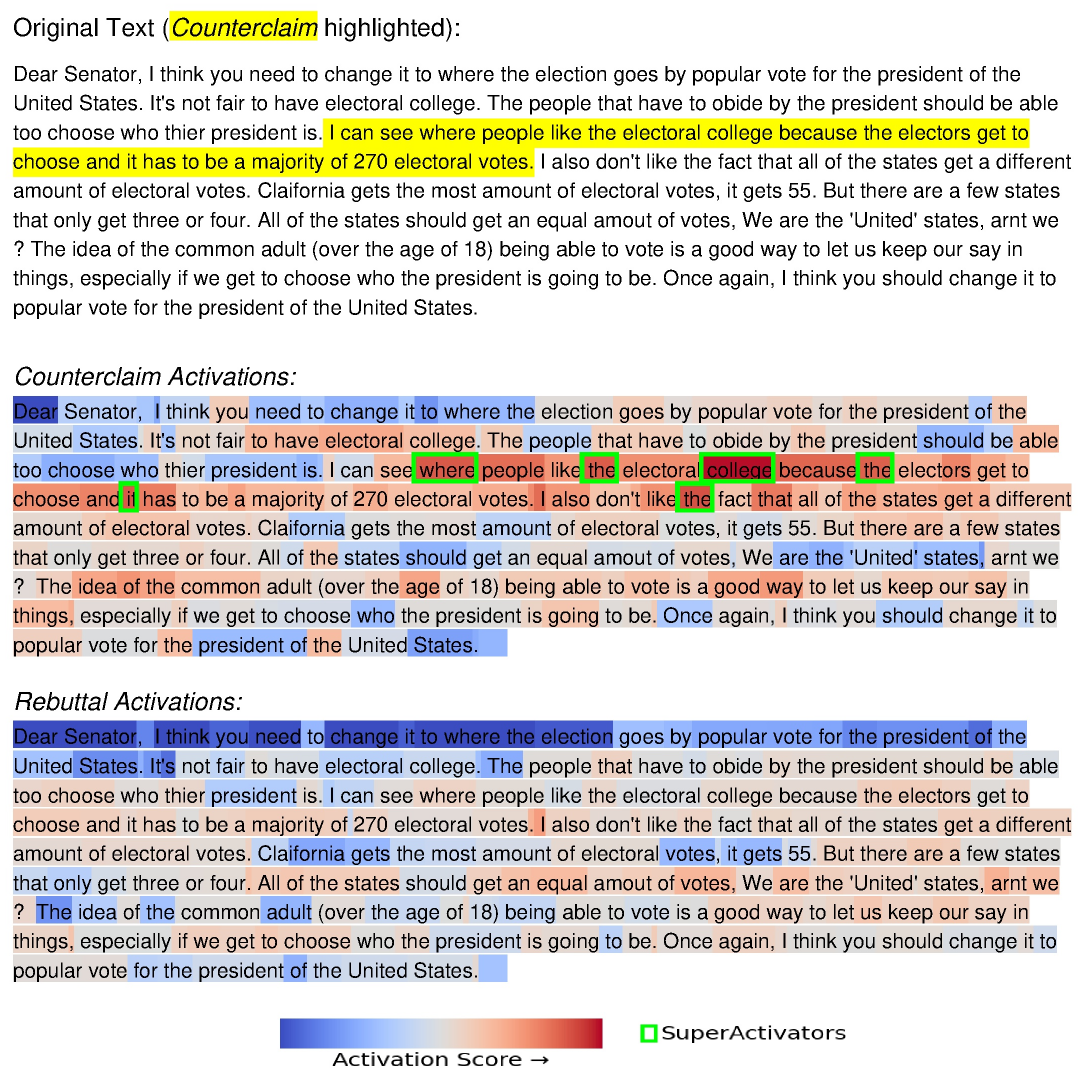}
    \vspace{-0.5cm}
    \caption{\emph{PERSUADE} -- Visualization of Concept Activations and \supers{}}
  \label{fig:superpatch-example-persuade}
\end{figure}

\begin{figure}[H]
  \centering
    \includegraphics[width=\textwidth]{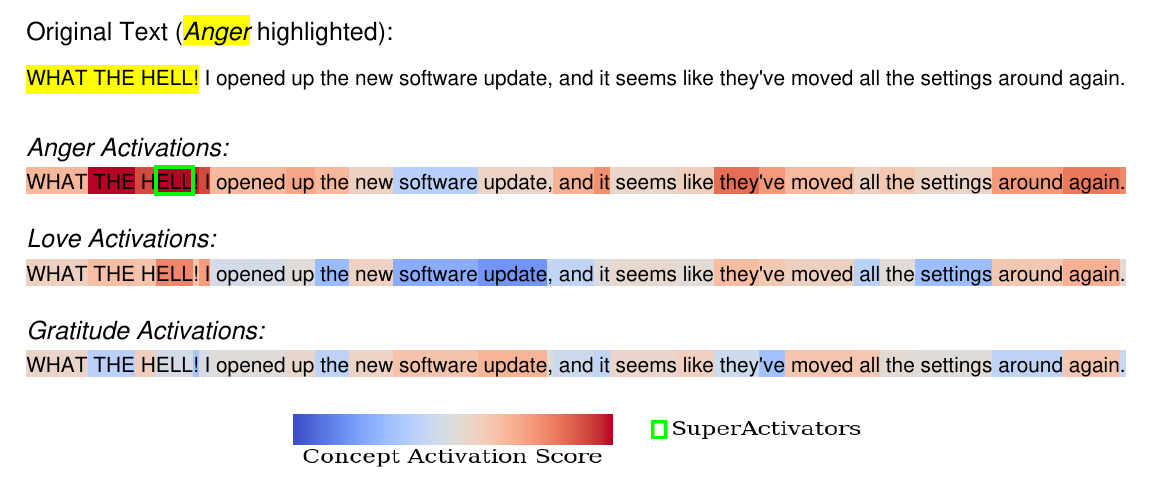}
    \vspace{-0.5cm}
    \caption{\emph{Augmented GoEmotions} -- Visualization of Concept Activations and \supers{}}
  \label{fig:superpatch-example-goemotions}
\end{figure}

% \begin{figure}[H]
%   \centering
%   % Top subfigure
%   \begin{subfigure}[t]{\textwidth}
%     \centering
%     \includegraphics[width=\textwidth]{Figures/Llama_Sarcasm_supertoken_example_742_linsep_concepts_BD_True_BN_False_Llama_patch_embeddings_percentthrumodel_100.pt.pdf}
%     \caption{Non-Sarcastic Version}
%   \end{subfigure}
%     % Bottom subfigure

%   \begin{subfigure}[h!]{\textwidth}
%     \centering
%     \includegraphics[width=\textwidth]{Figures/Llama_Sarcasm_supertoken_example_732_linsep_concepts_BD_True_BN_False_Llama_patch_embeddings_percentthrumodel_100.pt.pdf}
%     \caption{Sarcastic Version}
%   \end{subfigure}

%   \caption{\emph{Sarcasm} -- Visualization of Concept Activations and \supers{} (sarcastic and non-sarcastic version of same sentiment)}
%   \label{fig:superpatch-example-sarcasm}
% \end{figure}

\begin{figure}[H]
  \centering
  % Top subfigure
    \begin{subfigure}[t]{\textwidth}
    \centering
    \includegraphics[width=\textwidth]{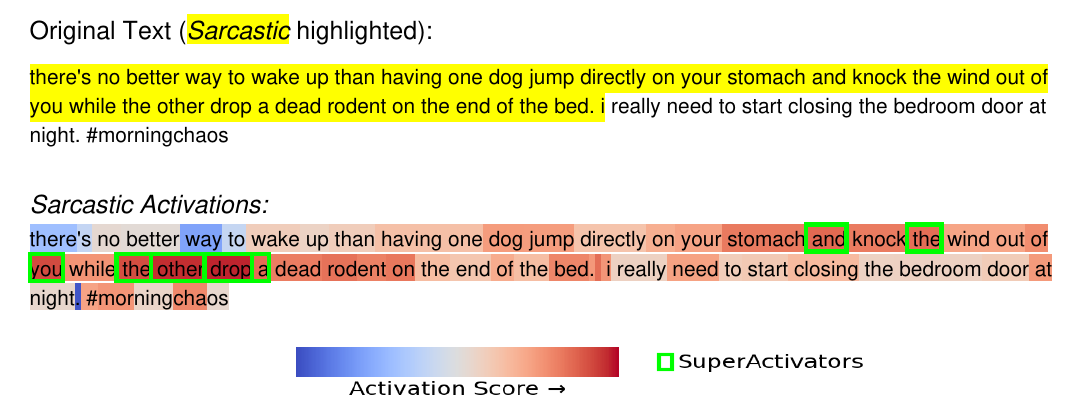}
    \caption{Sarcastic Sample}
  \end{subfigure}
  
  \begin{subfigure}[t]{\textwidth}
    \centering
    \includegraphics[width=\textwidth]{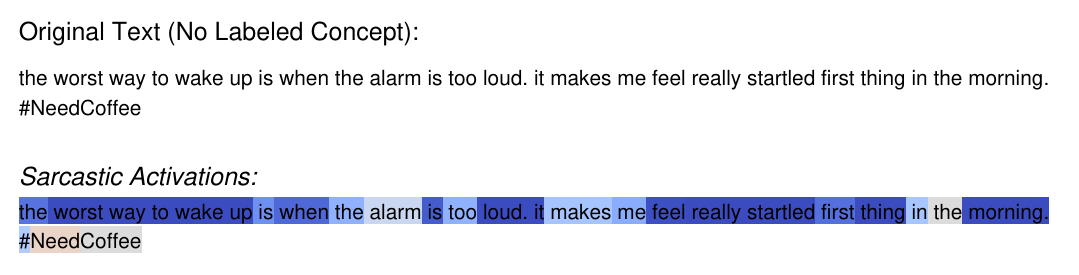}
    \caption{Non-Sarcastic Sample}
  \end{subfigure}

  \caption{\emph{iSarcasm} -- Visualization of Concept Activations and \supers{} (non-sarcastic and sarcastic text samples)}
  \label{fig:superpatch-example-isarcasm}
\end{figure}

\clearpage
\section{Evolution of Activation Separability Across Model Depth}
\label{app:super-motivation}
\begin{figure}[h!]
    \centering

    % Row 1
    \hfill
    \begin{subfigure}{0.49\textwidth}
        \centering
        \includegraphics[width=\linewidth]{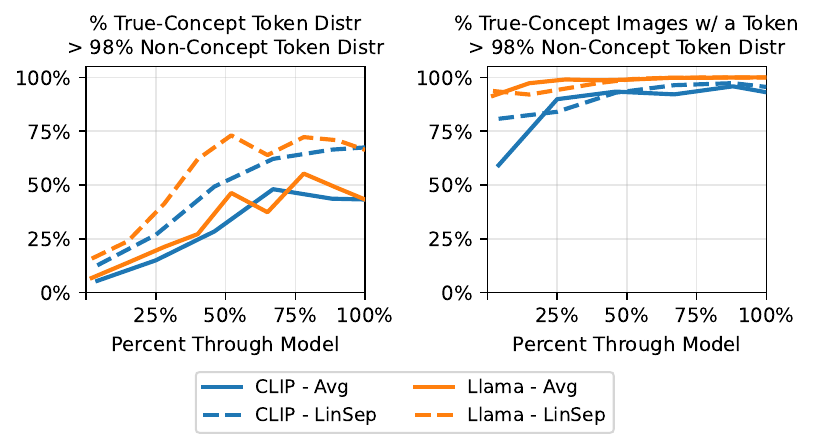}
        \vspace{-1.5em}
        \caption{COCO}
        \label{fig:coco-super-motivations}
    \end{subfigure}
    % \vspace{0.75em}
    \begin{subfigure}{0.49\textwidth}
        \centering
        \includegraphics[width=\linewidth]{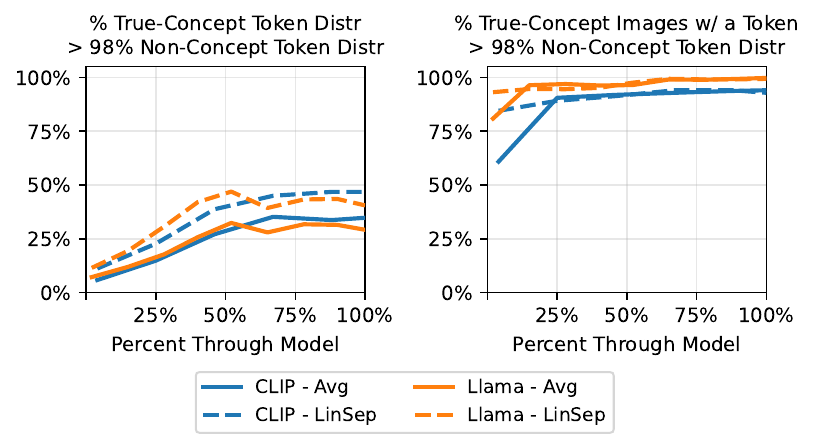}
        \vspace{-1.5em}
        \caption{Broden-OpenSurfaces}
        \label{fig:opensurfaces-super-motivations}
    \end{subfigure}
    
    \begin{subfigure}{0.49\textwidth}
        \centering
        \includegraphics[width=\linewidth]{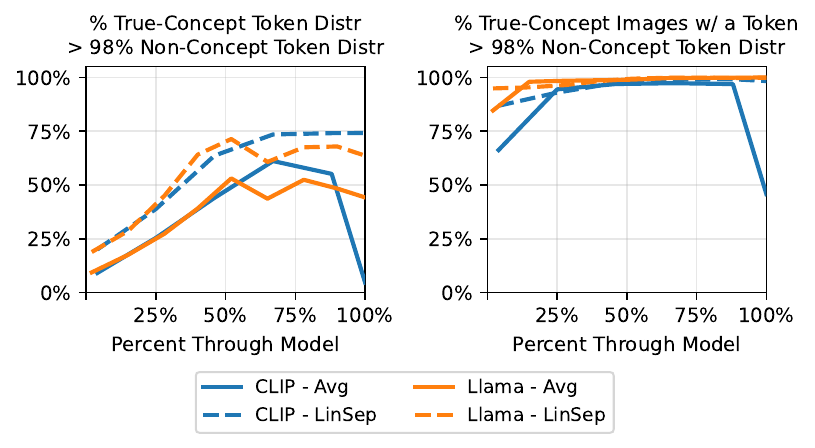}
        \vspace{-1.5em}
        \caption{Broden-Pascal}
        \label{fig:pascal-super-motivations}
    \end{subfigure}
    \hfill
    \begin{subfigure}{0.49\textwidth}
        \centering
        \includegraphics[width=\linewidth]{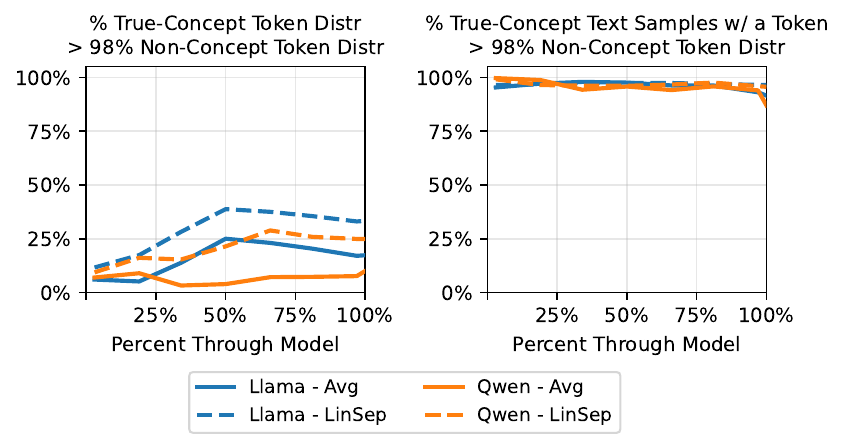}
        \vspace{-1.5em}
        \caption{PERSUADE}
        \label{fig:persuade-super-motivations}
    \end{subfigure}
    \vspace{0.75em}
    \begin{subfigure}{0.49\textwidth}
        \centering
        \includegraphics[width=\linewidth]{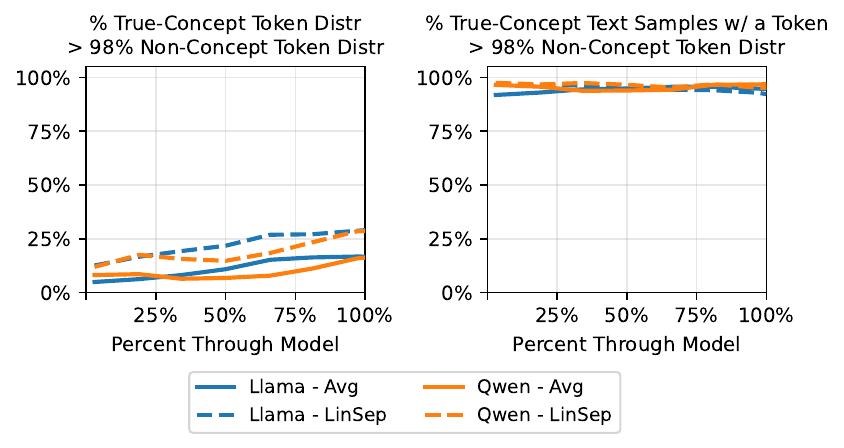}
        \vspace{-1.5em}
        \caption{El Capitan}
        \label{fig:elcapitan-super-motivations}
    \end{subfigure}
    \hfill
    \begin{subfigure}{0.49\textwidth}
        \centering
        \includegraphics[width=\linewidth]{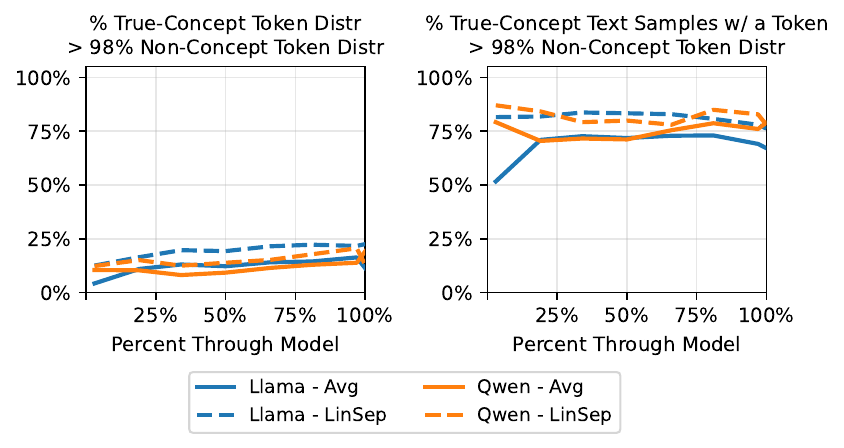}
        \vspace{-1.5em}
        \caption{GoEmotions}
        \label{fig:goemotions-super-motivations}
    \end{subfigure}

    \caption{Across datasets, models, and concept types, high-magnitude in-concept activations become more separable from $D_c^{\mathrm{out}}$ with depth, while maintaining high coverage of in-concept samples.}
    \label{fig:super-motivations-image}
\end{figure}

In this section, we examine how concept activation separability evolves throughout a model. As an initial empirical measure of separability, we consider an in-concept token activation separable from the empirical out-of-concept activation distribution $D_c^{\mathrm{out}}$ if it exceeds a high quantile $q_{0.98}(D_c^{\mathrm{out}})$. For each dataset, the left plots show the fraction of the in-concept activation distribution lying above this threshold as a function of model depth, averaged across concepts. The right plots show the percentage of in-concept samples (images, reviews, essays, etc.) containing at least one such activation, again averaged across concepts. Figure~\ref{fig:super-motivations-image} reports these quantities across various datasets and models, using both average and linear separator concept vectors.

As shown in the left plots, the fraction of $D_c^{\mathrm{in}}$ that is well separated from $D_c^{\mathrm{out}}$ generally increases with model depth, particularly in the early-to-middle layers, although this trend often plateaus in later layers. These gains are especially pronounced on image datasets, while text datasets generally exhibit smaller improvements and lower overall fractions of well-separated activations. The right plots show that, despite this limited token-level separation, a substantial proportion of in-concept samples contain at least one well-separated activation in many settings. 

Together, these results suggest that improvements in concept separability with depth are not uniform across the in-concept activation distribution. Rather, a subset of high-magnitude activations becomes particularly distinguishable from $D_c^{\mathrm{out}}$, and these activations can provide broad sample-level coverage even when much of the distribution remains overlapping. This motivates our focus on the high-activation tail of $D_c^{\mathrm{in}}$ in the analyses that follow.

% In your document:

\clearpage

\section{\texorpdfstring{$D_c^{\text{in}}$ and $D_c^{\text{out}}$ Across Additional Datasets}{Dcin and Dcout Across Datasets}}
\label{app:distributions}

We provide additional examples of the concept activation distributions shown in 
Figure~\ref{fig:acts-across-layers}. Across \emph{COCO}, \emph{Pascal}, \emph{PERSUADE}, \emph{El Capitan}, and \emph{GoEmotions} concepts, 
\emph{Llama-3.2-11B-Vision-Instruct} linear separator concepts show the same qualitative 
pattern: $D_c^{\mathrm{in}}$ and $D_c^{\mathrm{out}}$ become more distinct with depth, with 
the clearest separation concentrated in the high-activation tail of $D_c^{\mathrm{in}}$.

\begin{figure}[H]
    \includegraphics[width=\linewidth]{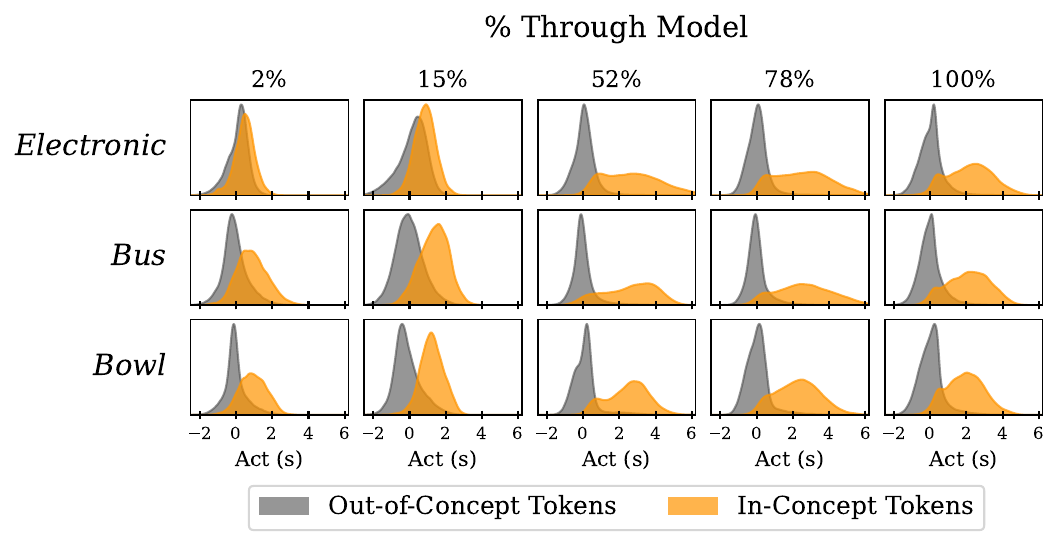}
    \vspace{-1.5em}
    \caption{\emph{COCO}: Concept Activation Distributions for \textit{electronic}, \textit{bus}, and \textit{bowl}.}
    \label{fig:dist-coco}
\end{figure}

% \begin{figure}[H]
%     \noindent\makebox[\linewidth][l]{%
%         \includegraphics[width=.84\linewidth]{Figures/distributions/Llama_Broden-OpenSurfaces_supers_False_activation_distributions_grid.png}
%     }
%     \caption{Broden-OpenSurfaces.}
%     \label{fig:dist-broden-opensurfaces}
% \end{figure}

\begin{figure}[H]
    \includegraphics[width=\linewidth]{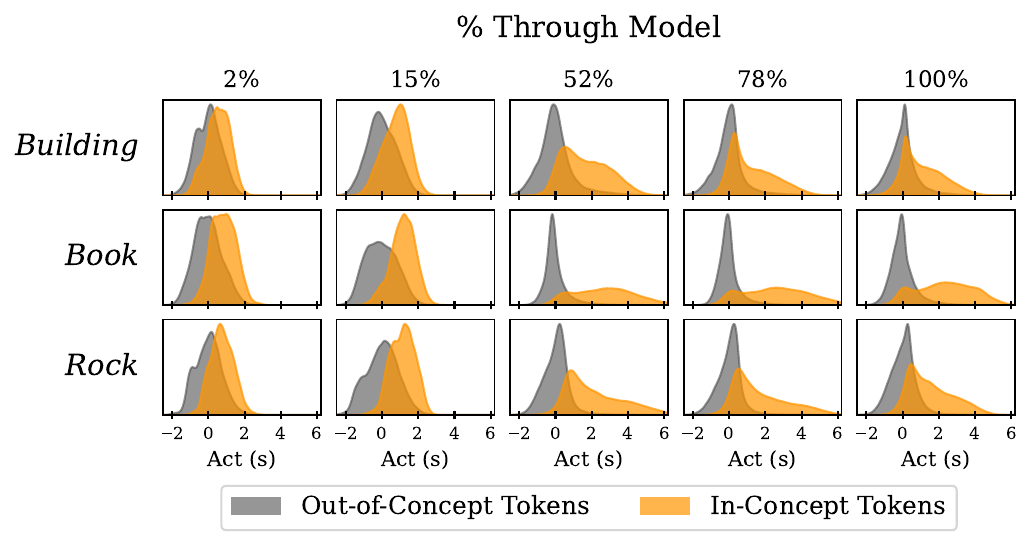}
    \vspace{-1.5em}
    \caption{\emph{Pascal}: Concept Activation Distributions for \textit{building}, \textit{book}, and \textit{rock}.}
    \label{fig:dist-broden-pascal}
\end{figure}

\begin{figure}[H]
    \includegraphics[width=\linewidth]{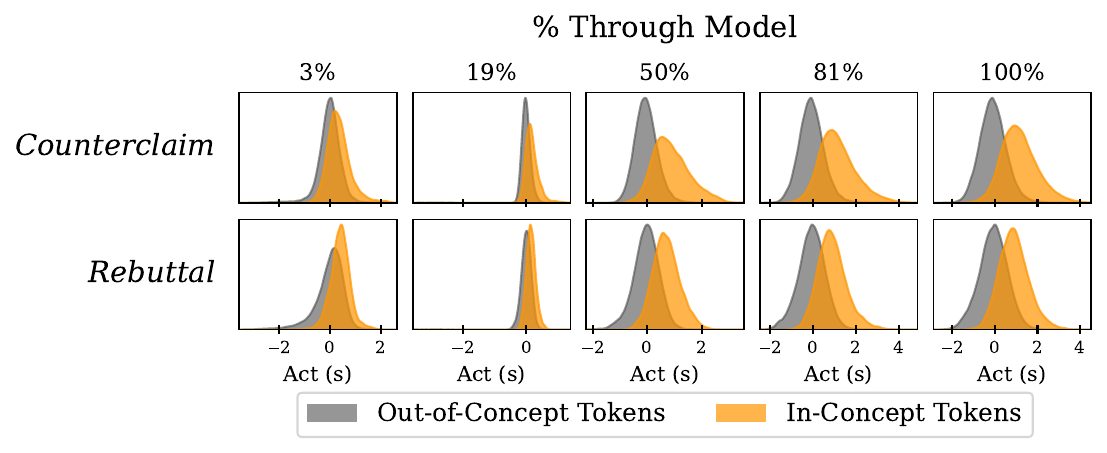}
    \vspace{-1.5em}
    \caption{\emph{PERSUADE}: Concept Activation Distributions for \textit{counterclaim} and \textit{rebuttal}.}
    \label{fig:dist-persuade}
\end{figure}
\vspace{-1em}

\begin{figure}[H]
    \includegraphics[width=\linewidth]{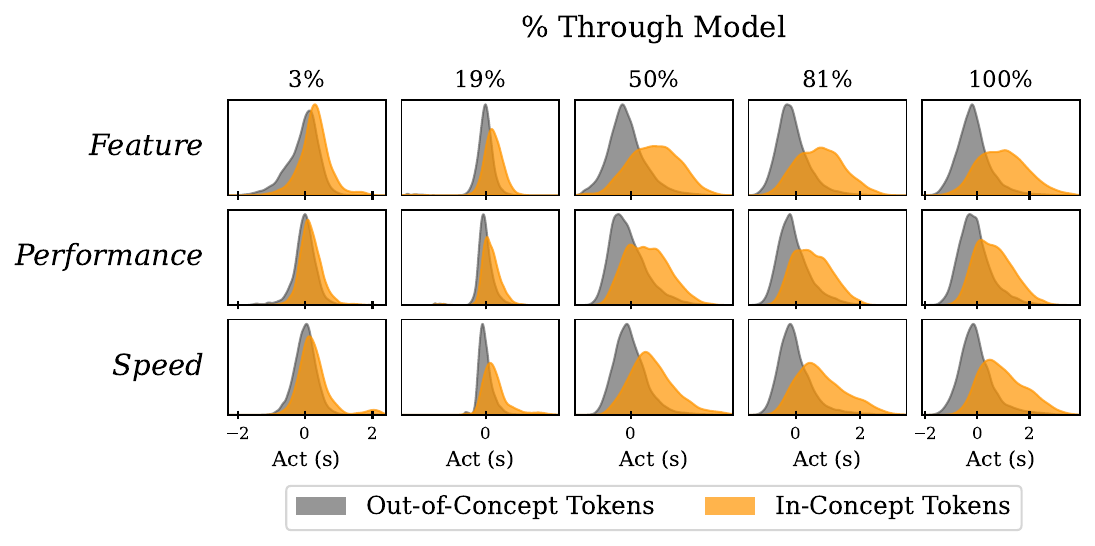}
    \vspace{-1.5em}
    \caption{\emph{El Capitan}: Concept Activation Distributions for \textit{feature}, \textit{performance}, and \textit{speed}.}
    \label{fig:dist-elcapitan}
\end{figure}

\vspace{-1em}

\begin{figure}[H]
    \includegraphics[width=\linewidth]{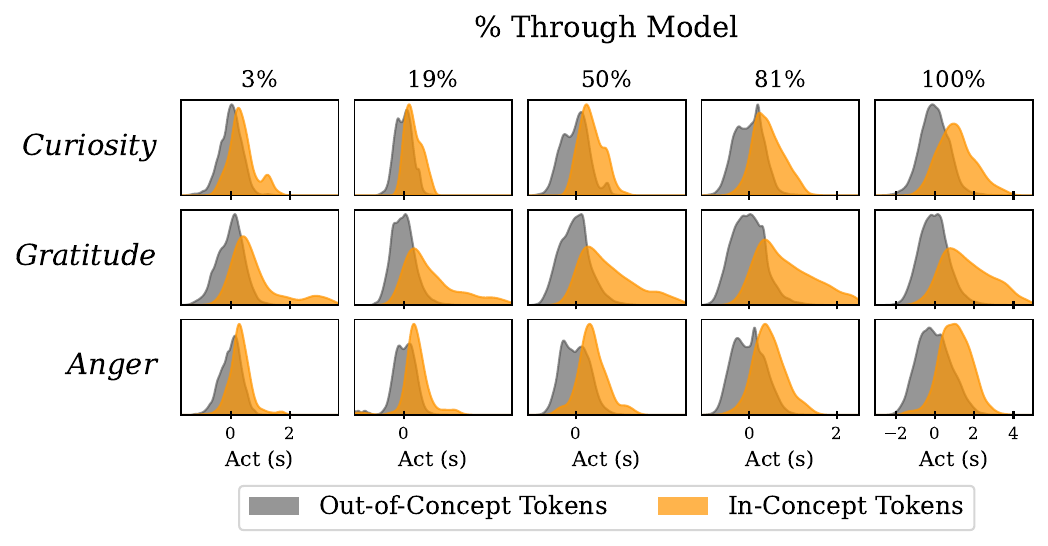}
    \vspace{-1.5em}
    \caption{\emph{GoEmotions}: Concept Activation Distributions for \textit{curiosity}, \textit{gratitude}, and \textit{anger}.}
    \label{fig:dist-goemotions}
\end{figure}

\clearpage

\section{Theoretical Analysis Underlying the \super{} Mechanism}
\label{app:theory}
This appendix provides proofs of the \textit{activation gap amplification} and \textit{tail-asymmetric amplification} theorems from the main text, along with additional analyses characterizing asymptotic equalization and showing that similar dynamics persist under bounded noise when the concept signal is strong enough to overcome non-concept noise.

\subsection{Setup and the scalar reduction}
\label{sec:setup}

Each token $i$ in a sequence of $N$ tokens is represented by $x_i \in \mathbb{R}^d$. A standard self-attention head updates token $i$ as
\[
x_i' = x_i + \sum_{j=1}^N \alpha_{ij} h_j, \qquad \alpha_{ij} = \softmax_j\!\left(\tfrac{q_i^\top k_j}{\sqrt{d_k}}\right), \quad q_i = W_Q x_i,\ k_j = W_K x_j,\ h_j = W_{OV} x_j.
\]

\paragraph{Modeling assumptions.} 
We study concepts that have been learned by the model and, following the linear representation hypothesis \citep{Kim2017InterpretabilityBF}, represent a concept by a unit direction $v\in\mathbb{R}^d$. The concept activation of token $i$ is $s_i=v^\top x_i$.

Building on the read-write decomposition of attention heads \citep{elhage2021mathematical} and prior work identifying semantically specialized heads \citep{Gandelsman2023InterpretingCI}, we formalize a concept-aligned head as one that reads along $v$ through QK and writes back along $v$ through OV. We verify in Appendix~\ref{app:head-validation} that heads with positive alignment along $v$ in both pathways exist for every concept and model we study. Writing $W_{OV}:=W_OW_V$, we decompose the head into concept-aligned and residual components:

\begin{itemize}
  \item[(A1)] \emph{QK alignment.} $\frac{1}{\sqrt{d_k}} W_Q^\top W_K = \gamma\, v v^\top + \varepsilon_{QK}$ with $\gamma > 0$ and $v^\top \varepsilon_{QK} v = 0$. The QK interaction routes attention using the concept direction with strength $\gamma$; all other directions contribute through $\varepsilon_{QK}$.
  \item[(A2)] \emph{OV alignment.} $W_{OV} = \lambda\, v v^\top + \varepsilon_{OV}$ with $\lambda > 0$ and $v^\top \varepsilon_{OV} v = 0$. The OV circuit writes to the concept direction with strength $\lambda$; all other directions contribute through $\varepsilon_{OV}$.
  \item[(A3)] \emph{Non-concept residual.} A per-token residual $\eta_i^{(l)}$ aggregates the contribution of all other layer components (other heads, MLP, residual stream). The full layer update is
  \[
  x_i^{(l+1)} = x_i^{(l)} + \sum_{j=1}^N \alpha_{ij}^{(l)} h_j^{(l)} + \eta_i^{(l)}.
  \]
\end{itemize}

\paragraph{Reduction to scalar dynamics.} Under (A1)--(A3), $h_j = \lambda s_j v + \varepsilon_{OV} x_j$  and the QK logits decompose as $q_i^\top k_j / \sqrt{d_k} = \gamma s_i s_j + \xi_{ij}$. Projecting the attention update onto $v$ collapses the dynamics to
\begin{equation}
s_i^{(l+1)} \;=\; s_i^{(l)} + \lambda \sum_{j=1}^N \alpha_{ij}^{(l)} s_j^{(l)} + \epsilon_i^{(l)}, \qquad \alpha_{ij}^{(l)} := \softmax_j\!\bigl(\gamma s_i^{(l)} s_j^{(l)} + \xi_{ij}^{(l)}\bigr),
\label{eq:scalar-dynamics}
\end{equation}
where $\xi_{ij}^{(l)} := x_i^{(l)\top} \varepsilon_{QK} x_j^{(l)}$ collects the QK perturbation (orthogonal to the concept direction by~(A1)) and $\epsilon_i^{(l)} := v^\top(\varepsilon_{OV} x_i^{(l)} + \eta_i^{(l)})$ aggregates the OV-pathway noise and the non-concept residual contribution from~(A3). We will write the full per-token layer map as
\[
T_i^{(l)}(s) \;:=\; s + \lambda f_i^{(l)}(\gamma s) + \epsilon_i^{(l)}, \qquad f_i^{(l)}(t) := \sum_{j=1}^N s_j^{(l)} \, \softmax_j\!\bigl(t s_j^{(l)} + \xi_{ij}^{(l)}\bigr),
\]
so that $s_i^{(l+1)} = T_i^{(l)}(s_i^{(l)})$.

\subsection{Noiseless regime}
\label{sec:noiseless}

In the noiseless regime, all layer-component perturbations vanish ($\varepsilon_{QK} = 0$, $\varepsilon_{OV} = 0$, $\eta_i^{(l)} = 0$, hence $\xi_{ij}^{(l)} = 0$ and $\epsilon_i^{(l)} = 0$). The row index of $T_i^{(l)}$ enters only through these noise terms, so $T_i^{(l)}$ collapses to a single common map $T^{(l)}(s) = s + \lambda f^{(l)}(\gamma s)$ shared by all tokens.

\begin{itemize}
  \item Theorem~\ref{thm:feedback} (\emph{activation gap amplification}): every pairwise concept activation gap amplifies multiplicatively across layers, $\Delta_{ab}^{(l+1)} = (1 + \lambda\gamma\,\overline{V}_{ab}^{(l)})\,\Delta_{ab}^{(l)}$ where $\overline{V}_{ab}^{(l)} > 0$ is the attention-weighted variance averaged over the gap; the overall ordering of tokens is preserved.
  \begin{itemize}
    \item Corollary~\ref{cor:concentration} (\emph{attention concentration on the poles}): the resulting growth in pole margins pulls each nonzero-activation token's attention exponentially toward the maximally and minimally activated tokens (the \emph{poles}) with each layer.
    \item Corollary~\ref{cor:equalization} (\emph{within-tail equalization}): same-tail activations equalize at asymptotic geometric rate $(1+\lambda)^{-1}$.
  \end{itemize}
  \item Theorem~\ref{thm:skew} (\emph{tail-asymmetric amplification}): the curvature of the layer-update map satisfies $\bigl(T^{(l)}\bigr)''(s) = \lambda\gamma^2 \cdot \E_{\alpha^{(l)}(\gamma s)}[(s^{(l)} - \E_{\alpha^{(l)}(\gamma s)} s^{(l)})^3]$, i.e., $\lambda\gamma^2$ times the skew (third central moment) of the attention-weighted activation distribution. When this distribution is positively skewed, $\bigl(T^{(l)}\bigr)'$ is locally increasing in $s$, so gaps in the heavier tail amplify at strictly faster rates than gaps in the bulk.
\end{itemize}

\subsubsection{Activation gap amplification and pole concentration}

We first record a softmax identity that underlies every result that follows.

\begin{lemma}[Softmax weighted-average monotonicity]
\label{lem:softmax-mono}
Let $s_1, \ldots, s_N \in \mathbb{R}$ with $s_a \neq s_b$ for some $a, b$. Define
\[
f(t) \;=\; \sum_{j=1}^N \frac{s_j\, e^{t s_j}}{\sum_{k=1}^N e^{t s_k}}.
\]
Then $f$ is strictly increasing on $\mathbb{R}$, and
\[
f'(t) \;=\; \Var_{\alpha(t)}\!\left[s\right] \;>\; 0,
\]
where $\alpha_j(t) = e^{t s_j} / \sum_k e^{t s_k}$.
\end{lemma}

\begin{proof}
With $\alpha_j(t) = e^{t s_j}/\sum_k e^{t s_k}$, direct differentiation gives $\alpha_j'(t) = \alpha_j(t)\,(s_j - f(t))$. Then
\[
f'(t) \;=\; \sum_j \alpha_j'(t)\, s_j \;=\; \sum_j \alpha_j(t)\,(s_j - f(t))\, s_j \;=\; \sum_j \alpha_j(t)\, s_j^2 - f(t)^2 \;=\; \Var_{\alpha(t)}[s].
\]
The variance is strictly positive whenever the $s_j$ are not all equal. (Equivalently, $f(t) = \frac{d}{dt} \log \sum_j e^{t s_j}$ is the derivative of a strictly convex function.)
\end{proof}

\begin{theorem}[Activation gap amplification]
\label{thm:feedback}
Suppose the activations $s_1^{(l)}, \ldots, s_N^{(l)}$ at layer $l$ are not all equal, and write $\Delta_{ab}^{(l)} := s_a^{(l)} - s_b^{(l)}$ for the gap between tokens $a, b$ at layer $l$. Then under \eqref{eq:scalar-dynamics}, for any pair with $s_a^{(l)} > s_b^{(l)}$,
\[
\Delta_{ab}^{(l+1)} \;=\; \bigl(1 + \lambda\gamma\,\overline{V}_{ab}^{(l)}\bigr)\,\Delta_{ab}^{(l)},
\]
where
\[
\overline{V}_{ab}^{(l)} \;:=\; \int_0^1 \Var_{\alpha^{(l)}(\gamma(s_b^{(l)} + u\Delta_{ab}^{(l)}))}\!\bigl[s^{(l)}\bigr]\, du.
\]
Since $\overline{V}_{ab}^{(l)} > 0$ (Lemma~\ref{lem:softmax-mono}), every pairwise gap amplifies multiplicatively across layers, and the overall ordering of tokens is preserved.
\end{theorem}

The magnitude of $\overline{V}_{ab}^{(l)}$ reflects the spread of the attention-weighted distribution: a wider weighting gives faster gap growth, a more concentrated one slower.

\begin{proof}
At layer $l$, the activations $s_1^{(l)}, \ldots, s_N^{(l)}$ are fixed, so only the query varies in $T^{(l)}(s) = s + \lambda f^{(l)}(\gamma s)$, and $s_q^{(l+1)} = T^{(l)}(s_q^{(l)})$ for $q \in \{a, b\}$. By the chain rule and Lemma~\ref{lem:softmax-mono},
\[
\bigl(T^{(l)}\bigr)'(s) \;=\; 1 + \lambda\gamma\,(f^{(l)})'(\gamma s) \;=\; 1 + \lambda\gamma\,\Var_{\alpha^{(l)}(\gamma s)}\!\bigl[s^{(l)}\bigr].
\]
By the fundamental theorem of calculus, $\Delta_{ab}^{(l+1)} = T^{(l)}(s_a^{(l)}) - T^{(l)}(s_b^{(l)}) = \int_{s_b^{(l)}}^{s_a^{(l)}} \bigl(T^{(l)}\bigr)'(t)\,dt$. Substituting $t = s_b^{(l)} + u\Delta_{ab}^{(l)}$ (so $dt = \Delta_{ab}^{(l)}\, du$) and the expression for $\bigl(T^{(l)}\bigr)'$ above,
\[
\Delta_{ab}^{(l+1)} \;=\; \Delta_{ab}^{(l)}\int_0^1 \bigl(T^{(l)}\bigr)'\!\bigl(s_b^{(l)} + u\Delta_{ab}^{(l)}\bigr)\,du \;=\; \bigl(1 + \lambda\gamma\,\overline{V}_{ab}^{(l)}\bigr)\Delta_{ab}^{(l)}.
\]
The variance is strictly positive at every $t$ when activations are not all equal, so $\overline{V}_{ab}^{(l)} > 0$ and the gap strictly amplifies; pairwise distinct activations therefore stay distinct, preserving the ordering.
\end{proof}

Two consequences of Theorem~\ref{thm:feedback} follow.

\begin{corollary}[Attention concentration on the poles]
\label{cor:concentration}
Let $A^{(l)} := \arg\max_j s_j^{(l)}$ and $B^{(l)} := \arg\min_j s_j^{(l)}$ denote the maximally and minimally activated tokens at layer $l$ (ties resolved by any fixed rule), with pole margins
\[
\delta_A^{(l)} := \min_{k \neq A^{(l)}}\bigl(s_{A^{(l)}}^{(l)} - s_k^{(l)}\bigr), \qquad \delta_B^{(l)} := \min_{k \neq B^{(l)}}\bigl(s_k^{(l)} - s_{B^{(l)}}^{(l)}\bigr).
\]
For any query token $i$, the total attention placed on non-pole tokens is exponentially bounded:
\[
1 - \bigl(\alpha_{iA^{(l)}}^{(l)} + \alpha_{iB^{(l)}}^{(l)}\bigr) \;\leq\; (N-2)\exp\!\bigl(-\gamma\, |s_i^{(l)}|\, \delta_E^{(l)}\bigr), \qquad E^{(l)} := \begin{cases} A^{(l)} & \text{if } s_i^{(l)} > 0, \\ B^{(l)} & \text{if } s_i^{(l)} < 0. \end{cases}
\]
\end{corollary}

Each pole margin is strictly increasing in $l$ by Theorem~\ref{thm:feedback}, so every nonzero-activation token has its attention weight pulled toward the two poles exponentially with each layer. (Degenerate case: if $s_i^{(l)} = 0$, the query has no concept preference and attention is uniform across all tokens, so the bound is vacuous.)

\begin{proof}
Choose $E^{(l)}$ so that the logit $\gamma s_i^{(l)} s_{E^{(l)}}^{(l)}$ is largest among $j$, i.e., $E^{(l)} = A^{(l)}$ when $s_i^{(l)} > 0$ and $E^{(l)} = B^{(l)}$ when $s_i^{(l)} < 0$. Bounding the softmax denominator from below by this largest term,
\[
\alpha_{ik}^{(l)} \;=\; \frac{e^{\gamma s_i^{(l)} s_k^{(l)}}}{\sum_j e^{\gamma s_i^{(l)} s_j^{(l)}}} \;\leq\; \frac{e^{\gamma s_i^{(l)} s_k^{(l)}}}{e^{\gamma s_i^{(l)} s_{E^{(l)}}^{(l)}}} \;=\; \exp\!\bigl(-\gamma\, s_i^{(l)}\,(s_{E^{(l)}}^{(l)} - s_k^{(l)})\bigr).
\]

When $s_i^{(l)} > 0$, the gap $s_{A^{(l)}}^{(l)} - s_k^{(l)} \geq \delta_A^{(l)}$ for $k \neq A^{(l)}$ gives $\alpha_{ik}^{(l)} \leq \exp(-\gamma\, s_i^{(l)}\, \delta_A^{(l)})$. When $s_i^{(l)} < 0$, write $\gamma s_i^{(l)} = -\gamma|s_i^{(l)}|$ and apply the same argument using $s_k^{(l)} - s_{B^{(l)}}^{(l)} \geq \delta_B^{(l)}$ for $k \neq B^{(l)}$. In both cases,
\[
\alpha_{ik}^{(l)} \;\leq\; \exp\!\bigl(-\gamma|s_i^{(l)}|\delta_E^{(l)}\bigr) \quad \text{for any } k \neq E^{(l)},
\]
and summing over the $N-2$ indices $k \neq A^{(l)}, B^{(l)}$ gives the displayed bound.

For margin monotonicity: for each $k \neq A^{(l)}$, $\Delta_{A,k}^{(l+1)} > \Delta_{A,k}^{(l)} \geq \delta_A^{(l)}$ by Theorem~\ref{thm:feedback}, and ordering preservation gives $A^{(l+1)} = A^{(l)}$, so $\delta_A^{(l+1)} = \min_{k \neq A^{(l)}}\Delta_{A,k}^{(l+1)} > \delta_A^{(l)}$ (analogously for $\delta_B^{(l)}$).
\end{proof}

\begin{corollary}[Within-tail equalization]
\label{cor:equalization}
Same-tail activations equalize: ratios $s_i^{(l)}/s_E^{(l)}$ converge to $1$ at asymptotic geometric rate $(1+\lambda)^{-1}$.

Quantitatively, fix a token $i$ on the same side as a pole $E^{(l)}$ (so $s_i^{(l)}$ and $s_{E^{(l)}}^{(l)}$ share a sign), and let $M_q^{(l)} := f^{(l)}(\gamma s_q^{(l)}) = \sum_j \alpha_{qj}^{(l)} s_j^{(l)}$ denote the attention-weighted average for token $q$. Define the deviation from equality and the relative deviation of $M_q$ from the pole,
\[
u_{i,E}^{(l)} \;:=\; 1 - \frac{s_i^{(l)}}{s_E^{(l)}}, \qquad \varepsilon_{q,E}^{(l)} \;:=\; \frac{M_q^{(l)} - s_E^{(l)}}{s_E^{(l)}} \text{ for } q \in \{i, E\},
\]
and set $\eta_l := \max\{|\varepsilon_{i,E}^{(l)}|, |\varepsilon_{E,E}^{(l)}|\}$. Then whenever $1 + \lambda - \lambda \eta_l > 0$,
\[
\bigl|u_{i,E}^{(l+1)}\bigr| \;\leq\; \frac{|u_{i,E}^{(l)}| + 2\lambda \eta_l}{1 + \lambda - \lambda \eta_l}.
\]
The contraction rate is $(1+\lambda)^{-1} + O(\eta_l)$, and $\eta_l \to 0$ as attention concentrates on the poles (Corollary~\ref{cor:concentration}).
\end{corollary}

Once attention concentrates on the poles (Corollary~\ref{cor:concentration}), every token within a tail receives essentially the same update $\lambda s_E^{(l)}$: its weighted average $M_q^{(l)}$ converges to $s_E^{(l)}$ as off-pole attention vanishes. At large depth, token activations within a tail therefore grow by approximately the same factor $(1+\lambda)$ per layer; their magnitudes grow without bound, while their ratios converge to $1$.

\begin{proof}
Plugging $s_q^{(l+1)} = s_q^{(l)} + \lambda s_E^{(l)}(1 + \varepsilon_{q,E}^{(l)})$ for $q \in \{i, E\}$ into $1 - s_i^{(l+1)}/s_E^{(l+1)}$ and rearranging,
\[
u_{i,E}^{(l+1)} \;=\; \frac{u_{i,E}^{(l)} + \lambda\bigl(\varepsilon_{E,E}^{(l)} - \varepsilon_{i,E}^{(l)}\bigr)}{1 + \lambda + \lambda \varepsilon_{E,E}^{(l)}}.
\]
The displayed bound follows by the triangle inequality applied separately to numerator and denominator.

For the rate, assume $s_E^{(l)} > 0$ (the negative case is symmetric); for $q \in \{i, E\}$,
\[
|M_q^{(l)} - s_E^{(l)}| \;=\; \Bigl|\sum_{k \neq E}\alpha_{qk}^{(l)}(s_E^{(l)} - s_k^{(l)})\Bigr| \;\leq\; (1 - \alpha_{qE}^{(l)})\,(s_E^{(l)} - s_B^{(l)}),
\]
so $|\varepsilon_{q,E}^{(l)}| \leq (1 - \alpha_{qE}^{(l)})\cdot (s_E^{(l)} - s_B^{(l)})/s_E^{(l)}$. Corollary~\ref{cor:concentration} gives $1 - \alpha_{qE}^{(l)} \to 0$ exponentially as the pole margins grow, while the ratio $(s_E^{(l)} - s_B^{(l)})/s_E^{(l)}$ remains bounded since both poles grow at the same geometric rate $1+\lambda$ once attention concentrates. Hence $\eta_l \to 0$, and the leading rate $(1+\lambda)^{-1}$ governs equalization.
\end{proof}

\subsubsection{Tail-asymmetric amplification}

Theorem~\ref{thm:feedback} establishes that activation gaps amplify multiplicatively at every layer, with per-layer rate $\bigl(T^{(l)}\bigr)'(s) = 1 + \lambda\gamma\,\Var_{\alpha^{(l)}(\gamma s)}[s^{(l)}] > 1$, but says nothing about how this rate varies with $s$ --- i.e., about the curvature $\bigl(T^{(l)}\bigr)''$. The next theorem shows that this curvature is exactly $\lambda\gamma^2$ times the skew of the attention-weighted activation distribution, exposing the mechanism that turns an initial positive skew into asymmetric tail expansion at depth.

\begin{theorem}[Tail-asymmetric amplification]
\label{thm:skew}
For every $s \in \mathbb{R}$ and every layer $l$ for which the activations are not all equal,
\[
\bigl(T^{(l)}\bigr)''(s) \;=\; \lambda \gamma^2 \cdot \E_{\alpha^{(l)}(\gamma s)}\!\Bigl[\bigl(s^{(l)} - \E_{\alpha^{(l)}(\gamma s)}[s^{(l)}]\bigr)^3\Bigr].
\]
That is, the curvature of the layer update at any point equals $\lambda\gamma^2$ times the skew of the activations under the attention distribution at that point, and inherits its sign.
\end{theorem}

\begin{proof}
Abbreviate $\alpha = \alpha^{(l)}$, $f = f^{(l)}$, $s = s^{(l)}$, with expectations under $\alpha(t)$. We compute $f''(t)$ and apply the chain rule, using $\bigl(T^{(l)}\bigr)''(s) = \lambda\gamma^2 f''(\gamma s)$ as in the proof of Theorem~\ref{thm:feedback}. From the proof of Lemma~\ref{lem:softmax-mono}, $\alpha_j'(t) = \alpha_j(t)(s_j - f(t))$ and $f'(t) = \E[s^2] - f(t)^2$. Differentiating the latter,
\[
\frac{d}{dt} \E[s^2] \;=\; \sum_j \alpha_j'(t)\, s_j^2 \;=\; \E[s^3] - f(t)\E[s^2],
\]
\[
\frac{d}{dt} f(t)^2 \;=\; 2 f(t)\bigl(\E[s^2] - f(t)^2\bigr),
\]
so that
\[
f''(t) \;=\; \E[s^3] - 3 f(t) \E[s^2] + 2 f(t)^3.
\]
Expanding the third central moment and using $\E[s] = f(t)$,
\[
\begin{aligned}
\E\!\bigl[(s - f(t))^3\bigr]
&= \E[s^3] - 3 f(t) \E[s^2]
   + 3 f(t)^2 \E[s] - f(t)^3 \\
&= \E[s^3] - 3 f(t) \E[s^2] + 2 f(t)^3 \\
&= f''(t),
\end{aligned}
\]
so $f''(t)$ equals the attention-weighted third central moment under $\alpha(t)$. Setting $t = \gamma s$ gives the stated formula.
\end{proof}

This explains the empirical observation (Figure~\ref{fig:acts-across-layers}) that the initial positive skew of $\mathcal{D}_c^{in}$ sharpens with depth: wherever the attention-weighted distribution is positively skewed, $\bigl(T^{(l)}\bigr)''(s) > 0$, so $\bigl(T^{(l)}\bigr)'(s)$ is strictly increasing in $s$. Since the gap amplification rate $1 + \lambda\gamma\overline{V}_{ab}^{(l)}$ from Theorem~\ref{thm:feedback} is exactly $\bigl(T^{(l)}\bigr)'$ averaged over the gap interval $[s_b^{(l)}, s_a^{(l)}]$, gaps in the upper tail amplify at strictly higher rates than gaps in the bulk, producing the SuperActivator tail of $\mathcal{D}_c^{in}$ that pulls away while the bulk grows more slowly and remains entangled with $\mathcal{D}_c^{out}$.

This is a finite-depth phenomenon. The mechanism amplifies any existing skew in the activation distribution rather than creating a tail from scratch: positive skew gets sharpened with depth, with the upper tail pulling away from the bulk at a strictly higher rate than the bulk gaps grow. At sufficient depth, Corollary~\ref{cor:equalization} eventually pulls every same-sign activation toward its corresponding pole, so within-tail relative differences disappear in the asymptotic limit. The empirical regime corresponds to the transient where skew amplification dominates: bulk tokens with small $|s|$ have not yet entered the asymptotic equalization regime because their attention concentration is slow --- the pole-margin product $|s_i^{(l)}|\delta_E^{(l)}$ in Corollary~\ref{cor:concentration} is small at small $|s_i^{(l)}|$ --- so the bulk lags behind the sharpening tail.

\subsection{Bounded-noise regime}
\label{sec:noisy}

We now ask how each of the noiseless results changes when the perturbations $\xi_{ij}^{(l)}$ and $\epsilon_i^{(l)}$ from~\eqref{eq:scalar-dynamics} are bounded but nonzero:
\begin{itemize}
\item[(N1)] $|\xi_{ij}^{(l)}| \leq \Xi$ and $|\epsilon_i^{(l)}| \leq \mathcal{E}$ for all $i, j, l$.
\end{itemize}
The unifying picture: \emph{each unconditional noiseless result becomes a result that holds either above a signal-to-noise threshold on the activation gaps or with extra depth required for the dynamics to climb past a noise floor}. Concretely:
\begin{itemize}
  \item Theorem~\ref{thm:feedback-noise} (\emph{gaps amplify multiplicatively, but only above a signal-to-noise threshold}): the multiplicative rate persists, plus an additive residual scaling with the global activation range $\Delta_{AB}^{(l)}$. Pairs above the signal-to-noise threshold amplify at the noiseless rate every layer; pairs below it never reliably amplify.
  \begin{itemize}
    \item Corollary~\ref{cor:concentration-noise} (\emph{attention concentrates on the poles, but it takes more layers}): the exponential rate is unchanged; an additive $+2\Xi$ in the exponent sets a noise floor that the signal must climb above before concentration becomes detectable.
    \item Corollary~\ref{cor:equalization-noise} (\emph{within-tail activations equalize, but more slowly}): the asymptotic contraction rate is the noiseless $(1+\lambda)^{-1}$, but equalization is delayed by the residual off-pole attention and converges only to a noise-floor neighborhood of equality.
  \end{itemize}
  \item Corollary~\ref{cor:skew-noise} (\emph{tail-asymmetric amplification holds verbatim under noisy attention}): the same identity holds, with the skew now taken under the row-$i$ noisy attention rather than the shared noiseless one; positive skew of the noisy distribution still produces preferential upper-tail amplification.
\end{itemize}

The gap-amplification proof uses the following softmax stability bound, which controls how much two softmax distributions can differ when their logits differ entrywise.

\begin{lemma}[Softmax stability under bounded logit perturbations]
\label{lem:softmax-stability}
Let $p_j = e^{\ell_j}/\sum_k e^{\ell_k}$ and $q_j = e^{\ell'_j}/\sum_k e^{\ell'_k}$ be two softmax distributions over $N$ items whose logits differ entrywise by at most $\Lambda$, i.e., $|\ell_j - \ell'_j| \leq \Lambda$ for all $j$. Then
\[
\frac{p_j}{q_j} \in [e^{-2\Lambda}, e^{2\Lambda}] \text{ for every } j, \qquad \sum_{j=1}^N |p_j - q_j| \leq e^{2\Lambda} - 1.
\]
\end{lemma}

\begin{proof}
Write $p_j/q_j = e^{\ell_j - \ell'_j} \cdot Z'/Z$ where $Z = \sum_k e^{\ell_k}$ and $Z' = \sum_k e^{\ell'_k}$. The per-token factor lies in $[e^{-\Lambda}, e^{\Lambda}]$ by hypothesis. For the partition functions, the entrywise bound $e^{\ell_k} \in [e^{-\Lambda}e^{\ell'_k}, e^{\Lambda}e^{\ell'_k}]$ gives $Z/Z' \in [e^{-\Lambda}, e^{\Lambda}]$, hence $Z'/Z \in [e^{-\Lambda}, e^{\Lambda}]$. Multiplying, $p_j/q_j \in [e^{-2\Lambda}, e^{2\Lambda}]$. Since $|p_j/q_j - 1| \leq e^{2\Lambda} - 1$, we have $|p_j - q_j| \leq (e^{2\Lambda} - 1) q_j$, and summing over $j$ gives $\sum_j |p_j - q_j| \leq e^{2\Lambda} - 1$.
\end{proof}

\begin{theorem}[Gaps amplify multiplicatively, but only above a signal-to-noise threshold]
\label{thm:feedback-noise}
Under~(N1), for any pair $a, b$ with $s_a^{(l)} > s_b^{(l)}$,
\[
\Delta_{ab}^{(l+1)} \;=\; \bigl(1 + \lambda\gamma\overline V_{ab}^{(l)}\bigr)\Delta_{ab}^{(l)} + R_{ab}^{(l)}, \qquad |R_{ab}^{(l)}| \leq \tfrac{1}{2}\lambda(e^{4\Xi}-1)\Delta_{AB}^{(l)} + 2\mathcal{E}.
\]
Here $R_{ab}^{(l)}$ is the residual contributed by the noise, added to the noiseless multiplicative update, with $\Delta_{AB}^{(l)} := \max_j s_j^{(l)} - \min_j s_j^{(l)}$, and
\[
\overline V_{ab}^{(l)} \;:=\; \int_0^1 \Var_{\beta_a(\gamma(s_b^{(l)} + u\Delta_{ab}^{(l)}))}[s^{(l)}]\, du, \qquad \beta_{a,j}(t) := \softmax_j(t s_j^{(l)} + \xi_{aj}^{(l)}),
\]
the noiseless quantity from Theorem~\ref{thm:feedback} with the noisy attention from query $a$ in place of the noiseless attention.
\end{theorem}

Multiplicative gap growth holds only when the gap $\Delta_{ab}^{(l)}$ is a substantial fraction of the activation spread $\Delta_{AB}^{(l)}$ --- that is, when the two tokens are already meaningfully separated relative to the rest of the population. The QK perturbation bound $\Xi$ enters exponentially through $(e^{4\Xi}-1)$, so small QK noise contributes negligibly but moderate QK noise can prevent any pair from amplifying. The concept-axis noise $\epsilon_i^{(l)}$ (OV-pathway perturbations and the aggregate of other heads, MLP, residual stream) contributes a fixed $2\mathcal{E}$ that does not depend on activation spread; whether it matters comes down to $\mathcal{E}/(\lambda\gamma)$, since strong concept-direction signal makes the additive jitter relatively negligible. Larger local variance $\overline V_{ab}^{(l)}$ over the gap interval also contributes to multiplicative gap growth.

The same threshold condition determines order preservation: the order between $a$ and $b$ is guaranteed only if $\Delta_{ab}^{(l)}$ exceeds the residual, which scales with the global spread $\Delta_{AB}^{(l)}$. Pairs with $\Delta_{ab} \sim \Delta_{AB}$ keep their order; pairs with $\Delta_{ab} \ll \Delta_{AB}$ may reshuffle.

\begin{proof}
By~\eqref{eq:scalar-dynamics}, the noisy gap update is
\[
\Delta_{ab}^{(l+1)} - \Delta_{ab}^{(l)} \;=\; \lambda\bigl(f_a^{(l)}(\gamma s_a^{(l)}) - f_b^{(l)}(\gamma s_b^{(l)})\bigr) + (\epsilon_a^{(l)} - \epsilon_b^{(l)}),
\]
with $f_q^{(l)}$ as defined in the setup. Unlike the noiseless case where $f_i^{(l)}$ collapses to a single function $f^{(l)}$ shared by all queries, here $f_q^{(l)}$ depends on the query $q$ through $\xi_{qj}^{(l)}$, so the difference cannot be handled by FTC directly. Insert an intermediate at $\gamma s_b^{(l)}$:
\[
f_a^{(l)}(\gamma s_a^{(l)}) - f_b^{(l)}(\gamma s_b^{(l)}) \;=\; \underbrace{f_a^{(l)}(\gamma s_a^{(l)}) - f_a^{(l)}(\gamma s_b^{(l)})}_{\text{(I): same query, two arguments}} \;+\; \underbrace{f_a^{(l)}(\gamma s_b^{(l)}) - f_b^{(l)}(\gamma s_b^{(l)})}_{\text{(II): two queries, same argument}}.
\]
Term (I) is the noiseless gap-amplification calculation applied to $f_a^{(l)}$ in place of $f^{(l)}$: chain rule gives $(f_a^{(l)})'(t) = \Var_{\beta_a(t)}[s^{(l)}]$, and FTC + change of variables (as in the proof of Theorem~\ref{thm:feedback}) yield $\text{(I)} = \gamma\,\overline V_{ab}^{(l)}\,\Delta_{ab}^{(l)}$.

Term (II) is the new noise contribution. Let $p_j := \softmax_j(\gamma s_b^{(l)} s_j^{(l)} + \xi_{aj}^{(l)})$ and $q_j := \softmax_j(\gamma s_b^{(l)} s_j^{(l)} + \xi_{bj}^{(l)})$. Since $\sum_j(p_j - q_j) = 0$, subtracting the midpoint $c := (s_A^{(l)} + s_B^{(l)})/2$ inside the sum and applying the triangle inequality,
\[
|\text{(II)}| \;=\; \Bigl|\sum_j (s_j^{(l)} - c)(p_j - q_j)\Bigr| \;\leq\; \max_j |s_j^{(l)} - c| \cdot \sum_j |p_j - q_j| \;\leq\; \tfrac{1}{2}\Delta_{AB}^{(l)} \sum_j |p_j - q_j|.
\]
The logits of $p, q$ differ entrywise by at most $|\xi_{aj}^{(l)} - \xi_{bj}^{(l)}| \leq 2\Xi$, so Lemma~\ref{lem:softmax-stability} with $\Lambda = 2\Xi$ gives $\sum_j |p_j - q_j| \leq e^{4\Xi} - 1$, hence $|\text{(II)}| \leq \tfrac{1}{2}(e^{4\Xi} - 1)\Delta_{AB}^{(l)}$. Combining with $|\epsilon_a^{(l)} - \epsilon_b^{(l)}| \leq 2\mathcal{E}$ from (N1) gives the stated bound.
\end{proof}

\begin{corollary}[Attention concentrates on the poles, but it takes more layers]
\label{cor:concentration-noise}
Under~(N1), for every token $i$, with $A^{(l)}, B^{(l)}, \delta_E^{(l)}$ as in Corollary~\ref{cor:concentration},
\[
1 - \bigl(\alpha_{iA^{(l)}}^{(l)} + \alpha_{iB^{(l)}}^{(l)}\bigr) \;\leq\; (N-2)\exp\!\bigl(-\gamma|s_i^{(l)}|\delta_E^{(l)} + 2\Xi\bigr).
\]
\end{corollary}

Below the threshold $\gamma|s_i|\delta_E < 2\Xi$ the right-hand side exceeds $1$ and the bound carries no information. Above it, the noisy bound is the noiseless bound from Corollary~\ref{cor:concentration} times $e^{2\Xi}$, so if the signal $|s_i|\delta_E$ clears the $2\Xi$ noise from the QK perturbations, attention concentrates on the poles at the noiseless exponential rate. The noise adds an extra $2\Xi/\gamma$ depth delay on top of the noiseless concentration case.

\begin{proof}
In the noisy case, the attention is $\alpha_{ik}^{(l)} = \softmax_j(\gamma s_i^{(l)} s_k^{(l)} + \xi_{ik}^{(l)})$ --- noiseless logit plus perturbation. As in the proof of Corollary~\ref{cor:concentration}, choose $E^{(l)}$ so that the noiseless logit $\gamma s_i^{(l)} s_{E^{(l)}}^{(l)}$ is largest, and bound the softmax denominator from below by this term:
\[
\alpha_{ik}^{(l)} \;\leq\; \frac{\exp(\gamma s_i^{(l)} s_k^{(l)} + \xi_{ik}^{(l)})}{\exp(\gamma s_i^{(l)} s_{E^{(l)}}^{(l)} + \xi_{iE^{(l)}}^{(l)})} \;=\; \exp\!\bigl(-\gamma s_i^{(l)}(s_{E^{(l)}}^{(l)} - s_k^{(l)}) + (\xi_{ik}^{(l)} - \xi_{iE^{(l)}}^{(l)})\bigr).
\]
The noiseless sign argument gives $s_i^{(l)}(s_{E^{(l)}}^{(l)} - s_k^{(l)}) \geq |s_i^{(l)}|\delta_E^{(l)}$. The new step is bounding the QK perturbation: by~(N1), $|\xi_{ik}^{(l)} - \xi_{iE^{(l)}}^{(l)}| \leq 2\Xi$. Combining, $\alpha_{ik}^{(l)} \leq \exp(-\gamma|s_i^{(l)}|\delta_E^{(l)} + 2\Xi)$, and summing over $k \neq A^{(l)}, B^{(l)}$ gives the claim.
\end{proof}

\begin{corollary}[Within-tail activations equalize, but more slowly]
\label{cor:equalization-noise}
As a consequence of Corollary~\ref{cor:concentration-noise}, the within-tail equalization of Corollary~\ref{cor:equalization} carries over for tokens whose attention concentrates, with a noise-dependent rate correction and convergence to a noise-floor neighborhood rather than to equality exactly. Specifically, for a same-sign token $i$ with $\gamma|s_i^{(l)}|\delta_E^{(l)} > 2\Xi$, with $u_{i,E}^{(l)}$, $\varepsilon_{q,E}^{(l)}$, and $\eta_l$ defined as in Corollary~\ref{cor:equalization} but using the noisy attention, then whenever the denominator is positive,
\[
\bigl|u_{i,E}^{(l+1)}\bigr| \;\leq\; \frac{|u_{i,E}^{(l)}| + 2\lambda\eta_l + 2\mathcal{E}/|s_E^{(l)}|}{1 + \lambda - \lambda\eta_l - \mathcal{E}/|s_E^{(l)}|}.
\]
\end{corollary}

The asymptotic contraction rate is $(1+\lambda)^{-1}$ as in the noiseless case, so equalization eventually proceeds at the noiseless rate. At finite depth, the contraction is slower by an amount controlled by the residual off-pole attention $\eta_l \lesssim e^{2\Xi - \gamma|s_i|\delta_E}$ (delayed concentration) and the additive jitter $\mathcal{E}/|s_E^{(l)}|$ (from $\epsilon_i^{(l)}$). Convergence is to a noise-floor neighborhood of equality of size $O(\eta_l + \mathcal{E}/(\lambda|s_E^{(l)}|))$, which shrinks with depth as $|s_i|\delta_E$ grows past $2\Xi$ and $|s_E^{(l)}|$ grows. Tokens whose attention does not concentrate do not equalize.

\begin{proof}
As in the proof of Corollary~\ref{cor:equalization}, but with the noisy update $s_q^{(l+1)} = s_q^{(l)} + \lambda s_E^{(l)}(1 + \varepsilon_{q,E}^{(l)}) + \epsilon_q^{(l)}$ (now including the additive jitter $\epsilon_q^{(l)}$): plug into the definition $u_{i,E}^{(l+1)} = 1 - s_i^{(l+1)}/s_E^{(l+1)}$ for $q \in \{i, E\}$ and rearrange to obtain
\[
u_{i,E}^{(l+1)} \;=\; \frac{u_{i,E}^{(l)} + \lambda(\varepsilon_{E,E}^{(l)} - \varepsilon_{i,E}^{(l)}) + (\epsilon_E^{(l)} - \epsilon_i^{(l)})/s_E^{(l)}}{1 + \lambda(1 + \varepsilon_{E,E}^{(l)}) + \epsilon_E^{(l)}/s_E^{(l)}}.
\]
Take absolute values, use $|\varepsilon_{q,E}^{(l)}| \leq \eta_l$ and $|\epsilon_q^{(l)}| \leq \mathcal{E}$ from (N1), and apply the triangle inequality to numerator and denominator separately.
\end{proof}

\begin{corollary}[Tail-asymmetric amplification holds verbatim under noisy attention]
\label{cor:skew-noise}
Differentiating the per-token map $T_i^{(l)}$ from~\eqref{eq:scalar-dynamics} gives
\[
(T_i^{(l)})''(s) \;=\; \lambda\gamma^2 \cdot \E_{\alpha_i^{(l)}(\gamma s)}\bigl[(s^{(l)} - \E_{\alpha_i^{(l)}(\gamma s)} s^{(l)})^3\bigr], \qquad \alpha_i^{(l)}(t)_j := \softmax_j(t s_j^{(l)} + \xi_{ij}^{(l)}),
\]
identical in form to Theorem~\ref{thm:skew}, with the only difference being that the expectation is taken over the noisy attention $\alpha_i^{(l)}$.
\end{corollary}

The mechanism is intact: wherever this distribution is positively skewed, $(T_i^{(l)})'$ is locally increasing, so gaps in the upper tail amplify faster than gaps in the bulk. It is these growing tail gaps that are more likely to overcome the gap-amplification noise threshold and continue to grow multiplicatively, while bulk pairs may stagnate or reshuffle. The tokens involved in these gaps have large $|s_i^{(l)}|\delta_E^{(l)}$, eventually clearing the $2\Xi$ floor in the concentration bound and having their attention pulled onto the poles, while bulk tokens may have attention remain distributed.

\begin{proof}
In the noisy case, the per-token map is $T_i^{(l)}(s) = s + \lambda f_i^{(l)}(\gamma s) + \epsilon_i^{(l)}$, with $f_i^{(l)}(t) = \sum_j s_j^{(l)} \softmax_j(t s_j^{(l)} + \xi_{ij}^{(l)})$. As in the proof of Theorem~\ref{thm:skew}, take the second derivative with respect to $s$. The constant $\epsilon_i^{(l)}$ vanishes under $\partial^2/\partial s^2$; the $\xi_{ij}^{(l)}$ inside the softmax shifts the logits but does not depend on $s$, so it is inert under differentiation. The remaining calculation is identical to the noiseless one, applied to $f_i^{(l)}$ in place of $f^{(l)}$, yielding the stated identity with the third moment now under the noisy attention $\alpha_i^{(l)}(\gamma s)$.
\end{proof}

\section{Empirical Support for the Theoretical Mechanism}
\subsection{Concept-Aligned Attention Heads Exist in All Tested Models}
\label{app:head-validation}

\begin{figure}[t!]
    \centering
    \includegraphics[width=\textwidth]{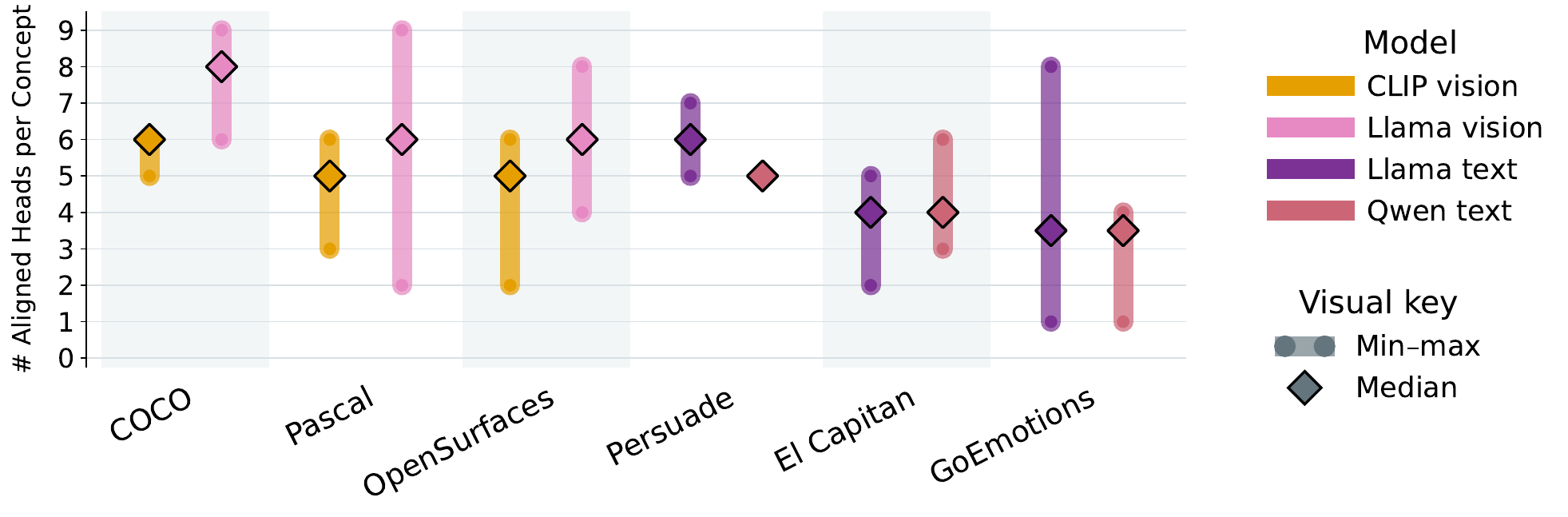}
    \caption{\textbf{All concepts have at least one concept-aligned head in every tested model}, with a median of 6 per concept, and nine at the most.}
    \label{fig:concept-aligned heads}
\end{figure}

We test whether attention heads exhibit the positive QK and OV alignment assumed by our theoretical mechanism. We fit concept directions $c_{\mathrm{in}}$ at the attention input and $c_{\mathrm{out}}$ immediately after the attention residual update. Each is fit from training positions as a covariance-ridge mean-difference direction between concept-positive and concept-negative representations, with ridge $\rho=10^{-4}$.

\paragraph{QK and OV alignment.}
For each head $h$, we measure concept alignment in its attention-routing and residual-write pathways. For QK alignment, we project $c_{\mathrm{in}}$ through the head's query and key projections, apply the architecture-specific Q/K projection normalization when present, and compute

$$
\hat{\gamma}_h
=
\frac{
q_h(c_{\mathrm{in}})^\top k_h(c_{\mathrm{in}})
}{
\sqrt{d_h}
}.
$$

Thus, $\hat{\gamma}_h>0$ means that same-sign concept activations contribute positively to their mutual attention logits, with larger values indicating stronger concept-driven attention. 

For OV alignment, we measure how a concept-aligned input is written back to the residual stream:

$$
\hat{\lambda}_h
=
c_{\mathrm{out}}^\top
\left(g_h W_O^{(h)}W_V^{(h)}\right)
c_{\mathrm{in}},
$$

where $g_h=1$ for ungated attention and otherwise denotes the learned attention-output gate. Positive $\hat{\lambda}_h$ therefore indicates that concept signal entering the head is written positively along the post-attention concept direction.

\paragraph{Concept-specific alignment.}
To test whether these positive alignments are specific to the concept rather than generic properties of the head, we compare each head against $5{,}000$ random concept directions matched to the joint covariance structure of the pre- and post-attention representations. These null directions undergo the same fitting and normalization procedure, yielding null-standardized scores $z_{\gamma,h}$ and $z_{\lambda,h}$. We score each head as $J_h=\min(z_{\gamma,h},z_{\lambda,h})$ when both $\hat{\gamma}_h>0$ and $\hat{\lambda}_h>0$, and $J_h=-\infty$ otherwise, requiring both pathways to show unusually strong positive alignment.

For each concept and depth, we select the head with the largest $J_h$. We correct for searching over heads by repeating the same selection procedure for every null draw, yielding an empirical one-sided $p$-value for the selected head. We then apply Benjamini--Hochberg correction across all concept$\times$depth tests within each dataset/model cell. A concept--depth pair is \emph{supported} when its selected head has positive QK and OV coefficients, $p\le0.05$, and FDR-adjusted $q\le0.05$.

Figure~\ref{fig:concept-aligned heads} shows that concept-aligned heads are consistently present in all evaluated models and datasets. Every concept has at least one head with significant positive QK and OV alignment in each model, with a median of six aligned heads per concept and as many as nine. Alignment is especially widespread in the vision models, while the text models show somewhat fewer aligned heads and greater variation across concepts.

\begin{figure}[t!]
    \centering
    \includegraphics[width=\textwidth]{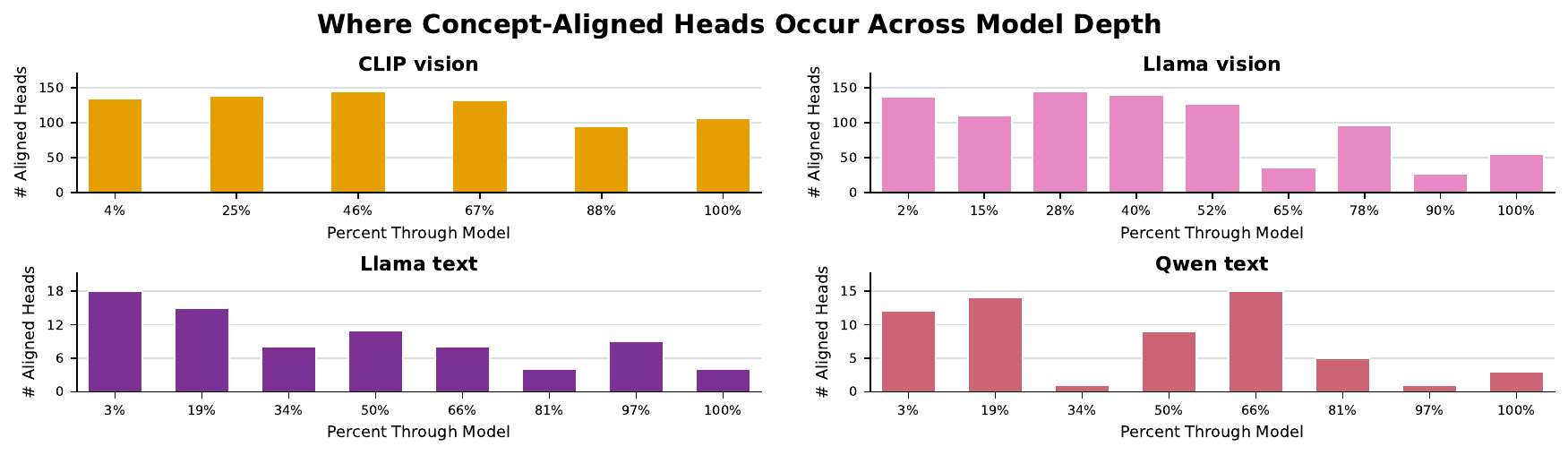}
    \caption{Early-to-mid layers are most likely to have concept-aligned heads.}
    \label{fig:aligned-heads-over-layers}
\end{figure}

In Figure~\ref{fig:aligned-heads-over-layers}, we plot where concept-aligned heads occur across model depth. Although aligned heads appear throughout the network, they are most concentrated in early-to-middle layers and become less frequent toward the final layers. This lines up with our empirical observation in Appendix~\ref{app:super-motivation} that activation separation occurs more rapidly in early-to-middle layers.

\subsection{Ablating Concept-Aligned Heads Preferentially Reduces the Tail--Bulk Gap}
\label{app:aligned-head-ablation}

We next test whether these concept-aligned heads causally contribute to the activation separation predicted by the \super{} Mechanism. For computational tractability, we conduct this analysis on \emph{COCO} and \emph{GoEmotions} and, for each model, evaluate the percent through model (PTM) containing the largest number of concept-aligned heads. All concepts are represented at the selected PTMs, except \emph{amusement}, \emph{gratitude}, and \emph{joy} in \emph{Qwen}, which are evaluated only in \emph{Llama}. Within each concept, we rank heads by their previously measured QK--OV alignment using $r_h=\sqrt{\max(z_{\gamma,h},0)\max(z_{\lambda,h},0)}$ for heads with positive $\hat{\gamma}_h$ and $\hat{\lambda}_h$. We then cumulatively ablate heads from most to least aligned and compare against $64$ random head orderings that remove the same number of heads at each step. Ablation removes the selected head's complete output immediately before the attention output projection while leaving the remainder of the model unchanged.

To measure the effect, we use the unablated attention-input activations along $c_{\mathrm{in}}$ to fix the top $5\%$ of positions as the \emph{tail} and the $40$--$60$th percentile as the \emph{bulk}. We retain these positions under every ablation and, after the attention residual update, score them along $c_{\mathrm{out}}$. At each cumulative ablation step, we measure how much the difference between mean tail and mean bulk activation decreases relative to the unablated model. We summarize each ablation curve by averaging this reduction across all intermediate numbers of removed heads, excluding the all-head endpoint where every ordering is identical.

\begin{figure}[t!]
    \centering
    \includegraphics[width=\textwidth]{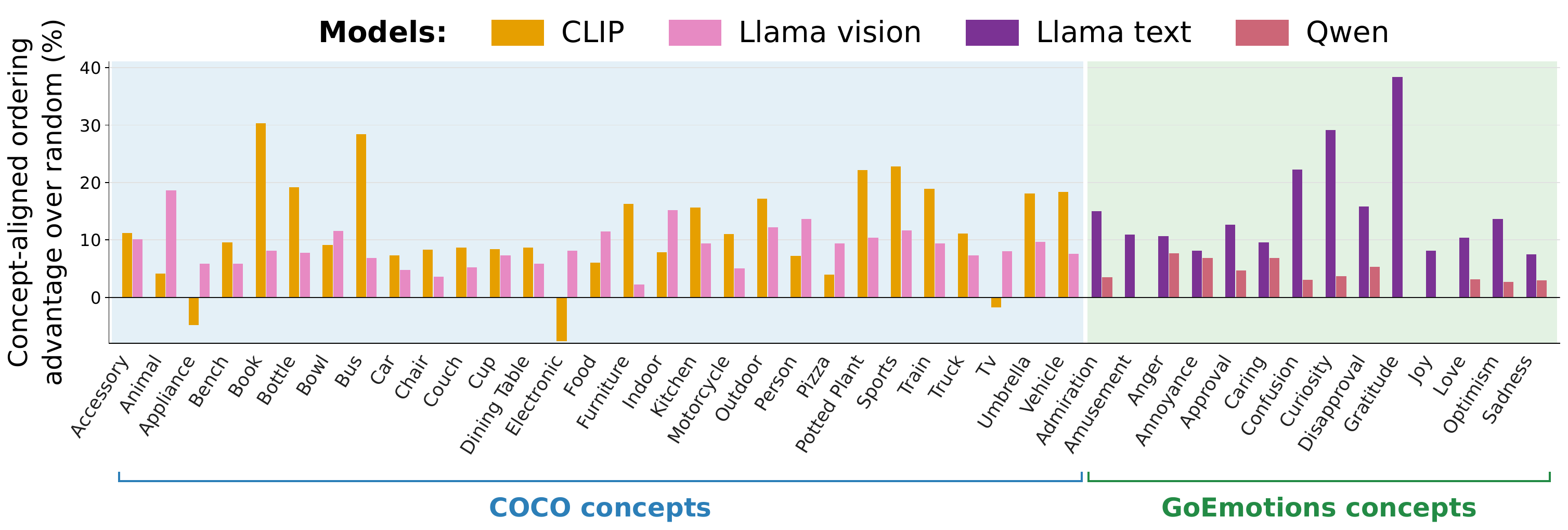}
    \caption{\textbf{Concept-aligned heads preferentially contribute to the local tail--bulk activation gap.} For each concept, heads are cumulatively ablated from most to least QK--OV aligned and compared with $64$ random head orderings. Bars show the aligned-order ablation advantage over the mean random ordering, normalized by the unablated tail--bulk gap; positive values indicate that removing concept-aligned heads reduces the gap more strongly than removing randomly ordered heads. This effect is positive for $80$ of $83$ concept--model pairs.}
    \label{fig:causal-results}
\end{figure}

Figure~\ref{fig:causal-results} compares the aligned-head ordering against the mean random ordering for each concept, normalized by that concept's baseline tail--bulk gap. Positive values indicate that ablating concept-aligned heads reduces the tail--bulk separation earlier and more strongly than randomly ordered heads. This effect appears consistently: $80$ of $83$ concept--model pairs ($96\%$) are positive. The only exceptions are \textit{Appliance}, \textit{Electronic}, and \textit{TV} in \emph{CLIP}. These are closely related \emph{COCO} categories and exhibit similar negative effects, although the source of this shared behavior is unclear; the corresponding concepts remain positive in \emph{Llama vision}. Thus, heads identified solely from their QK--OV alignment are consistently more important than random heads for driving the local separation between extreme and typical concept activations.

\subsection{Constructed Attention Heads Demonstrate the Predicted Amplification and Equalization Dynamics}
\label{sec:constructed-heads}

To directly examine the dynamics predicted by our theory, we construct synthetic attention heads around a fixed concept direction $v\in\mathbb{R}^{16}$. For each experimental condition, we first specify the desired effective QK and OV operators and then factor them via SVD into explicit weight matrices $W_Q,W_K,W_V\in\mathbb{R}^{8\times16}$ and $W_O\in\mathbb{R}^{16\times8}$. Without noise, these matrices are constructed to satisfy
\[
\frac{W_Q^\top W_K}{\sqrt{8}}=\gamma vv^\top,
\qquad
W_OW_V=\lambda vv^\top.
\]
Thus, $\gamma$ directly controls concept-aligned attention routing and $\lambda$ controls the corresponding concept-aligned residual write.

We then construct a synthetic sequence of $200$ token representations to pass through this head. To match the empirically observed positive skew of concept activations, we initialize each token with $s_i^{(0)}=0.5+\operatorname{LogNormal}(-1,0.7)$ along $v$, together with an orthogonal component $z_i$, so $x_i^{(0)}=s_i^{(0)}v+z_i$ and $z_i\perp v$. For each of $100$ seeds, we repeatedly apply the same frozen head for $100$ residual-attention steps using standard vector-valued attention,
\[
Q=XW_Q^\top,\qquad K=XW_K^\top,\qquad V=XW_V^\top,
\]
\[
A=\operatorname{softmax}\!\left(QK^\top/\sqrt{8}\right),
\qquad
X'=X+(AV)W_O^\top.
\]
Here, ``depth'' refers only to repeated application of this same constructed head, rather than distinct learned transformer layers. The resulting vector states are projected onto $v$ only after each update to obtain the scalar concept activations $s_i^{(\ell)}=v^\top x_i^{(\ell)}$ used for the plotted diagnostics.

\begin{figure}[t]
    \centering
    \includegraphics[width=\textwidth]{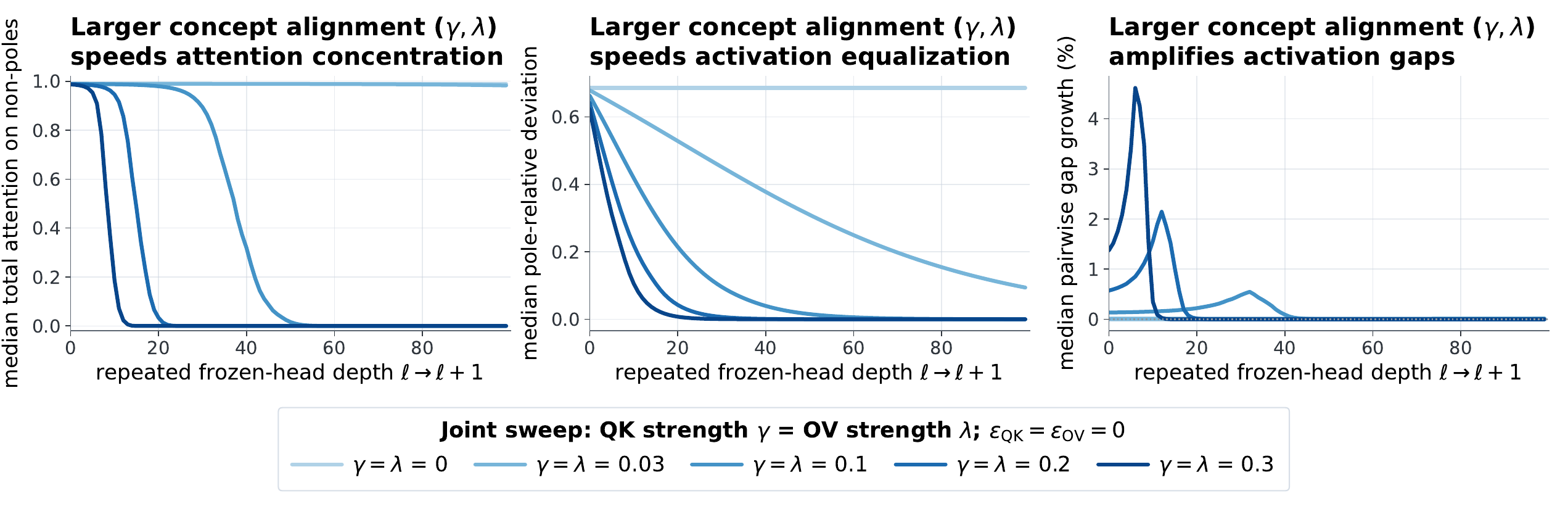}
    \caption{\textbf{Stronger concept alignment accelerates attention concentration and relative equalization while amplifying activation gaps.} We construct rank-one QK/OV-aligned attention heads and jointly vary $\gamma=\lambda$, with no off-concept terms. As predicted by our theory, repeated application of the same frozen head shows that larger alignment concentrates attention on the activation poles sooner (left), drives activations closer to the upper pole relative to its magnitude (middle), and produces stronger pairwise activation-gap growth until relative equalization dominates (right).}
    \label{fig:concept-aligned-comps}
\end{figure}

\textbf{Stronger concept alignment accelerates both activation gap amplification and equalization.}
We first use purely concept-aligned heads with no off-concept QK or OV components and jointly increase $\gamma=\lambda$. Figure~\ref{fig:concept-aligned-comps} tracks three quantities predicted by our analysis: (1) how much attention remains on non-pole tokens, (2) how far other activations remain from the maximum activation relative to the maximum's magnitude, and (3) how much pairwise activation differences grow from one step to the next. As predicted by our theory, increasing concept alignment causes attention to concentrate on the poles substantially earlier and in turn, activations approach the upper pole more quickly in \emph{relative} scale. More concept alignment also produces larger pairwise-gap amplification in the \super{} regime, until equalization eventually dominates.

\textbf{Non-concept components can delay or suppress these dynamics.}
We next ask how non-concept factors affect these dynamics while holding the concept alignment fixed at $\gamma=\lambda=0.1$. We introduce three orthogonal directions $q,k,r\perp v$: $q$ and $k$ create controlled perturbations to the attention logits, while $r$ creates a controlled perturbation to the OV write. These features are added to the synthetic token representations, and we construct
\[
M_{\mathrm{QK}}
=
\gamma vv^\top+\frac{\Xi}{a^2}qk^\top,
\qquad
M_{\mathrm{OV}}
=
\lambda vv^\top+\sigma\frac{\mathcal E}{a}vr^\top,
\]
where $a=0.25/\sqrt{3}$, $\Xi$ and $\mathcal E$ control the QK and OV noise strengths, and $\sigma\in\{-1,+1\}$ sets the direction of the OV perturbation. As before, we factor these effective operators into explicit $W_Q,W_K,W_V,W_O$ matrices and run the full vector-valued attention update. The orthogonal token features are chosen adversarially so that the added terms realize bounded disturbances from our theoretical analysis.

\begin{figure}[H]
    \centering
    \includegraphics[width=\textwidth]{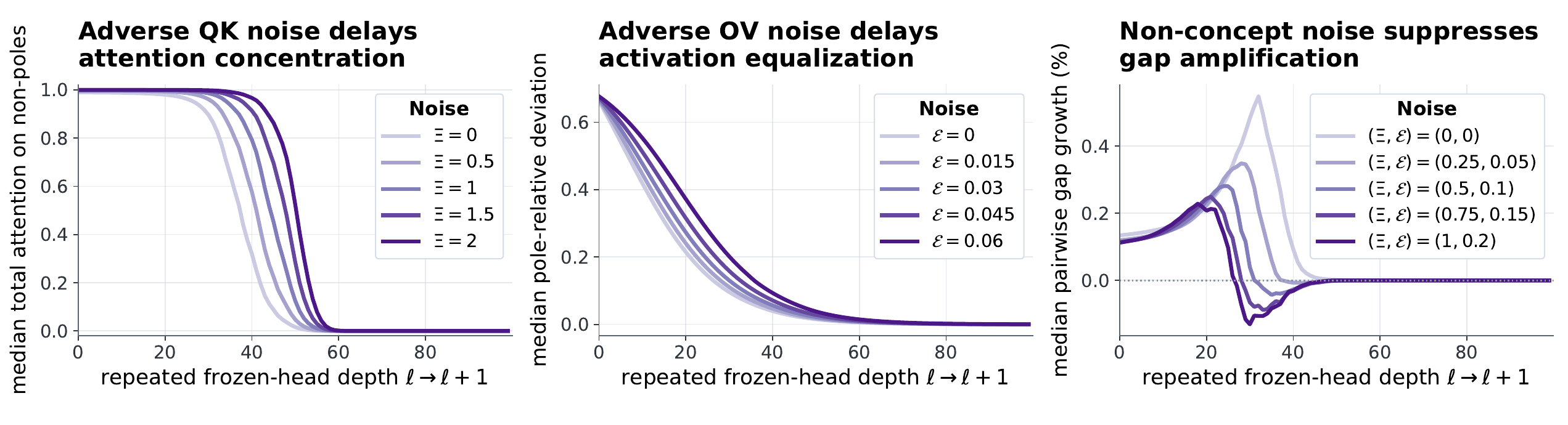}
    \caption{\textbf{Adverse non-concept noise delays or suppresses the predicted dynamics.} Holding the concept-aligned component fixed at $\gamma=\lambda=0.1$, we add controlled orthogonal perturbations to the constructed QK and OV operators. Increasing QK noise delays attention concentration onto the activation poles (left), increasing OV noise delays pole-relative equalization (middle), and jointly adverse QK and OV noise suppresses pairwise activation-gap growth and can even reverse it (right).}
    \label{fig:noise-comps}
\end{figure}

Figure~\ref{fig:noise-comps} confirms these effects predicted in the theory: increasing adverse QK noise delays attention concentration, adverse OV noise delays pole-relative equalization, and jointly adverse QK and OV components progressively suppress pairwise-gap amplification and can temporarily reverse it.

\section{Experimental Configurations}
\label{app:experiment-configs}

\subsection{Embedding Models}
\label{app:models-embeddings}
For images, we extract both input token and [CLS] token embeddings from the \emph{CLIP} ViT-L/14~\citep{Radford2021LearningTV} and \emph{Llama-3.2-11B-Vision-Instruct}~\citep{Dubey2024TheL3} models. For text, we use \emph{Llama-3.2-11B-Vision-Instruct} and \emph{Qwen3-Embedding-4B}~\citep{Zhang2025Qwen3EA}. Since these models lack an explicit [CLS] token for text inputs, we approximate a [CLS]-style representation by averaging token embeddings, a strategy found to be effective in prior work~\citep{Choi2021EvaluationOB, Tang2024PoolingAA}. For each model, we obtain embeddings across multiple layers. To ensure comparability, we normalize and mean-center each layer’s embeddings using statistics computed from the training set.  

To make the computation feasible, we evaluate models at a fixed set of percentage depths through the network, rather than at every layer. The chosen checkpoint percentages are \emph{CLIP}: [4, 25, 46, 67, 88, 100], \emph{Llama-Vision}: [2, 15, 28, 40, 52, 65, 78, 90, 100], \emph{Llama-Text}: [3, 19, 34, 50, 66, 81, 97, 100], \emph{Qwen}: [3, 19, 34, 50, 66, 81, 97, 100].

\subsection{Concept Extraction Methods}
\label{app:concept-extraction-techniques}
Throughout, let $x$ denote a sample (image or text), and $z(x) \in \mathbb{R}^d$ its embedding obtained from the underlying model. For a ground-truth concept $c$, let $\mathcal{X}^+_c$ denote the set of samples labeled positive for $c$. We use $v_c \in \mathbb{R}^d$ to denote the concept vector associated with $c$, and $v_j$ to denote candidate concept vectors discovered by an unsupervised method.  
All concepts are constructed only using embeddings from the training set.

We extract concepts using supervised methods, unsupervised methods, and a prompting baseline. Concept representations are computed at both the token level, using embeddings from input tokens, and the [CLS] level, using embeddings from the [CLS] tokens, which lie in a distinct representational space optimized for sequence-level summarization.

\paragraph{Supervised Methods:}  
\begin{enumerate}
    \item \textbf{Mean Prototypes}~\citep{Zou2023RepresentationEA}: Each concept vector is defined as the average embedding of all positive examples,
    \[
    v_c = \frac{1}{|\mathcal{X}^+_c|} \sum_{x \in \mathcal{X}^+_c} z(x).
    \]

    \item \textbf{Linear Separators (LinSep)}~\citep{Kim2017InterpretabilityBF}: For each concept $c$, we train a linear model (without bias) to distinguish positives from negatives. For training, we balance positive and negative samples and use \texttt{BCEWithLogitsLoss} with the Adam optimizer (learning rate $0.01$). We train for up to 100 epochs with a batch size of 32, apply weight decay of $1\text{e}{-4}$, and decay the learning rate by a factor of 0.5 every 10 epochs. Early stopping is used with a patience of 15 epochs and a tolerance of 3, which sets the minimum improvement required to continue training. The resulting normal vector of the separating hyperplane is used as the concept vector:
    \[
    v_c = w_c.
    \]
\end{enumerate}

\paragraph{Unsupervised Methods:}  
\begin{enumerate}
    \item \textbf{K-Means Prototypes}~\citep{Ghorbani2019AutomatingID}: We cluster embeddings using FAISS GPU~\citep{Johnson2017BillionScaleSS} with Euclidean distance, a maximum of 300 iterations, and $k{=}1000$ for token-level embeddings and $k{=}50$ for [CLS] embeddings. The choice of $k$ was determined experimentally using an elbow curve. Token-level embeddings are finer-grained and therefore benefit from a larger number of clusters. Each cluster centroid is used as a concept vector:
    \[
    v_j = \mu_j = \frac{1}{|\mathcal{C}_j|} \sum_{x \in \mathcal{C}_j} z(x).
    \]

    \item \textbf{Cluster-Based Separators (K-LinSep)}: We first assign soft labels to embeddings based on their K-means cluster membership, then train linear separators with the same procedure described above to predict whether an embedding belongs to a given cluster. The normal vectors of these separators are treated as concept directions:
    \[
    v_{ij} = w_{ij}.
    \]

    \item \textbf{Sparse Autoencoders (SAEs)}~\citep{bricken2023monosemanticity}: SAEs learn a sparse reconstruction
    \[
    z(x) \approx W h(x), \quad h(x) \in \mathbb{R}^m \ \text{ sparse}, \quad v_j = w_j,
    \]
    where each column $w_j$ of $W$ corresponds to a candidate concept. Because SAE training is computationally expensive, we use pretrained SAEs; see Appendix~\ref{app:saes} for architectural and implementation details.
\end{enumerate}

To ensure we can evaluate against unsupervised methods, each ground-truth concept $c$ is matched to the unsupervised cluster $v_j$ that achieves the highest validation F$_1$ score for detecting $c$:
\[
v_c = \arg\max_{v_j} \;\; \text{F}_1^{\text{val}}(c, v_j).
\]

\paragraph{Prompt Baseline:}  
As an additional baseline, when the model supports prompting, we query the corresponding model directly. For each sample $x$ and concept $c$, we prompt:
\[
\text{``Is the concept of $c$ present in the following? $x$''}.
\]
Prior works have employed similar zero-shot prompting baselines successfully~\citep{Wu2025AxBenchSL, Robicheaux2025Roboflow100VLAM, Tillman2025InvestigatingTP}. We perform greedy generation (temperature 0) and label the sample $x$ as containing concept $c$ if and only if the response is ``Yes''.

\subsection{Dataset Overview}
\label{app:dataset_details}

We primarily evaluate on six image and text datasets:

\textbf{COCO \citep{Lin2014MicrosoftCC}:} We use the 2017 validation set of \emph{MS-COCO}, containing 5{,}000 images with everyday scenes involving people, objects, and natural contexts. Each image comes with human-annotated segmentations, providing dense labels for both object categories and broader supercategories.

\textbf{Broden–Pascal  \citep{everingham2010pascalvoc} and Broden–OpenSurfaces \citep{bell13opensurfaces}:} We use 4{,}503 samples from Pascal and 3{,}578 samples from OpenSurfaces. These are subsets of the Broden dataset \citep{bau2017network}, which unifies multiple segmentation datasets into a single benchmark for concept-based interpretability research. Pascal primarily contains natural images with segmented objects from diverse categories such as animals, vehicles, and household items, while OpenSurfaces emphasizes fine-grained material and surface property annotations (e.g., wood, fabric, metal). We chose these two subsets because they focus on patch-level segmentation where concepts do not necessarily span the entire image.

\textbf{PERSUADE \citep{CROSSLEY2024100865}:} We draw from the Feedback Prize training split of PERSUADE 2.0, a corpus of argumentative essays written by U.S. students in grades 6--12 across 15 writing prompts. Essays are annotated at the span level with seven discourse elements, providing localized labels for argumentative structure within longer passages. From the competition-training essays containing fewer than 800 words, we sample 8{,}868 essays for our experiments to maintain computational tractability.

\textbf{El Capitan \citep{ElCapitanDataset}:} We use 2{,}232 English-language customer reviews of OS X El Capitan, comprising 10{,}348 sentences. Each sample consists of the full review body, constructed by joining its constituent sentences. The corpus provides human-annotated topic labels at both the review and sentence levels, enabling localized analysis of opinions about operating-system features, performance, updates, and related issues.

\textbf{GoEmotions (Augmented):} We use 5{,}427 samples from the GoEmotions dataset \citep{Demszky2020GoEmotionsAD}, a human-annotated collection of Reddit comments labeled with 27 emotion categories. We augment selected samples by embedding emotional sentences within surrounding neutral context, tagging the emotional span while preserving natural paragraph flow.

We additionally evaluate on two synthetic datasets that provide greater control over concept presence and localization. While these settings allow for more targeted analysis, the generation and augmentation procedures may introduce artifacts not present in naturally occurring data: 

\textbf{CLEVR (Single-Object) \citep{Johnson2016CLEVRAD}:} A synthetic dataset of 1{,}000 images, each containing a red, green, or blue object with shape sphere, cylinder, or cube. Images and segmentation masks are generated programmatically, allowing fine-grained control over object properties and patch-level annotations. 

\textbf{iSarcasm (Augmented):} We adapt 1{,}734 samples from the original iSarcasm dataset \citep{Oprea2019iSarcasmAD}, which provides sarcastic tweets alongside non-sarcastic rewrites conveying the same meaning (both provided by the original authors). We augment these by embedding sarcastic and non-sarcastic sentences into short paragraphs of neutral context, with sarcastic spans explicitly marked.

\subsection{Text Augmentation Pipelines and Prompts}
\label{app:prompts}

This section describes the augmentation pipelines used for generating and adapting
text datasets, along with the exact prompts. Our goal was to create datasets with
localized token-level concept spans, since most publicly available text datasets
only provide sample-level (sentence, tweet, comment, etc.) labels. Generation and
augmentation are performed via controlled prompting of GPT-4o
\citep{Hurst2024GPT4oSC}. 

\subsubsection{iSarcasm Augmentation}

\textbf{Dataset Overview:}
The original iSarcasm dataset contains sarcastic tweets paired with author-provided
sincere rewrites conveying the same meaning. We extend this dataset synthetically
by surrounding the sarcastic tweets with literal, neutral context, ensuring
precise span-level supervision. Only sarcastic samples are selected for
augmentation, and for each sarcastic input we generate both a sarcastic augmented
post and a non-sarcastic rewrite.

\textbf{Augmentation Pipeline:}
Each sarcastic input is expanded into casual, paragraph-like text using controlled
prompting of GPT-4o. To introduce variation, random structural features are
applied:
\begin{itemize}
    \item 20\% chance of forcing a \texttt{[Sarcasm][Trigger]} structure.
    \item 15\% chance of adding emojis or hashtags.
    \item Otherwise, random choice among \texttt{[Sarc][Trig]},
          \texttt{[Trig][Sarc]}, and \texttt{[Trig][Sarc][Trig]}.
\end{itemize}

\textbf{Sarcastic Augmentation Prompt:}

You are a data annotation machine. Your only goal is to produce perfectly literal
text that follows the rules. You must not be creative or clever. You must not
generate any figurative language outside of the provided tags.

Your Task:
You will be given a sarcastic tweet and its true meaning. Rewrite the tweet by
embedding it within a strictly literal train of thought that matches the
original's casual tone.

Structure: [Randomly choose or force specific structure]
[Optional emoji/hashtag instruction if selected]

Constraints Checklist:
- The tone is casual and informal.
- The added text is not redundant.
- Outside <SARCASM> tags is strictly literal and descriptive.
- The original sarcastic tweet is fully preserved within <SARCASM> tags.
- Output contains ONLY the final post.

Input Sarcastic Tweet: \{sarcastic tweet\}\\
Sincere Meaning (for your context): \{rephrased text\}

Your Output:

\textbf{Non-Sarcastic Augmentation Prompt:}

You are a data annotation machine. Your only goal is to produce perfectly literal
text that follows the rules. You must not be creative or clever. You must not
invent new details.

Your Task:
Take a sincere idea and expand it slightly into a personal, casual post,
remaining 100\% faithful to the original meaning.

[Optional emoji/hashtag instruction if selected]

Constraints Checklist:
- The tone is casual and informal.
- The entire post is strictly literal and descriptive.
- No sarcasm, irony, overstatement, or rhetorical questions.
- The post must be 100\% faithful to the meaning of the original idea.
- Output contains ONLY the final post.

Input Sincere Idea: \{rephrased text\}

Your Output:

\textbf{Verification Process:}
Outputs are verified via flexible matching with progressively lenient checks:
exact matching (case-insensitive), whitespace normalization,
URL/punctuation removal, and word-overlap thresholds. If all attempts fail, the
original tweet is wrapped in \texttt{<SARCASM>} tags as a fallback.

\textbf{Example:}

Input sarcastic tweet:
``The only thing I got from college is a caffeine addiction.''

Input sincere rephrase:
``College is really difficult, expensive, tiring, and I often question if a degree
is worth the stress.''

Sarcastic augmentation:
``I just checked my calendar and saw how many assignments are due this week.
\textless SARCASM\textgreater the only thing I got from college is a caffeine
addiction \textless /SARCASM\textgreater''

Non-sarcastic rewrite:
``college is really difficult. it's also expensive and tiring. sometimes i find
myself questioning if getting a degree is worth all the stress.''

\subsubsection{GoEmotions Augmentation}

\textbf{Dataset Overview:}
GoEmotions is a large-scale dataset of Reddit comments labeled with up to 27
fine-grained emotions. We extend it synthetically by surrounding the original
emotional comment with strictly neutral filler context, ensuring the emotional
span remains localized and clearly marked with \texttt{<EMOTION>} tags.

\textbf{Augmentation Pipeline:}
Every comment in GoEmotions is augmented without filtering, following a two-step
process:
\begin{enumerate}
    \item \textbf{Step 1: Generation.} A ``Neutral Filler Machine'' prompt is used
    to generate five diverse neutral-context options embedding the original
    emotional comment.
    \item \textbf{Step 2: Selection.} A ``Grader'' prompt evaluates the five drafts
    and selects the best single option according to neutrality and naturalness.
\end{enumerate}

To increase variation, a random structure is sampled per comment:
\begin{itemize}
    \item 50\% chance: \texttt{[Emotion][Context]}
    \item 25\% chance: \texttt{[Context][Emotion]}
    \item 25\% chance: \texttt{[Context][Emotion][Context]}
\end{itemize}

\textbf{Step 1 — Neutral Filler Prompt:}

You are a Neutral Filler Machine. Your task is to generate neutral, non-emotional
text to surround a given Reddit comment.

Task:
- Preserve the original emotional comment exactly inside <EMOTION> tags.
- Generate five unique and diverse neutral contexts that flow naturally.
- All options must follow the required structure.

Constraints:
- Text outside <EMOTION> must be strictly neutral (no emotion leakage).
- Sound natural and casual like a Reddit post.
- No redundancy with the emotional comment.

Input Emotional Comment: \texttt{\{emotional\_comment\}}
Primary Emotion(s): \texttt{\{emotion\_labels\_str\}}
Required Structure: \texttt{\{structure\_choice\}}

Your Output: Five options, each in the correct structure.

\textbf{Step 2 — Selection Prompt:}

You are a data annotation quality assurance specialist.
Your task is to select the best draft among five options.

Checklist:
- Context must be strictly neutral (no emotions).
- Flow naturally as a Reddit comment.
- No contradiction or redundancy.
- Only output the single best final option.

Draft Options:
\texttt{\{draft\_options\}}

Your Final, Best Output:

\textbf{Verification Process:}
The augmented comments are verified using flexible string matching to ensure that
the original text is preserved inside \texttt{<EMOTION>} tags. We allow up to five
retry attempts with progressively lenient checks. If all attempts fail, the
fallback is to wrap the original comment directly in \texttt{<EMOTION>} tags.

\textbf{Example:}

Original emotional comment (gratitude):
``I didn't know that, thank you for teaching me something today!''

Augmented output:
``A comment explained the process behind recycling plastics and how it affects the
environment.
\textless EMOTION\textgreater I didn't know that, thank you for teaching me
something today! \textless /EMOTION\textgreater''

\subsection{Concepts Used in Experiments}
\label{app:concepts}
For the MS-COCO, GoEmotions, and Broden datasets, we filter concepts using minimum sample thresholds (100–300 samples, depending on the dataset) to ensure sufficient data for reliable concept construction, though future work could examine \supers{} in underfit settings. The semantic concepts used in our experiments are listed here:

\begin{itemize}
    \item \textbf{COCO:} accessory, animal, appliance, bench, book, bottle, bowl, bus, car, chair, couch, cup, dining table, electronic, food, furniture, indoor, kitchen, motorcycle, outdoor, person, pizza, potted plant, sports, train, truck, tv, umbrella, vehicle
    \item \textbf{Broden–OpenSurfaces:} brick, cardboard, carpet, ceramic, concrete, fabric, food, fur, glass, granite, hair, laminate, leather, metal, mirror, painted, paper, plastic-clear, plastic-opaque, rock, rubber, skin, tile, wallpaper, wicker, wood
    \item \textbf{Broden–Pascal:} airplane, bicycle, bird, boat, body, book, building, bus, cap, car, cat, cup, dog, door, ear, engine, grass, hair, horse, leg, mirror, motorbike, mountain, painting, person, pottedplant, saddle, screen, sky, sofa, table, track, train, tvmonitor, wheel, wood, arm, bag, beak, bottle, box, cabinet, ceiling, chain wheel, chair, coach, curtain, eye, eyebrow, fabric, fence, floor, foot, ground, hand, handle bar, head, headlight, light, mouth, muzzle, neck, nose, paw, plant, plate, plaything, pole, pot, road, rock, rope, shelves, sidewalk, signboard, stern, tail, torso, tree, wall, water, windowpane, wing
    \item \textbf{PERSUADE:} counterclaim, rebuttal
    \item \textbf{El Capitan:} update, performance, el capitan, feature, speed, mail, download, installation, yosemite 
    \item \textbf{GoEmotions:} confusion, joy, sadness, anger, love, caring, optimism, amusement, curiosity, disapproval, approval, annoyance, gratitude, admiration
    \item \textbf{CLEVR:} blue, green, red, cube, cylinder, sphere
    \item \textbf{iSarcasm:} sarcastic
\end{itemize}

\section{Concept Formalisms in More Detail}
\label{app:detection-formalisms}

We provide a detailed formalization of concept detection and activation aggregation strategies, focusing on transformer architectures given their demonstrated effectiveness across modalities.

\textbf{Model Representations.}\hspace{0.5em}  
Let $f$ be a trained transformer model that processes an input $x \in \mathcal{X}$ (an image or a text sequence) into a set of hidden representations. At a given layer $\ell$, we extract token-level embeddings 
\[
f_\ell(x) = \{\, z^{\text{tok}}_1(x), \dots, z^{\text{tok}}_{n(x)}(x), z^{\text{cls}}(x) \,\}, 
\quad z^{\text{tok}}_i(x), z^{\text{cls}}(x) \in \mathbb{R}^d.
\]
Here $z^{\text{tok}}_i(x)$ denotes the representation of the $i$-th token (or image patch), and $z^{\text{cls}}(x)$ denotes the [CLS]-style representation summarizing the full input.

\paragraph{Concept Vectors and Activation Scores.}  
For any semantic concept $c$, we define a \textbf{concept vector} $v_c \in \mathbb{R}^d$, extracted via one of the techniques in Appendix~\ref{app:concept-extraction-techniques}. Intuitively, $v_c$ represents a direction in embedding space along which the concept $c$ is encoded.  
The \textbf{activation score} of an embedding $z$ with respect to concept $c$ is defined as
\[
s_c(z) = \langle z, v_c \rangle.
\]
If $v_c$ is derived as a cluster centroid, this corresponds to cosine similarity (for normalized embeddings). If $v_c$ is derived from a linear separator, it corresponds to the signed distance from the separating hyperplane. Intuitively, $s_c(z)$ measures the alignment of $z$ with concept $c$: large positive values indicate that $z$ strongly encodes features associated with $c$, while negative values suggest opposition or absence.  

We aim to characterize, for each concept $c$, the distribution of activation scores across many samples.  
Let $\mathcal{D}_c^{\text{in}}$ and $\mathcal{D}_c^{\text{out}}$ denote the population-level distributions of activation scores for in-concept and out-of-concept tokens, respectively.  
Empirically, we approximate these distributions using finite datasets $D_c^{\text{in}}$ and $D_c^{\text{out}}$ constructed from observed activations.  Let $Z$ denote the set of all tokens across samples, and let $S_c = \{\, s_c(z) : z \in Z \,\}$ be their corresponding activation scores.  
If $Z_c^{\text{in}} \subseteq Z$ are the tokens labeled concept-positive for concept $c$ and $Z_c^{\text{out}}$ are the tokens drawn from samples that do \emph{not} contain $c$ (thus excluding out-of-concept tokens from samples containing $c$ to avoid self-attention leakage), then
\[
D_c^{\text{in}} = \{\, s_c(z) : z \in Z_c^{\text{in}} \,\}, 
\qquad
D_c^{\text{out}} = \{\, s_c(z) : z \in Z_c^{\text{out}} \,\},
\]
which serve as empirical samples from $\mathcal{D}_c^{\text{in}}$ and $\mathcal{D}_c^{\text{out}}$. 
We use $Q_q(\mathcal{D})$ to denote the population $q$-quantile of a distribution $\mathcal{D}$, and $q_q(D)$ to denote its empirical estimate computed from a finite sample $D$.

\paragraph{Concept Detection.}  
The goal of concept detection is to determine whether a sample $x$ contains a concept $c$~\citep{Wu2025AxBenchSL}. Transformer models produce a collection of activation scores at the token level, but for detection we require a single score per sample. This necessitates an \textbf{aggregation operator} that interprets the set of token-level activations as a sample-level score.  

Let $S_c(x) = \{s_{c,1}(x), \dots, s_{c,n(x)}(x), s_{c,\text{cls}}(x)\}$ denote the set of activation scores for concept $c$ on input $x$, where $s_{c,i}(x)$ is the score for the $i$-th token and $s_{c,\text{cls}}(x)$ is the score for the [CLS] token. An aggregation operator is any function
\[
G: \mathbb{R}^{n(x)+1} \to \mathbb{R}, \quad s_c^{\text{agg}}(x) = G(S_c(x)).
\]
Given a calibrated threshold $\tau_c$, detection is performed by
\[
\hat{y}_c(x) = \mathbf{1}\!\left[\, s_c^{\text{agg}}(x) \geq \tau_c \,\right].
\]

Because prior work has shown that different concepts may emerge at different layers of a transformer~\citep{Salam2025LargeLM, Yu2024LatentCE, Dalvi2022DiscoveringLC}, we calibrate the layer separately for each concept to avoid enforcing a strict shared choice. This calibration is also performed independently for each aggregation strategy, ensuring that no operator is unfairly advantaged or disadvantaged due to layer-specific biases.

\paragraph{Standard Aggregation Strategies.}  
Prior work has considered several choices of $G$, each operating on the same token-level activations (with the exception of [CLS], which uses separately trained concept vectors since sample-level and input token-level representations occupy different spaces):
\begin{itemize}
    \item \textbf{Max pooling ($G_{\text{max}}$):} 
    \[
    G_{\text{max}}(S_c(x)) = \max\{s_{c,1}(x), \dots, s_{c,n(x)}(x), s_{c,\text{cls}}(x)\}.
    \] 
    Takes the strongest activation across input tokens. This is effective for isolating the most distinct concept signals~\citep{Tillman2025InvestigatingTP, Wu2025AxBenchSL}. This aggregation strategy is also implied by the \super{} Mechanism.

    \item \textbf{Mean pooling ($G_{\text{mean}}$):} 
    \[
    G_{\text{mean}}(S_c(x)) = \tfrac{1}{n(x)} \sum_{i=1}^{n(x)} s_{c,i}(x).
    \] 
    Averages over all tokens. This ensures that no part of the input is ignored and can capture distributed concept signals. Used in multiple studies such as \cite{McKenzie2025DetectingHI} and \cite{Suresh2025FromNT}.   

    \item \textbf{Last token ($G_{\text{last}}$):} 
    \[
    G_{\text{last}}(S_c(x)) = s_{c,n(x)}(x).
    \] 
    Uses the last input token activation. For autoregressive models, the final token often encodes sequence-level information, making it a plausible summary for concept detection. Used in \cite{Chen2025PersonaVM}, \cite{Tillman2025InvestigatingTP}, and \cite{Tang2024PoolingAA}.  

    \item \textbf{[CLS]-only ($G_{\text{cls}}$):} 
    \[
    G_{\text{cls}}(S_c(x)) = s_{c,\text{cls}}(x).
    \] 
    Uses only the [CLS] token score. Since CLS tokens are trained to attend to all inputs, they are natural candidates for summarizing sample-level concepts, and this strategy has been adopted in works such as \cite{Nejadgholi2022TowardsPF}, \cite{Yu2024LatentCE}, and \cite{Behrendt2025MaxPoolBERTEB}.  

    \item \textbf{Random token ($G_{\text{rand}}$):} 
    \[
    G_{\text{rand}}(S_c(x)) = s_{c,j}(x), \quad j \sim \text{Unif}\{1,\dots,n(x)\}.
    \] 
    Selects an input token activation uniformly at random. While a weak baseline, self-attention mechanisms distribute information broadly, so even a randomly chosen token may retain meaningful concept cues.  
\end{itemize}
These operators differ only in how they aggregate activations; they do not alter how concept vectors are trained. For each concept and aggregation method, the detection threshold $\tau_c$ is tuned continuously on the validation set to maximize detection performance. At test time, we apply the same aggregation operator $G$ to the sample activations and predict concept presence by comparing the resulting score against the fixed validation-selected threshold $\tau_c$.

\paragraph{Sparsity-Controlled Max Detection.}
To characterize how sparse the most reliable concept signal is within the in-concept activation distribution, we use a \super{}-based detector whose decision threshold is parameterized directly by a sparsity level $\delta_c$. 

Formally, let $D_{c,\mathrm{val}}^{\mathrm{in}}$ denote the activation scores of all labeled in-concept tokens in the validation set. For a sparsity level $\delta_c$, we define the \super{} threshold as $\tau_c^{\text{super}}(\delta_c) = Q_{1-\delta_c}(D_{c,\mathrm{val}}^{\mathrm{in}})$, so that approximately a fraction $\delta_c$ of in-concept token activations exceed the threshold.

At test time, we apply max aggregation and predict concept presence when the maximum token activation exceeds $\tau_c^{\text{super}}(\delta_c)$. For the sparsity analysis, we evaluate detection F1 across a fixed grid of \(\delta_c\) values to characterize how performance varies with the in-concept tail fraction. The main detection results instead use the continuously tuned threshold described above, calibrated using sample-level concept-presence labels.

\section{Comprehensive Concept Detection Results}
\label{app:detection-results}
The following tables compare \super{}-based aggregation with other standard aggregation and prompting baselines across datasets, models, and concept types. The first table provides random and constant-predictor detection performances for each dataset–model combination for reference. Each table reports the average $F_1$ detection scores, computed as the mean across concepts weighted by their frequency in the test set. Error bars denote 95\% bootstrap confidence intervals obtained by resampling evaluation samples and recomputing detection metrics. For each model/concept-type combination, the best-performing detection method on average is shown in \textbf{bold} and the second best is \underline{underlined}.

Across our six primary image and text datasets, SuperAct-based aggregation achieves the highest average $F_1$ in 47 of 48 model/concept-type combinations. In the single exception, the reported uncertainty intervals overlap. The strongest alternative varies between mean pooling (MeanTok), CLS token aggregation (CLSTok), and prompting. 

Results on CLEVR largely follow this trend, with SuperAct achieving the highest $F_1$ in six of eight configurations and tying in one. However, results on iSarcasm are less consistent, with SuperAct leading in only one of eight configurations, although four of the seven remaining cases have overlapping error bars. This discrepancy may be related to artifacts introduced during synthetic data generation, though the underlying cause remains unclear.

\begin{table}[h]
\centering
% Requires \usepackage{booktabs,graphicx} in the preamble.
% To use \include inside a table float, add to the main preamble:
%   \usepackage{etoolbox}
%   \AtBeginEnvironment{table}{\let\include\input}
% The duplicated model dimension has been removed; every result is unchanged.
\begingroup
\footnotesize
\caption{Baseline Detection $F_1$}
\setlength{\tabcolsep}{3pt}
\renewcommand{\arraystretch}{1.15}
\resizebox{\linewidth}{!}{%
\begin{tabular}{@{}lcccccccc@{}}
\toprule
Baseline & COCO & \shortstack{Open\\Surfaces} & Pascal & PERSUADE & \shortstack{El\\Capitan} & GoEmotions & CLEVR & iSarcasm \\
\midrule
Constant Positive & $0.317\,{\scriptstyle\pm 0.007}$ & $0.341\,{\scriptstyle\pm 0.009}$ & $0.380\,{\scriptstyle\pm 0.005}$ & $0.402\,{\scriptstyle\pm 0.019}$ & $0.415\,{\scriptstyle\pm 0.019}$ & $0.102\,{\scriptstyle\pm 0.007}$ & $0.504\,{\scriptstyle\pm 0.033}$ & $0.676\,{\scriptstyle\pm 0.047}$ \\
Random & $0.262\,{\scriptstyle\pm 0.010}$ & $0.281\,{\scriptstyle\pm 0.012}$ & $0.307\,{\scriptstyle\pm 0.006}$ & $0.334\,{\scriptstyle\pm 0.023}$ & $0.337\,{\scriptstyle\pm 0.024}$ & $0.096\,{\scriptstyle\pm 0.010}$ & $0.402\,{\scriptstyle\pm 0.043}$ & $0.505\,{\scriptstyle\pm 0.064}$ \\
\bottomrule
\end{tabular}%
}
\endgroup

\end{table}

\begin{table}[h]
\centering
% COCO
% Match to Broden-Pascal source style.
% Requires booktabs, multirow, makecell, tabletextsize, tablenumbersize.
\centering
\setlength{\tabcolsep}{4.7pt}
\caption{Concept detection $F_1$ for the \textbf{COCO} dataset.}
\tabletextsize{
\begin{tabular}{l l c ccc c c}
    \toprule
    Model & \multirow{2}{*}{\makecell{Concept\\Type}} & \multicolumn{6}{c}{Concept Detection Methods} \\
    \cmidrule(lr){3-8}
     &  & RandTok & LastTok & MeanTok & CLSTok & Prompt & \textbf{SuperAct} \\
    \midrule
\multirow{4}{*}{CLIP} & Avg & {\tablenumbersize 0.368 {\scriptsize $\pm$ 0.013}} & {\tablenumbersize 0.347 {\scriptsize $\pm$ 0.012}} & {\tablenumbersize 0.531 {\scriptsize $\pm$ 0.014}} & \underline{{\tablenumbersize 0.679 {\scriptsize $\pm$ 0.012}}} & -- & \textbf{{\tablenumbersize 0.718 {\scriptsize $\pm$ 0.012}}} \\
 & Linsep & {\tablenumbersize 0.514 {\scriptsize $\pm$ 0.014}} & {\tablenumbersize 0.516 {\scriptsize $\pm$ 0.014}} & {\tablenumbersize 0.681 {\scriptsize $\pm$ 0.013}} & \underline{{\tablenumbersize 0.702 {\scriptsize $\pm$ 0.012}}} & -- & \textbf{{\tablenumbersize 0.791 {\scriptsize $\pm$ 0.011}}} \\
 & K-Means & {\tablenumbersize 0.350 {\scriptsize $\pm$ 0.013}} & {\tablenumbersize 0.350 {\scriptsize $\pm$ 0.013}} & {\tablenumbersize 0.543 {\scriptsize $\pm$ 0.014}} & \underline{{\tablenumbersize 0.582 {\scriptsize $\pm$ 0.012}}} & -- & \textbf{{\tablenumbersize 0.701 {\scriptsize $\pm$ 0.012}}} \\
 & K-Linsep & {\tablenumbersize 0.347 {\scriptsize $\pm$ 0.013}} & {\tablenumbersize 0.368 {\scriptsize $\pm$ 0.014}} & \underline{{\tablenumbersize 0.591 {\scriptsize $\pm$ 0.014}}} & {\tablenumbersize 0.570 {\scriptsize $\pm$ 0.013}} & -- & \textbf{{\tablenumbersize 0.713 {\scriptsize $\pm$ 0.012}}} \\
\midrule
\multirow{4}{*}{Llama} & Avg & {\tablenumbersize 0.340 {\scriptsize $\pm$ 0.012}} & {\tablenumbersize 0.315 {\scriptsize $\pm$ 0.011}} & {\tablenumbersize 0.551 {\scriptsize $\pm$ 0.013}} & {\tablenumbersize 0.531 {\scriptsize $\pm$ 0.014}} & \underline{{\tablenumbersize 0.686 {\scriptsize $\pm$ 0.010}}} & \textbf{{\tablenumbersize 0.754 {\scriptsize $\pm$ 0.012}}} \\
\bottomrule
\end{tabular}
}
\label{tab:coco-detection}

\end{table}

\begin{table}[h]
\centering
% OpenSurfaces
% Match to Broden-Pascal source style.
% Requires booktabs, multirow, makecell, tabletextsize, tablenumbersize.
\centering
\setlength{\tabcolsep}{4.7pt}
\caption{Concept detection $F_1$ for the \textbf{OpenSurfaces} dataset.}
{\tabletextsize 
\begin{tabular}{l l c ccc c c}
    \toprule
    Model & \multirow{2}{*}{\makecell{Concept\\Type}} & \multicolumn{6}{c}{Concept Detection Methods} \\
    \cmidrule(lr){3-8}
     &  & RandTok & LastTok & MeanTok & CLSTok & Prompt & \textbf{SuperAct} \\
    \midrule
\multirow{4}{*}{CLIP} & Avg & {\tablenumbersize 0.360 {\scriptsize $\pm$ 0.013}} & {\tablenumbersize 0.354 {\scriptsize $\pm$ 0.012}} & {\tablenumbersize 0.465 {\scriptsize $\pm$ 0.014}} & \underline{{\tablenumbersize 0.488 {\scriptsize $\pm$ 0.014}}} & -- & \textbf{{\tablenumbersize 0.546 {\scriptsize $\pm$ 0.015}}} \\
 & Linsep & {\tablenumbersize 0.416 {\scriptsize $\pm$ 0.014}} & {\tablenumbersize 0.413 {\scriptsize $\pm$ 0.013}} & \underline{{\tablenumbersize 0.523 {\scriptsize $\pm$ 0.014}}} & {\tablenumbersize 0.499 {\scriptsize $\pm$ 0.014}} & -- & \textbf{{\tablenumbersize 0.557 {\scriptsize $\pm$ 0.014}}} \\
 & K-Means & {\tablenumbersize 0.340 {\scriptsize $\pm$ 0.013}} & {\tablenumbersize 0.339 {\scriptsize $\pm$ 0.013}} & \underline{{\tablenumbersize 0.469 {\scriptsize $\pm$ 0.014}}} & {\tablenumbersize 0.437 {\scriptsize $\pm$ 0.014}} & -- & \textbf{{\tablenumbersize 0.533 {\scriptsize $\pm$ 0.015}}} \\
 & K-Linsep & {\tablenumbersize 0.348 {\scriptsize $\pm$ 0.013}} & {\tablenumbersize 0.345 {\scriptsize $\pm$ 0.013}} & \underline{{\tablenumbersize 0.481 {\scriptsize $\pm$ 0.015}}} & {\tablenumbersize 0.432 {\scriptsize $\pm$ 0.014}} & -- & \textbf{{\tablenumbersize 0.530 {\scriptsize $\pm$ 0.015}}} \\
\midrule
\multirow{4}{*}{Llama} & Avg & {\tablenumbersize 0.352 {\scriptsize $\pm$ 0.012}} & {\tablenumbersize 0.332 {\scriptsize $\pm$ 0.011}} & {\tablenumbersize 0.480 {\scriptsize $\pm$ 0.014}} & {\tablenumbersize 0.439 {\scriptsize $\pm$ 0.014}} & \underline{{\tablenumbersize 0.491 {\scriptsize $\pm$ 0.014}}} & \textbf{{\tablenumbersize 0.533 {\scriptsize $\pm$ 0.014}}} \\
\bottomrule
\end{tabular}
}
\label{tab:opensurfaces-detection}

\end{table}

\begin{table}[h]
\centering
% Broden-Pascal
% Match to Broden-Pascal source style.
% Requires booktabs, multirow, makecell, tabletextsize, tablenumbersize.
\centering
\setlength{\tabcolsep}{4.7pt}
\caption{Concept detection $F_1$ for the \textbf{Broden-Pascal} dataset.}
{\tabletextsize 
\begin{tabular}{l l c ccc c c}
    \toprule
    Model & \multirow{2}{*}{\makecell{Concept\\Type}} & \multicolumn{6}{c}{Concept Detection Methods} \\
    \cmidrule(lr){3-8}
     &  & RandTok & LastTok & MeanTok & CLSTok & Prompt & \textbf{SuperAct} \\
    \midrule
\multirow{4}{*}{CLIP} & Avg & {\tablenumbersize 0.426 {\scriptsize $\pm$ 0.007}} & {\tablenumbersize 0.420 {\scriptsize $\pm$ 0.006}} & {\tablenumbersize 0.581 {\scriptsize $\pm$ 0.007}} & \underline{{\tablenumbersize 0.720 {\scriptsize $\pm$ 0.006}}} & -- & \textbf{{\tablenumbersize 0.792 {\scriptsize $\pm$ 0.006}}} \\
 & Linsep & {\tablenumbersize 0.497 {\scriptsize $\pm$ 0.007}} & {\tablenumbersize 0.492 {\scriptsize $\pm$ 0.006}} & {\tablenumbersize 0.698 {\scriptsize $\pm$ 0.006}} & \underline{{\tablenumbersize 0.746 {\scriptsize $\pm$ 0.006}}} & -- & \textbf{{\tablenumbersize 0.824 {\scriptsize $\pm$ 0.005}}} \\
 & K-Means & {\tablenumbersize 0.423 {\scriptsize $\pm$ 0.007}} & {\tablenumbersize 0.443 {\scriptsize $\pm$ 0.007}} & {\tablenumbersize 0.614 {\scriptsize $\pm$ 0.007}} & \underline{{\tablenumbersize 0.652 {\scriptsize $\pm$ 0.006}}} & -- & \textbf{{\tablenumbersize 0.786 {\scriptsize $\pm$ 0.006}}} \\
 & K-Linsep & {\tablenumbersize 0.414 {\scriptsize $\pm$ 0.007}} & {\tablenumbersize 0.450 {\scriptsize $\pm$ 0.007}} & \underline{{\tablenumbersize 0.648 {\scriptsize $\pm$ 0.007}}} & {\tablenumbersize 0.636 {\scriptsize $\pm$ 0.007}} & -- & \textbf{{\tablenumbersize 0.774 {\scriptsize $\pm$ 0.006}}} \\
\midrule
\multirow{4}{*}{Llama} & Avg & {\tablenumbersize 0.403 {\scriptsize $\pm$ 0.006}} & {\tablenumbersize 0.393 {\scriptsize $\pm$ 0.006}} & {\tablenumbersize 0.624 {\scriptsize $\pm$ 0.007}} & {\tablenumbersize 0.622 {\scriptsize $\pm$ 0.007}} & \underline{{\tablenumbersize 0.680 {\scriptsize $\pm$ 0.006}}} & \textbf{{\tablenumbersize 0.786 {\scriptsize $\pm$ 0.006}}} \\
\bottomrule
\end{tabular}
}
\label{tab:pascal-detection}

\end{table}

\begin{table}[h]
\centering
% PERSUADE
% Match to Broden-Pascal source style.
% Requires booktabs, multirow, makecell, tabletextsize, tablenumbersize.
\centering
\setlength{\tabcolsep}{4.7pt}
\caption{Concept detection $F_1$ for the \textbf{PERSUADE} dataset.}
{\tabletextsize 
\begin{tabular}{l l c ccc c c}
    \toprule
    Model & \multirow{2}{*}{\makecell{Concept\\Type}} & \multicolumn{6}{c}{Concept Detection Methods} \\
    \cmidrule(lr){3-8}
     &  & RandTok & LastTok & MeanTok & CLSTok & Prompt & \textbf{SuperAct} \\
    \midrule
\multirow{4}{*}{Llama} & Avg & {\tablenumbersize 0.412 {\scriptsize $\pm$ 0.020}} & {\tablenumbersize 0.401 {\scriptsize $\pm$ 0.020}} & {\tablenumbersize 0.484 {\scriptsize $\pm$ 0.024}} & \underline{{\tablenumbersize 0.486 {\scriptsize $\pm$ 0.026}}} & {\tablenumbersize 0.286 {\scriptsize $\pm$ 0.026}} & \textbf{{\tablenumbersize 0.591 {\scriptsize $\pm$ 0.025}}} \\
 & Linsep & {\tablenumbersize 0.422 {\scriptsize $\pm$ 0.024}} & {\tablenumbersize 0.419 {\scriptsize $\pm$ 0.022}} & {\tablenumbersize 0.510 {\scriptsize $\pm$ 0.025}} & \underline{{\tablenumbersize 0.551 {\scriptsize $\pm$ 0.027}}} & {\tablenumbersize 0.286 {\scriptsize $\pm$ 0.026}} & \textbf{{\tablenumbersize 0.639 {\scriptsize $\pm$ 0.026}}} \\
 & K-Means & {\tablenumbersize 0.398 {\scriptsize $\pm$ 0.020}} & {\tablenumbersize 0.400 {\scriptsize $\pm$ 0.021}} & \underline{{\tablenumbersize 0.434 {\scriptsize $\pm$ 0.021}}} & {\tablenumbersize 0.421 {\scriptsize $\pm$ 0.021}} & {\tablenumbersize 0.286 {\scriptsize $\pm$ 0.026}} & \textbf{{\tablenumbersize 0.605 {\scriptsize $\pm$ 0.024}}} \\
 & K-Linsep & {\tablenumbersize 0.396 {\scriptsize $\pm$ 0.020}} & {\tablenumbersize 0.405 {\scriptsize $\pm$ 0.020}} & \underline{{\tablenumbersize 0.445 {\scriptsize $\pm$ 0.024}}} & {\tablenumbersize 0.421 {\scriptsize $\pm$ 0.021}} & {\tablenumbersize 0.286 {\scriptsize $\pm$ 0.026}} & \textbf{{\tablenumbersize 0.608 {\scriptsize $\pm$ 0.026}}} \\
\bottomrule
\end{tabular}
}
\label{tab:persuade-detection}

\end{table}

\begin{table}[h]
\centering
% El Capitan
% Match to Broden-Pascal source style.
% Requires booktabs, multirow, makecell, tabletextsize, tablenumbersize.
\centering
\setlength{\tabcolsep}{4.7pt}
\caption{Concept detection $F_1$ for the \textbf{El Capitan} dataset.}
{\tabletextsize 
\begin{tabular}{l l c ccc c c}
    \toprule
    Model & \multirow{2}{*}{\makecell{Concept\\Type}} & \multicolumn{6}{c}{Concept Detection Methods} \\
    \cmidrule(lr){3-8}
     &  & RandTok & LastTok & MeanTok & CLSTok & Prompt & \textbf{SuperAct} \\
    \midrule
\multirow{4}{*}{Llama} & Avg & {\tablenumbersize 0.430 {\scriptsize $\pm$ 0.023}} & {\tablenumbersize 0.434 {\scriptsize $\pm$ 0.023}} & {\tablenumbersize 0.505 {\scriptsize $\pm$ 0.024}} & {\tablenumbersize 0.503 {\scriptsize $\pm$ 0.026}} & \underline{{\tablenumbersize 0.512 {\scriptsize $\pm$ 0.029}}} & \textbf{{\tablenumbersize 0.525 {\scriptsize $\pm$ 0.023}}} \\
 & Linsep & {\tablenumbersize 0.471 {\scriptsize $\pm$ 0.025}} & {\tablenumbersize 0.452 {\scriptsize $\pm$ 0.023}} & {\tablenumbersize 0.524 {\scriptsize $\pm$ 0.025}} & \underline{{\tablenumbersize 0.580 {\scriptsize $\pm$ 0.026}}} & {\tablenumbersize 0.512 {\scriptsize $\pm$ 0.029}} & \textbf{{\tablenumbersize 0.696 {\scriptsize $\pm$ 0.022}}} \\
 & K-Means & {\tablenumbersize 0.407 {\scriptsize $\pm$ 0.024}} & {\tablenumbersize 0.437 {\scriptsize $\pm$ 0.024}} & \underline{{\tablenumbersize 0.569 {\scriptsize $\pm$ 0.026}}} & {\tablenumbersize 0.494 {\scriptsize $\pm$ 0.026}} & {\tablenumbersize 0.512 {\scriptsize $\pm$ 0.029}} & \textbf{{\tablenumbersize 0.682 {\scriptsize $\pm$ 0.022}}} \\
 & K-Linsep & {\tablenumbersize 0.397 {\scriptsize $\pm$ 0.024}} & {\tablenumbersize 0.446 {\scriptsize $\pm$ 0.024}} & \underline{{\tablenumbersize 0.540 {\scriptsize $\pm$ 0.027}}} & {\tablenumbersize 0.494 {\scriptsize $\pm$ 0.026}} & {\tablenumbersize 0.512 {\scriptsize $\pm$ 0.029}} & \textbf{{\tablenumbersize 0.679 {\scriptsize $\pm$ 0.022}}} \\
\bottomrule
\end{tabular}
}
\label{tab:el_capitan-detection}

\end{table}

\begin{table}[h]
\centering
% GoEmotions
% Match to Broden-Pascal source style.
% Requires booktabs, multirow, makecell, tabletextsize, tablenumbersize.
\centering
\setlength{\tabcolsep}{4.7pt}
\caption{Concept detection $F_1$ for the \textbf{GoEmotions} dataset.}
{\tabletextsize 
\begin{tabular}{l l c ccc c c}
    \toprule
    Model & \multirow{2}{*}{\makecell{Concept\\Type}} & \multicolumn{6}{c}{Concept Detection Methods} \\
    \cmidrule(lr){3-8}
     &  & RandTok & LastTok & MeanTok & CLSTok & Prompt & \textbf{SuperAct} \\
    \midrule
\multirow{4}{*}{Llama} & Avg & {\tablenumbersize 0.131 {\scriptsize $\pm$ 0.021}} & {\tablenumbersize 0.172 {\scriptsize $\pm$ 0.021}} & {\tablenumbersize 0.177 {\scriptsize $\pm$ 0.021}} & \underline{{\tablenumbersize 0.274 {\scriptsize $\pm$ 0.026}}} & {\tablenumbersize 0.252 {\scriptsize $\pm$ 0.028}} & \textbf{{\tablenumbersize 0.330 {\scriptsize $\pm$ 0.026}}} \\
 & Linsep & {\tablenumbersize 0.187 {\scriptsize $\pm$ 0.026}} & {\tablenumbersize 0.210 {\scriptsize $\pm$ 0.023}} & {\tablenumbersize 0.275 {\scriptsize $\pm$ 0.025}} & \underline{{\tablenumbersize 0.348 {\scriptsize $\pm$ 0.029}}} & {\tablenumbersize 0.252 {\scriptsize $\pm$ 0.028}} & \textbf{{\tablenumbersize 0.449 {\scriptsize $\pm$ 0.027}}} \\
 & K-Means & {\tablenumbersize 0.083 {\scriptsize $\pm$ 0.017}} & {\tablenumbersize 0.156 {\scriptsize $\pm$ 0.022}} & {\tablenumbersize 0.175 {\scriptsize $\pm$ 0.021}} & {\tablenumbersize 0.165 {\scriptsize $\pm$ 0.022}} & \underline{{\tablenumbersize 0.252 {\scriptsize $\pm$ 0.028}}} & \textbf{{\tablenumbersize 0.387 {\scriptsize $\pm$ 0.026}}} \\
 & K-Linsep & {\tablenumbersize 0.122 {\scriptsize $\pm$ 0.022}} & {\tablenumbersize 0.171 {\scriptsize $\pm$ 0.023}} & \underline{{\tablenumbersize 0.305 {\scriptsize $\pm$ 0.029}}} & {\tablenumbersize 0.165 {\scriptsize $\pm$ 0.021}} & {\tablenumbersize 0.252 {\scriptsize $\pm$ 0.028}} & \textbf{{\tablenumbersize 0.452 {\scriptsize $\pm$ 0.028}}} \\
\bottomrule
\end{tabular}
}
\label{tab:goemotions-detection}

\end{table}

\begin{table}[h]
\centering
% CLEVR
% Match to Broden-Pascal source style.
% Requires booktabs, multirow, makecell, tabletextsize, tablenumbersize.
\centering
\setlength{\tabcolsep}{4.7pt}
\caption{Concept detection $F_1$ for the \textbf{CLEVR} dataset.}
{\tabletextsize 
\begin{tabular}{l l c ccc c c}
    \toprule
    Model & \multirow{2}{*}{\makecell{Concept\\Type}} & \multicolumn{6}{c}{Concept Detection Methods} \\
    \cmidrule(lr){3-8}
     &  & RandTok & LastTok & MeanTok & CLSTok & Prompt & \textbf{SuperAct} \\
    \midrule
\multirow{4}{*}{CLIP} & Avg & {\tablenumbersize 0.479 {\scriptsize $\pm$ 0.035}} & {\tablenumbersize 0.484 {\scriptsize $\pm$ 0.036}} & {\tablenumbersize 0.566 {\scriptsize $\pm$ 0.037}} & \underline{{\tablenumbersize 0.983 {\scriptsize $\pm$ 0.010}}} & -- & \textbf{{\tablenumbersize 0.989 {\scriptsize $\pm$ 0.008}}} \\
 & Linsep & {\tablenumbersize 0.475 {\scriptsize $\pm$ 0.035}} & {\tablenumbersize 0.527 {\scriptsize $\pm$ 0.040}} & \underline{{\tablenumbersize 0.805 {\scriptsize $\pm$ 0.030}}} & \textbf{{\tablenumbersize 0.994 {\scriptsize $\pm$ 0.005}}} & -- & \textbf{{\tablenumbersize 0.994 {\scriptsize $\pm$ 0.006}}} \\
 & K-Means & {\tablenumbersize 0.447 {\scriptsize $\pm$ 0.038}} & {\tablenumbersize 0.578 {\scriptsize $\pm$ 0.041}} & {\tablenumbersize 0.670 {\scriptsize $\pm$ 0.042}} & \underline{{\tablenumbersize 0.957 {\scriptsize $\pm$ 0.016}}} & -- & \textbf{{\tablenumbersize 0.993 {\scriptsize $\pm$ 0.006}}} \\
 & K-Linsep & {\tablenumbersize 0.448 {\scriptsize $\pm$ 0.039}} & {\tablenumbersize 0.613 {\scriptsize $\pm$ 0.041}} & {\tablenumbersize 0.779 {\scriptsize $\pm$ 0.035}} & \textbf{{\tablenumbersize 0.980 {\scriptsize $\pm$ 0.011}}} & -- & \underline{{\tablenumbersize 0.976 {\scriptsize $\pm$ 0.012}}} \\
\midrule
\multirow{4}{*}{Llama} & Avg & {\tablenumbersize 0.483 {\scriptsize $\pm$ 0.037}} & {\tablenumbersize 0.532 {\scriptsize $\pm$ 0.038}} & {\tablenumbersize 0.774 {\scriptsize $\pm$ 0.028}} & {\tablenumbersize 0.960 {\scriptsize $\pm$ 0.015}} & \underline{{\tablenumbersize 0.987 {\scriptsize $\pm$ 0.008}}} & \textbf{{\tablenumbersize 1.000 {\scriptsize $\pm$ 0.000}}} \\
\bottomrule
\end{tabular}
}
\label{tab:clevr-detection}

\end{table}

\begin{table}[h]
\centering
% iSarcasm
% Match to Broden-Pascal source style.
% Requires booktabs, multirow, makecell, tabletextsize, tablenumbersize.
\centering
\setlength{\tabcolsep}{4.7pt}
\caption{Concept detection $F_1$ for the \textbf{iSarcasm} dataset.}
{\tabletextsize 
\begin{tabular}{l l c ccc c c}
    \toprule
    Model & \multirow{2}{*}{\makecell{Concept\\Type}} & \multicolumn{6}{c}{Concept Detection Methods} \\
    \cmidrule(lr){3-8}
     &  & RandTok & LastTok & MeanTok & CLSTok & Prompt & \textbf{SuperAct} \\
    \midrule
\multirow{4}{*}{Llama} & Avg & {\tablenumbersize 0.679 {\scriptsize $\pm$ 0.047}} & \underline{{\tablenumbersize 0.823 {\scriptsize $\pm$ 0.045}}} & {\tablenumbersize 0.712 {\scriptsize $\pm$ 0.050}} & \textbf{{\tablenumbersize 0.860 {\scriptsize $\pm$ 0.038}}} & {\tablenumbersize 0.789 {\scriptsize $\pm$ 0.047}} & {\tablenumbersize 0.817 {\scriptsize $\pm$ 0.044}} \\
 & Linsep & {\tablenumbersize 0.708 {\scriptsize $\pm$ 0.054}} & \underline{{\tablenumbersize 0.905 {\scriptsize $\pm$ 0.033}}} & {\tablenumbersize 0.859 {\scriptsize $\pm$ 0.039}} & {\tablenumbersize 0.892 {\scriptsize $\pm$ 0.035}} & {\tablenumbersize 0.789 {\scriptsize $\pm$ 0.047}} & \textbf{{\tablenumbersize 0.918 {\scriptsize $\pm$ 0.029}}} \\
 & K-Means & {\tablenumbersize 0.678 {\scriptsize $\pm$ 0.050}} & \textbf{{\tablenumbersize 0.820 {\scriptsize $\pm$ 0.044}}} & {\tablenumbersize 0.791 {\scriptsize $\pm$ 0.045}} & \underline{{\tablenumbersize 0.808 {\scriptsize $\pm$ 0.043}}} & {\tablenumbersize 0.789 {\scriptsize $\pm$ 0.047}} & {\tablenumbersize 0.754 {\scriptsize $\pm$ 0.050}} \\
 & K-Linsep & {\tablenumbersize 0.648 {\scriptsize $\pm$ 0.053}} & \textbf{{\tablenumbersize 0.870 {\scriptsize $\pm$ 0.037}}} & {\tablenumbersize 0.815 {\scriptsize $\pm$ 0.044}} & {\tablenumbersize 0.810 {\scriptsize $\pm$ 0.046}} & {\tablenumbersize 0.789 {\scriptsize $\pm$ 0.047}} & \underline{{\tablenumbersize 0.867 {\scriptsize $\pm$ 0.037}}} \\
\bottomrule
\end{tabular}
}
\label{tab:isarcasm-detection}

\end{table}

\clearpage

\section{Ablation: How Does Sparsity Affect Average \super{} Detection Performance?}
\label{app:sparsity-ablation}

In this section, we evaluate how concept detection performance varies with the in-concept tail fraction $\delta$. Smaller $\delta$ restricts detection to increasingly extreme activations, while larger $\delta$ admits progressively more of the distribution (at $\delta=100\%$, the entire in-concept distribution is admitted, so strong detection would imply near-complete separation from out-of-concept activations). For each value of $\delta$, we use the same \super{}-based aggregation rule and evaluate whether the maximum token activation exceeds the corresponding threshold. Reported $F_1$ values are averages of the per-concept detection $F_1$, weighted by concept frequency and evaluated at each concept's best-performing layer on the validation set.

Across all model-dataset combinations, detection generally performs best at small in-concept tail fractions (typically between 5-10\%) and drops sharply as $\delta$ increases. Thus, the strongest concept signal is highly concentrated in the \super{} tail rather than extending broadly throughout the distribution.

\begin{figure}[h]
    \centering
    \begin{subfigure}{\textwidth}
        \centering
        \includegraphics[width=\linewidth]{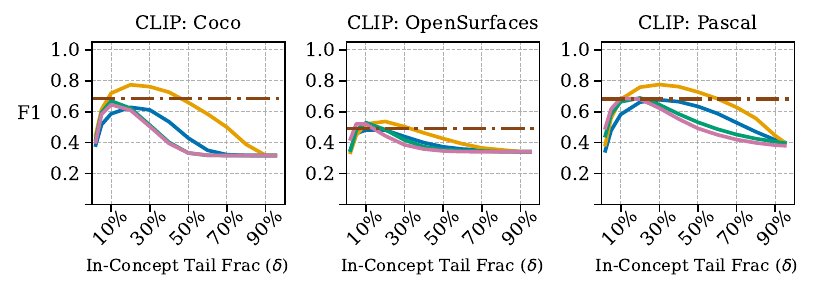}
    \end{subfigure}
    \begin{subfigure}{\textwidth}
        \centering
        \includegraphics[width=\linewidth]{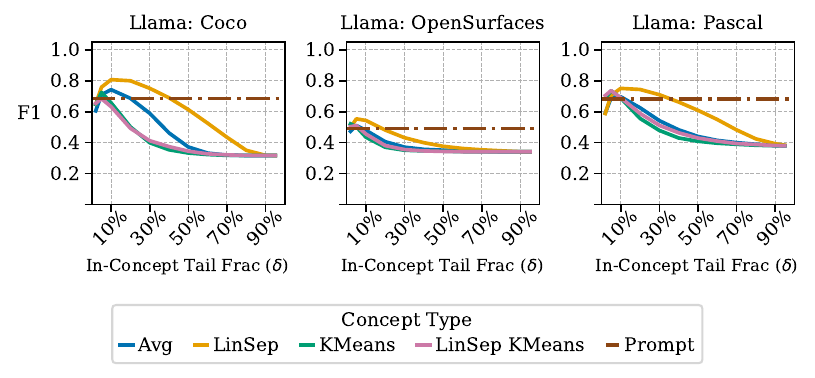}
    \end{subfigure}
    \caption{Image Domain -- Detection $F_1$ over Sparsity Level $\delta$}
\end{figure}

\begin{figure}[h]
    \centering
    \begin{subfigure}{\textwidth}
        \centering
        \includegraphics[width=\linewidth]{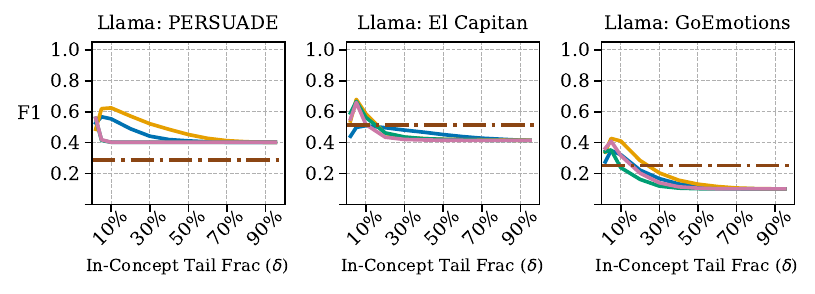}
    \end{subfigure}
    \begin{subfigure}{\textwidth}
        \centering
        \includegraphics[width=\linewidth]{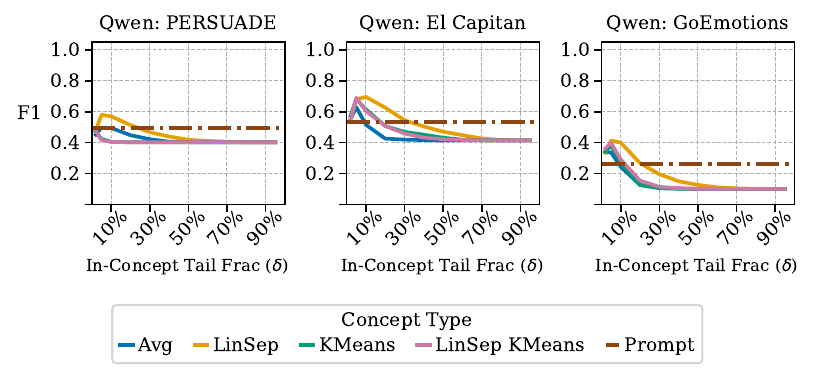}
    \end{subfigure}
    \caption{Text Domain -- Detection $F_1$ over Sparsity Level $\delta$}
\end{figure}

\clearpage

\section{Ablation: How Does Optimal Sparsity for \super{} Detection Vary Across Model Layers?}
\label{app:optimal-sparsity-across-layers}
\begin{figure}[h!]
    \centering

    % ===== Middle subfigure =====
    \begin{subfigure}{\textwidth}
        \centering
        \includegraphics[width=\textwidth]{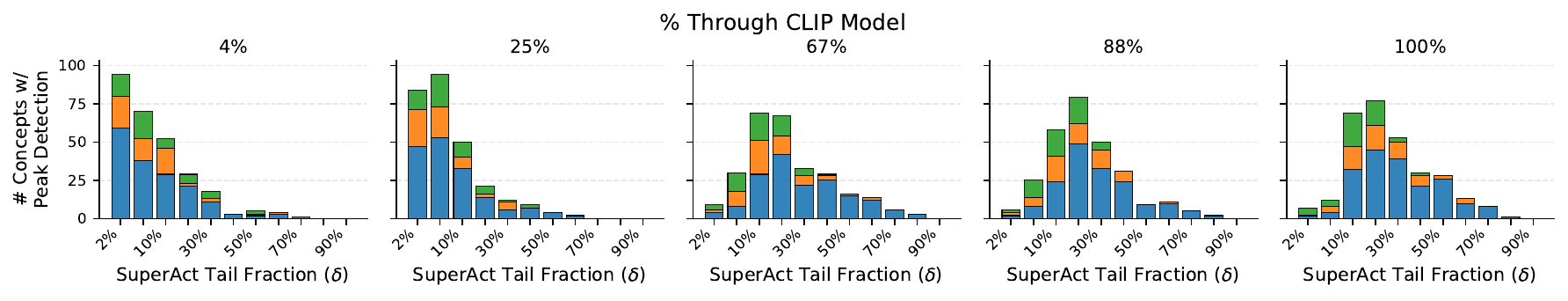}
    \end{subfigure}

    \begin{subfigure}{\textwidth}
        \centering
        \includegraphics[width=\textwidth]{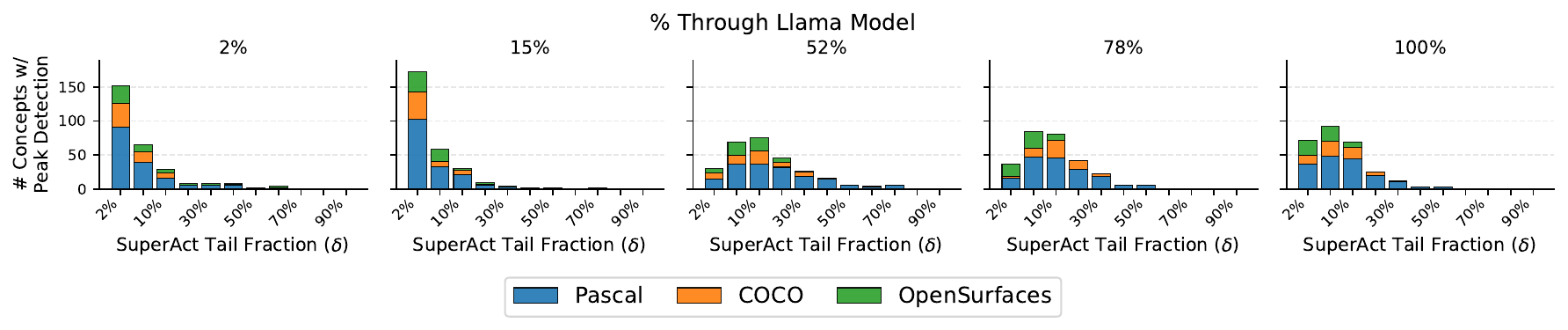}
    \end{subfigure}

    \caption{Image Domain -- Optimal Sparsity over Layers}
    \label{fig:sparsity-across-layers-image}
\end{figure}

\begin{figure}[h!]
    \centering
    \begin{subfigure}{\textwidth}
        \centering
        \includegraphics[width=\textwidth]{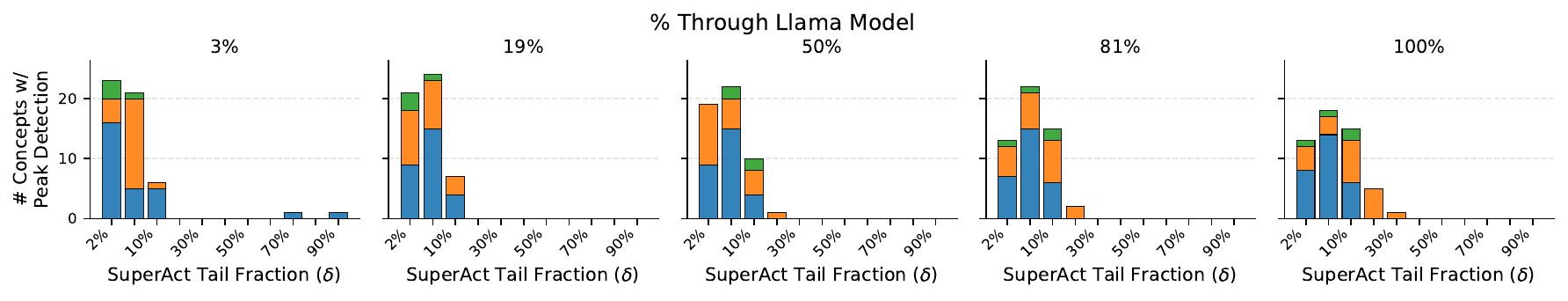}
    \end{subfigure}

    % ===== Top subfigure =====
    \begin{subfigure}{\textwidth}
        \centering
        \includegraphics[width=\textwidth]{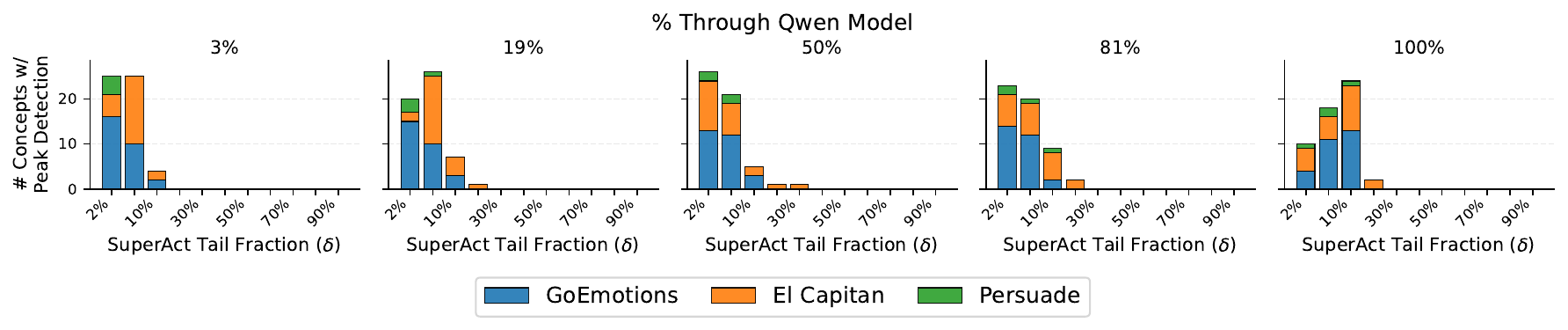}
    \end{subfigure}

    \caption{Text Domain -- Optimal Sparsity over Layers}
    \label{fig:sparsity-across-layers-text}
\end{figure}

Appendix~\ref{app:sparsity-ablation} showed that detection performance peaks at small in-concept tail fractions when each concept is evaluated at its best layer. Here we ask whether the optimal tail sparsity itself depends on depth. For each layer and each concept, we sweep the tail fraction $\delta$ and record the value that maximizes validation detection $F_1$. Figures~\ref{fig:sparsity-across-layers-image} and \ref{fig:sparsity-across-layers-text} show, for each layer, the distribution of this optimal $\delta$ across concepts, stacked by dataset.

At every layer and in both domains, most concepts peak at $\delta \leq 30\%$ and almost none at $\delta \geq 50\%$. The optimum shifts only modestly with depth, from the smallest tail fractions in early layers toward $10$--$30\%$ (image) or $10\%$ (text) in
later layers. Thus the most reliable concept evidence lies in the sparse tail regardless of depth.

\clearpage
\section{Ablation: How many \supers{} do most samples have?}
\label{app:superdetector-cdfs}
\begin{figure}[h!]
    \centering

    % Row 1
    \begin{subfigure}{0.41\textwidth}
        \centering
        \includegraphics[width=\linewidth]{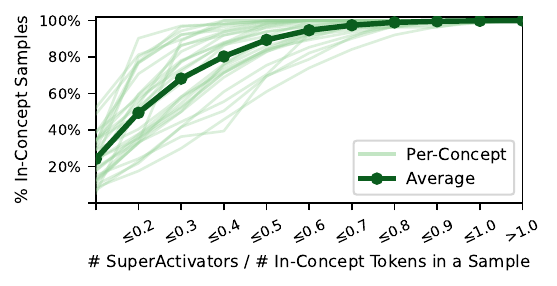}
        \vspace{-1.75em}
        \caption{COCO}
    \end{subfigure}
    \hspace{0.5em}
    \begin{subfigure}{0.41\textwidth}
        \centering
        \includegraphics[width=\linewidth]{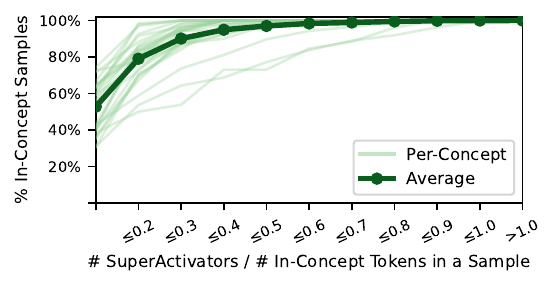}
        \vspace{-1.75em}
        \caption{OpenSurfaces}
    \end{subfigure}

    \vspace{0.75em}

    % Row 2
    \begin{subfigure}{0.41\textwidth}
        \centering
        \includegraphics[width=\linewidth]{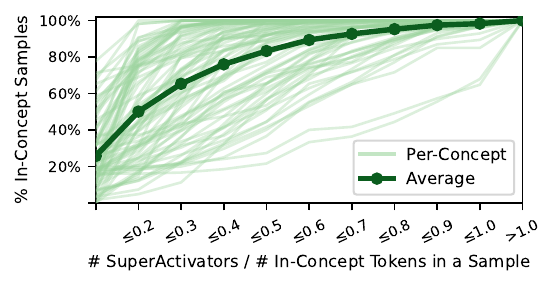}
        \vspace{-1.75em}
        \caption{Pascal}
    \end{subfigure}
    \hspace{0.5em}
    \begin{subfigure}{0.41\textwidth}
        \centering
        \includegraphics[width=\linewidth]{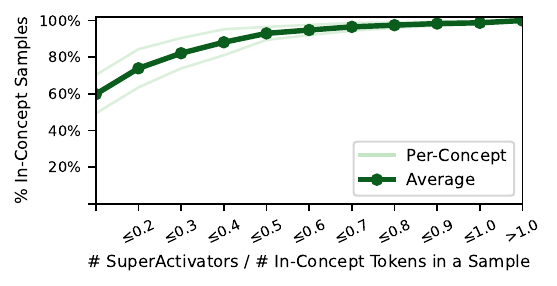}
        \vspace{-1.75em}
        \caption{PERSUADE}
    \end{subfigure}
    
    \vspace{0.75em}

    \begin{subfigure}{0.41\textwidth}
        \centering
        \includegraphics[width=\linewidth]{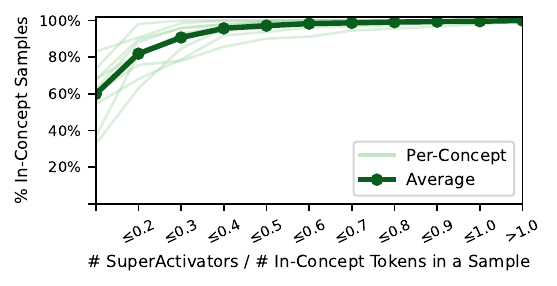}
        \vspace{-1.75em}
        \caption{El Capitan}
    \end{subfigure}
    \hspace{0.5em}
    \begin{subfigure}{0.41\textwidth}
        \centering
        \includegraphics[width=\linewidth]{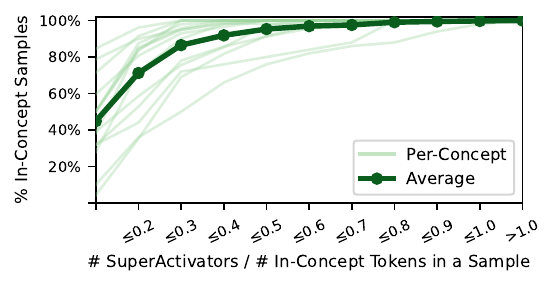}
        \vspace{-1.75em}
        \caption{GoEmotions}
    \end{subfigure}

    \vspace{0.75em}

    \caption{Cumulative distribution functions showing, for each concept and on average across a dataset, the ratio of \supers{} to in-concept tokens in each in-concept test sample.}
    \label{fig:superactivator-cdfs}
\end{figure}

Figure \ref{fig:superactivator-cdfs} shows cumulative distribution functions for the \emph{Llama-3.2-11B-Vision-Instruct} linear-separator concepts, using each concept’s validation-selected model layer and optimal sparsity level $\delta$ on the test set. For each in-concept sample, we plot the ratio of \supers{} in the sample to the number of in-concept tokens, which normalizes for varying concept-span lengths and allows \supers{} to appear anywhere in the sequence.

These plots indicate that most in-concept samples contain only a small set of reliable concept signals relative to the total number of in-concept tokens. Across all datasets, the majority of in-concept samples contain no more than roughly one \super{} for every five in-concept tokens.
 
% In your document:

\clearpage
\section{Ablation: How does concept detection performance vary with depth?}
\label{app:detection-results-across-layers}
\begin{figure}[h!]
\centering
\includegraphics[width=\textwidth]{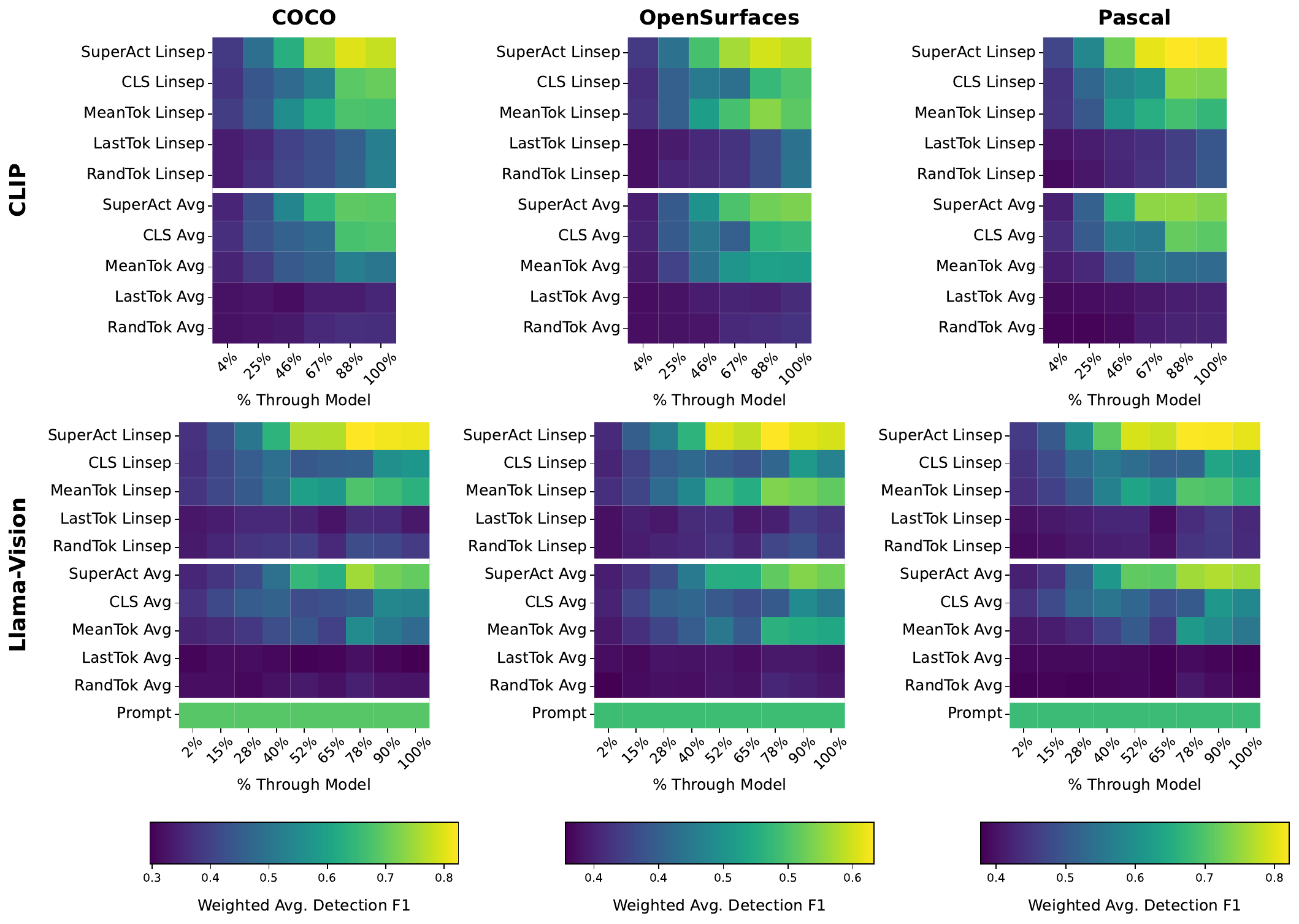}
\caption{\super{} detection across image datasets.}
\label{fig:detection-across-model-image}
\end{figure}

\begin{figure}[h!]
\centering
\includegraphics[width=\textwidth]{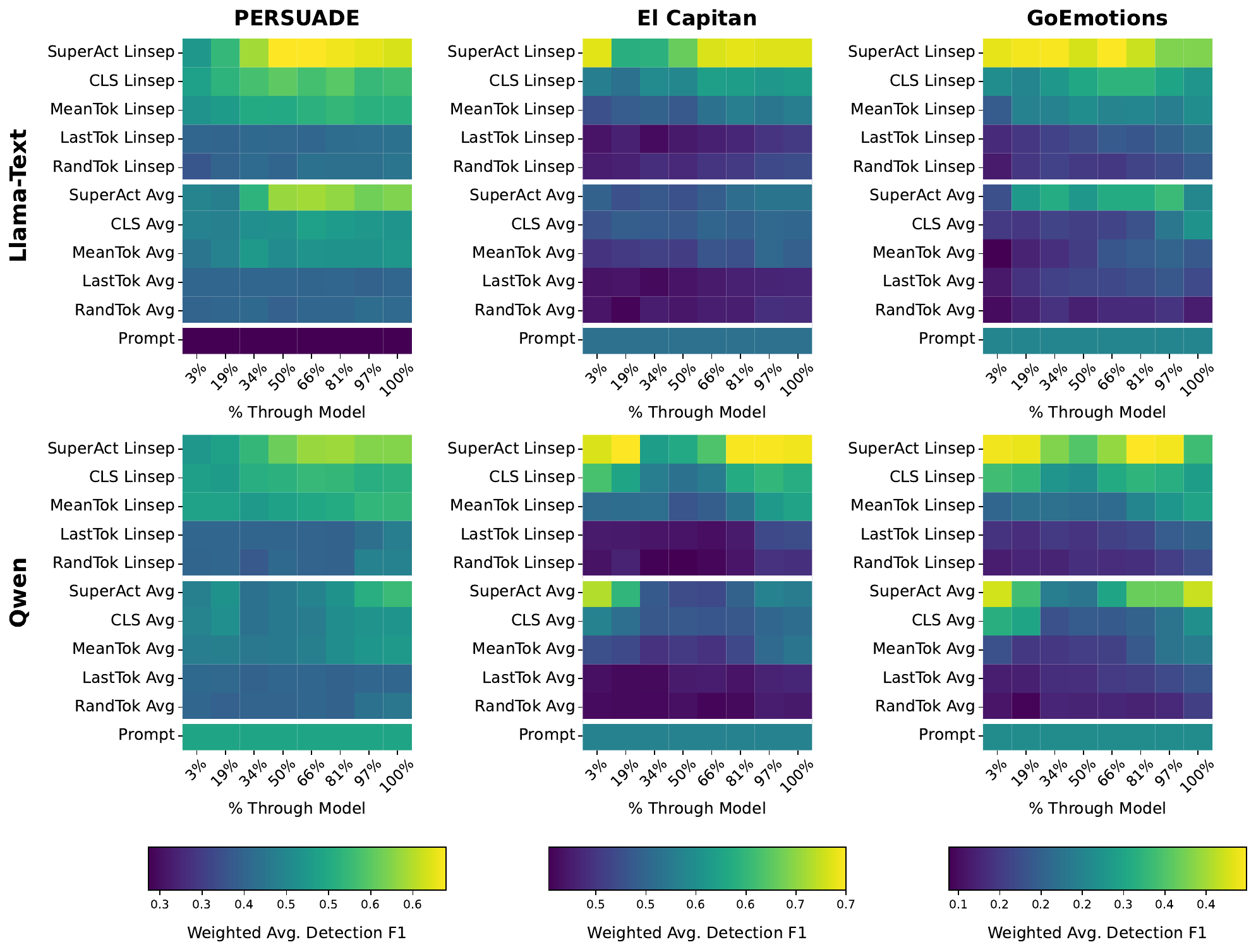}
\caption{\super{} detection across text datasets.}
\label{fig:detection-across-model-text}
\end{figure}

In this section, we investigate how average concept detection performance evolves throughout model depth. Figures~\ref{fig:detection-across-model-image} and~\ref{fig:detection-across-model-text} visualize heatmaps of the average detection $F_{1}$ scores as a function of transformer layer depth for image and text datasets, respectively. Each heatmap reports the mean $F_{1}$ score for each dataset, model, concept type, and detection scheme, computed over a grid of model depths. These heatmaps help illustrate how concept signals emerge and strengthen at different stages within the network.

In the vision domain, detection performance generally increases with depth, plateauing around the middle layers and declining slightly at the final layer. This behavior aligns with findings from prior work~\citep{Salam2025LargeLM, Yu2024LatentCE, Dalvi2022DiscoveringLC}, which report that mid-level and late-level layers often capture the richest and most separable semantic information. Text models show a broadly similar progression, though with more variability: some datasets and concepts exhibit earlier local peaks, but performance generally improves again toward the middle and later layers.

\clearpage

\section{Ablation: Which Model Layers Yield the Most Separable Concepts?}
\label{app:best-layer-per-concept}
In this section, we identify where in the model concepts are most separable, that is, at which layers concept vectors achieve their highest detection performance. For each dataset, we plot the frequency of concept vectors that achieve their best $F_{1}$ detection scores at each model layer. These trends are shown for the \super{}-based detection scheme as well as for [CLS]-, mean-, and last-token--based detection methods. All results in this analysis use linear separator concept vectors derived from the \emph{Llama-3.2-11B-Vision-Instruct} model. 

Across the image datasets, concept vectors most often achieve their best detection performance in later layers, with peaks concentrated around roughly $78$--$90\%$ through the model. A similar overall pattern appears for the text datasets, although their optimal layers are more dispersed and some concepts peak substantially earlier. Overall, the most separable concept representations tend to occur in the later portions of the model.

\begin{figure}[h!]
    \centering

    % -------- Row 1 --------
    \begin{subfigure}{0.49\textwidth}
        \centering
        \includegraphics[width=\textwidth]{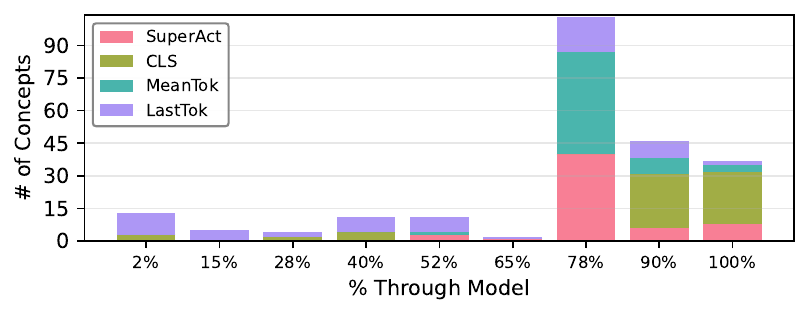}
        \vspace{-1.5em}
        \caption{\emph{COCO}}
    \end{subfigure}
    \hfill
    \begin{subfigure}{0.49\textwidth}
        \centering
        \includegraphics[width=\textwidth]{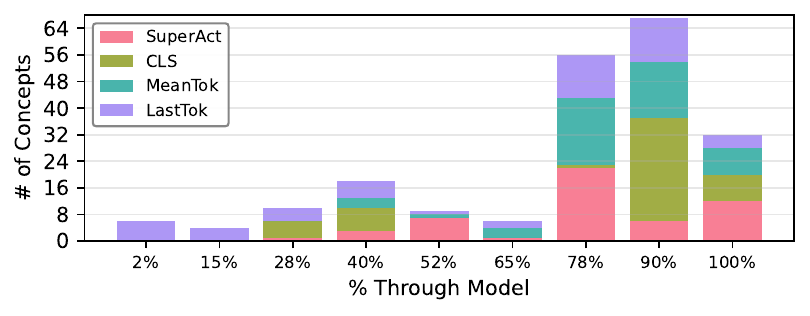}
        \vspace{-1.5em}
        \caption{\emph{Broden-OpenSurfaces}}
    \end{subfigure}

    \vspace{0.5em}

    % -------- Row 2 --------
    \begin{subfigure}{0.49\textwidth}
        \centering
        \includegraphics[width=\textwidth]{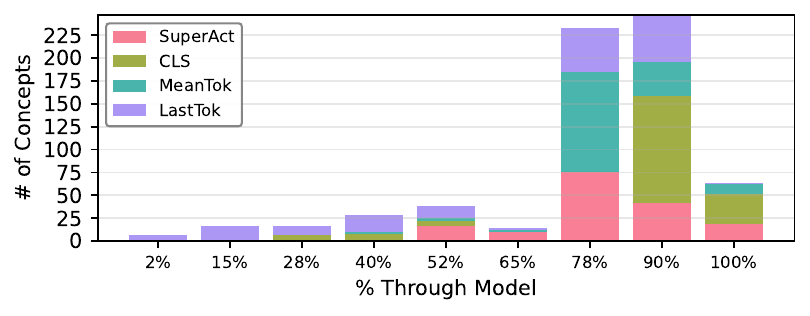}
        \vspace{-1.5em}
        \caption{\emph{Broden-Pascal}}
    \end{subfigure}
    \hfill
    \begin{subfigure}{0.49\textwidth}
        \centering
        \includegraphics[width=\textwidth]{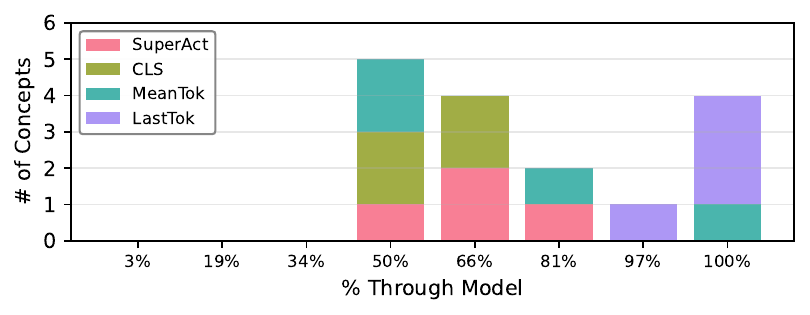}
        \vspace{-1.5em}
        \caption{\emph{PERSUADE}}
    \end{subfigure}

    \vspace{0.5em}

    \begin{subfigure}{0.49\textwidth}
        \centering
        \includegraphics[width=\textwidth]{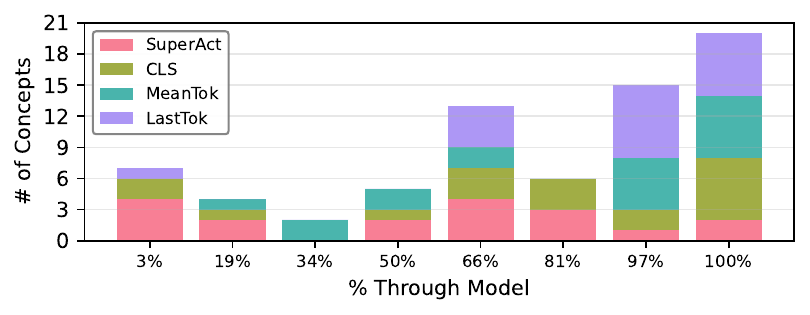}
        \vspace{-1.5em}
        \caption{\emph{El Capitan}}
    \end{subfigure}\hfill
    \begin{subfigure}{0.49\textwidth}
        \centering
        \includegraphics[width=\textwidth]{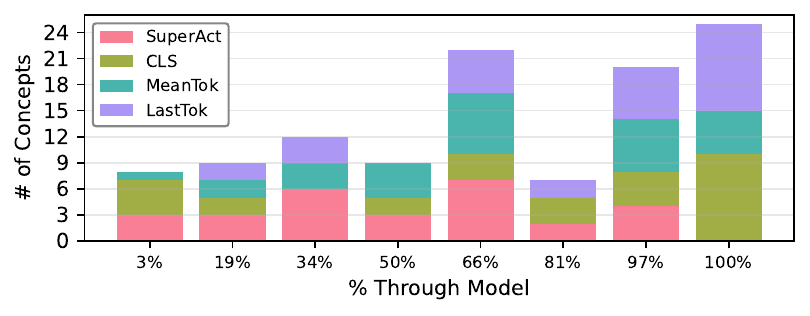}
        \vspace{-1.5em}
        \caption{\emph{GoEmotions}}
    \end{subfigure}

    \vspace{0.5em}

    \caption{Distribution of Optimal Detection Layers Across Aggregation Methods/Datasets}
    \label{fig:ptm_distributions}
\end{figure}

\clearpage
\section{Ablation: Do \supers{} have a positional dependence?}
\label{app:pos-dependencies}

\begin{figure}[h!]
    \centering

    % -------- Row 1 --------
    \begin{subfigure}{0.49\textwidth}
        \centering
        \includegraphics[width=\textwidth]{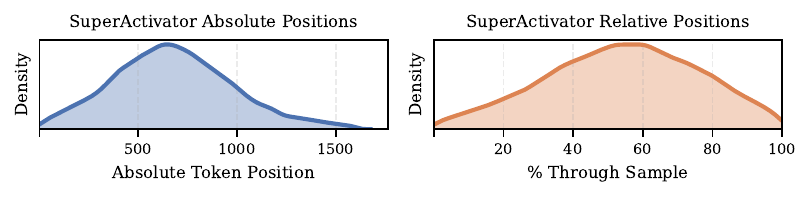}
        \vspace{-1.5em}
        \caption{\emph{COCO}}
    \end{subfigure}
    \hfill
    \begin{subfigure}{0.49\textwidth}
        \centering
        \includegraphics[width=\textwidth]{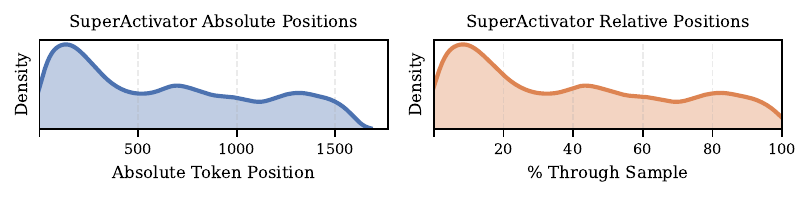}
        \vspace{-1.5em}
        \caption{\emph{Broden-OpenSurfaces}}
    \end{subfigure}

    \vspace{0.5em}

    % -------- Row 2 --------
    \begin{subfigure}{0.49\textwidth}
        \centering
        \includegraphics[width=\textwidth]{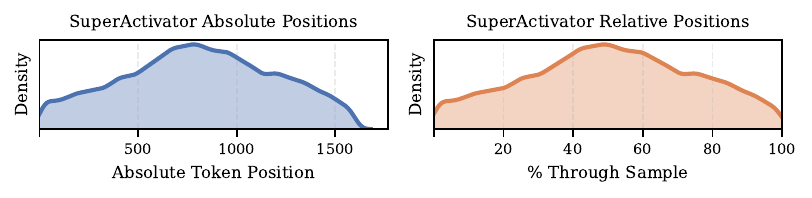}
        \vspace{-1.5em}
        \caption{\emph{Broden-Pascal}}
    \end{subfigure}
    \hfill
    \begin{subfigure}{0.49\textwidth}
        \centering
        \includegraphics[width=\textwidth]{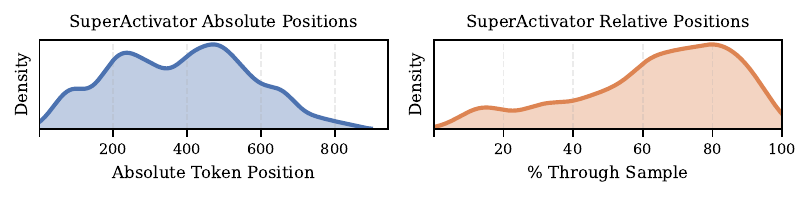}
        \vspace{-1.5em}
        \caption{\emph{PERSUADE}}
    \end{subfigure}

    \vspace{0.5em}

    \begin{subfigure}{0.49\textwidth}
        \centering
        \includegraphics[width=\textwidth]{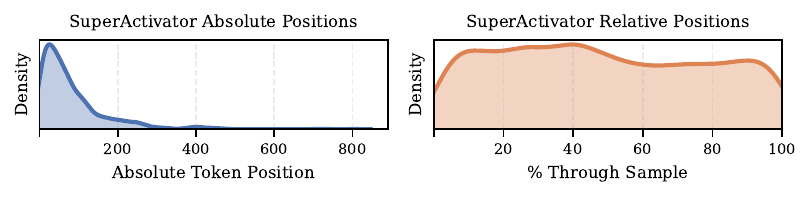 }
        \vspace{-1.5em}
        \caption{\emph{El Capitan}}
    \end{subfigure}\hfill
    \begin{subfigure}{0.49\textwidth}
        \centering
        \includegraphics[width=\textwidth]{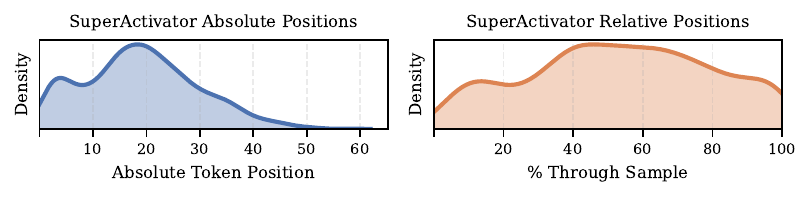}
        \vspace{-1.5em}
        \caption{\emph{GoEmotions}}
    \end{subfigure}

    \vspace{0.5em}

    \caption{\super{} Position Distribution Across Image and Text Domains}
    \label{fig:position-dependence}
\end{figure}

To check whether the \super{} Mechanism is driven by positional dependencies rather than genuine concept sensitivity, we plot the distribution of \supers{} across test splits (Figure \ref{fig:position-dependence}). For each dataset, we use \emph{Llama-3.2-11B-Vision-Instruct} linear separator concept \supers{}, defined at the concept-specific model depth and sparsity level $\delta$ that yield the best detection performance on the validation set. The left panels show absolute \super{} token positions, while the right panels show relative positions normalized to the length of each sample.

We do not observe positional biases. \emph{El Capitan} and \emph{GoEmotions} show some concentration at earlier absolute token positions, but this largely reflects variation in sample length, including a small number of much longer text samples. After normalizing position by sample length, the distributions are substantially more uniform, suggesting that \supers{} are not tied to a particular relative position within a sample.

\vspace{0.5in}

\clearpage
\section{\super{} Threshold Robustness}
\label{app:fixed-delta-detection}
Section~\ref{sec:detection-results} showed that the most reliable indicators of concept presence consistently lie in the extreme high tail of the in-concept activation distribution. Here, we test how robust this sparse-tail behavior is to using a single fixed sparsity level across concepts rather than tuning $\delta$ separately for each one.

Table \ref{tab:llama_linsep_fixed_delta_comparison} compares fully tuned \super{}-based detection, which selects the threshold $\tau_c^{\text{super}}$ continuously for each concept to maximize validation $F_1$, with a shared fixed sparsity level of $\delta=10\%$, alongside the strongest non-\super{} baseline. We report the average F1 of \emph{Llama-3.2-11B-Vision-Instruct} linear separator concepts, weighted by concept frequency, across all eight datasets. Using a fixed sparsity level across concepts, fixed-$\delta$ \super{} detection outperforms the strongest baseline on seven of eight datasets and remains within 0.024 $F_1$ of the fully tuned detector on five. The largest degradation occurs on PERSUADE, where fixed-$\delta$ also underperforms the baseline.

These results generally show that a sparsity of $\delta=10\%$ is robust across diverse datasets. This suggests that a similarly sparse fraction of the in-concept activation distribution contains reliable concept evidence across settings, and that per-concept tuning is not necessary to take advantage of the \super{} Mechanism.

\begin{table}[h]
\centering
% Built from llama_linsep_fixed_delta_comparison(1).csv; source values preserved.
% Requires \usepackage{booktabs}, \usepackage{float}, and \usepackage{graphicx}.
\begin{table}[H]
\centering
\small
\setlength{\tabcolsep}{8pt}
\caption{\textbf{Fixed-sparsity SuperAct detection compared with fully tuned SuperAct and the runner-up baseline.} Results use Llama-3.2-11B-Vision-Instruct linear separators. The fixed setting uses $\delta=10\%$. Reported uncertainties are 95\% bootstrap confidence intervals. \textbf{Bold} denotes the highest mean per row; \underline{underline} denotes the second-highest.}
\label{tab:llama_linsep_fixed_delta_comparison}
\resizebox{\columnwidth}{!}{%
\begin{tabular}{llccc}
    \toprule
    & & \multicolumn{3}{c}{Concept Detection $F_1$} \\
    \cmidrule(lr){3-5}
    Dataset & \shortstack{Runner-up\\Baseline} & \shortstack{Runner-up\\Score} & \shortstack{SuperAct\\(Tuned)} & \shortstack{SuperAct\\($\delta=10\%$)} \\
    \midrule
    CLEVR         & Prompt  & 0.987 {\scriptsize$\pm 0.008$} & \underline{0.997} {\scriptsize$\pm 0.004$} & \textbf{1.000} {\scriptsize$\pm 0.000$} \\
    COCO          & Prompt  & 0.686 {\scriptsize$\pm 0.010$} & \textbf{0.827} {\scriptsize$\pm 0.010$} & \underline{0.803} {\scriptsize$\pm 0.009$} \\
    OpenSurfaces  & MeanTok & 0.514 {\scriptsize$\pm 0.014$} & \textbf{0.567} {\scriptsize$\pm 0.015$} & \underline{0.547} {\scriptsize$\pm 0.013$} \\
    Pascal        & MeanTok & 0.709 {\scriptsize$\pm 0.006$} & \textbf{0.823} {\scriptsize$\pm 0.005$} & \underline{0.777} {\scriptsize$\pm 0.005$} \\
    \midrule
    PERSUADE      & CLSTok  & \underline{0.551} {\scriptsize$\pm 0.027$} & \textbf{0.639} {\scriptsize$\pm 0.026$} & 0.516 {\scriptsize$\pm 0.021$} \\
    El Capitan    & CLSTok  & 0.580 {\scriptsize$\pm 0.026$} & \textbf{0.696} {\scriptsize$\pm 0.022$} & \underline{0.625} {\scriptsize$\pm 0.022$} \\
    GoEmotions    & CLSTok  & 0.348 {\scriptsize$\pm 0.029$} & \underline{0.449} {\scriptsize$\pm 0.027$} & \textbf{0.459} {\scriptsize$\pm 0.025$} \\
    iSarcasm      & LastTok & \underline{0.905} {\scriptsize$\pm 0.033$} & \textbf{0.918} {\scriptsize$\pm 0.029$} & \textbf{0.918} {\scriptsize$\pm 0.029$} \\
    \bottomrule
\end{tabular}%
}
\vspace{-1em}
\end{table}

\end{table}

\clearpage

\section{Qualitative Visualizations of \supers{} Over Model Layers}
\label{app:qual-super-over-layers}
Next, we present qualitative examples from each primary dataset illustrating how the \super{} mechanism manifests across layers of the \emph{Llama-3.2-11B-Vision-Instruct} model. Each example visualizes linear separator activations for several concepts within a single test sample, along with the corresponding \supers{} identified using layer-specific, concept-calibrated thresholds.

% \begin{figure}[h]
%     \centering
%     \includegraphics[width=\textwidth]{Figures/qual_across_layers/qualoverlayers_CLEVR_idx_116_Llama_linsep.pdf}
%     \caption{\emph{CLEVR} -- \supers{} Across \emph{Llama-3.2-11B-Vision-Instruct} Layers}
% \end{figure}

\begin{figure}[h]
    \centering
    \includegraphics[width=\textwidth]{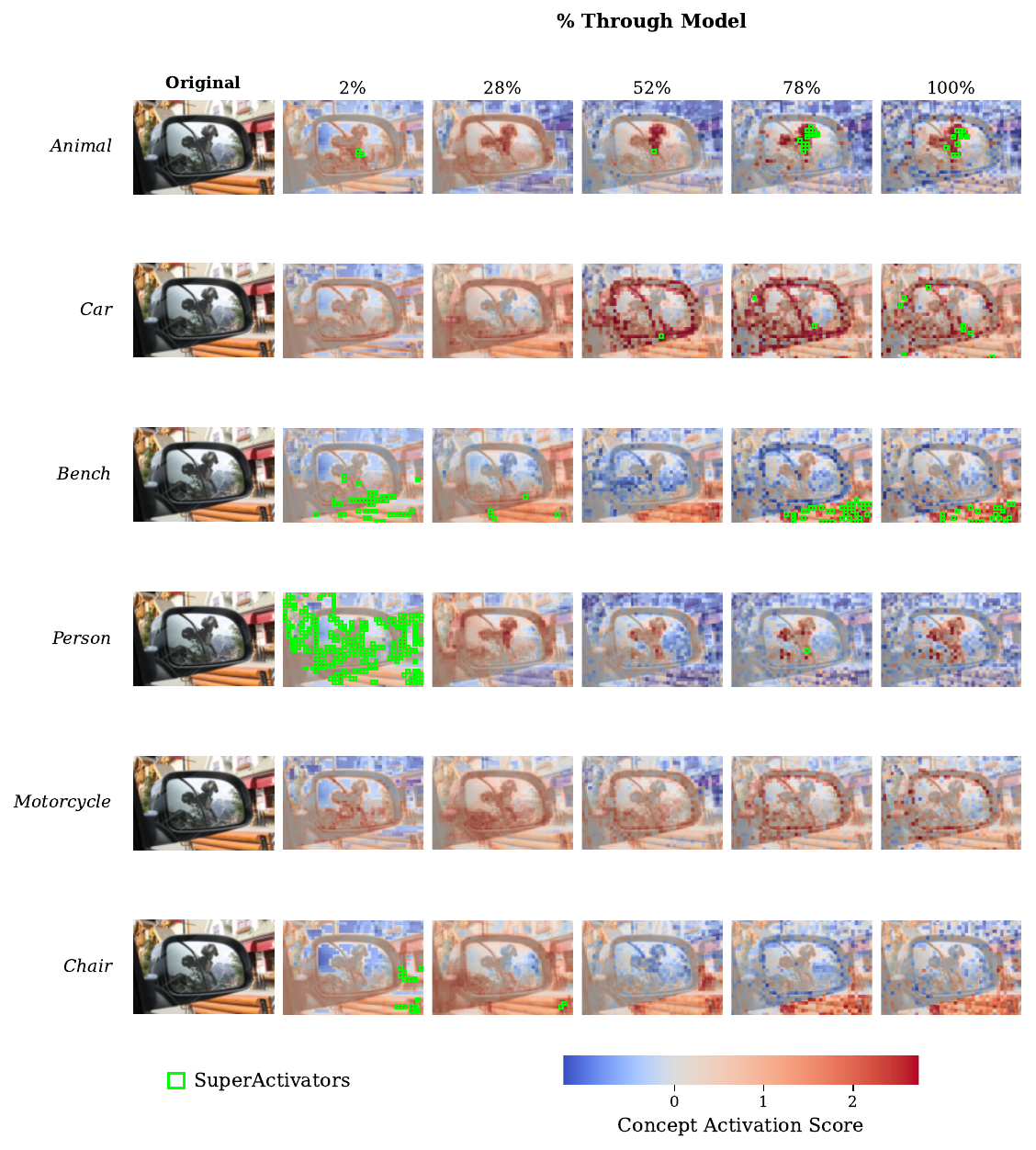}
    \caption{\emph{COCO} -- \supers{} Across \emph{Llama-3.2-11B-Vision-Instruct} Layers}
\end{figure}

\begin{figure}[h]
    \centering
    \includegraphics[width=\textwidth]{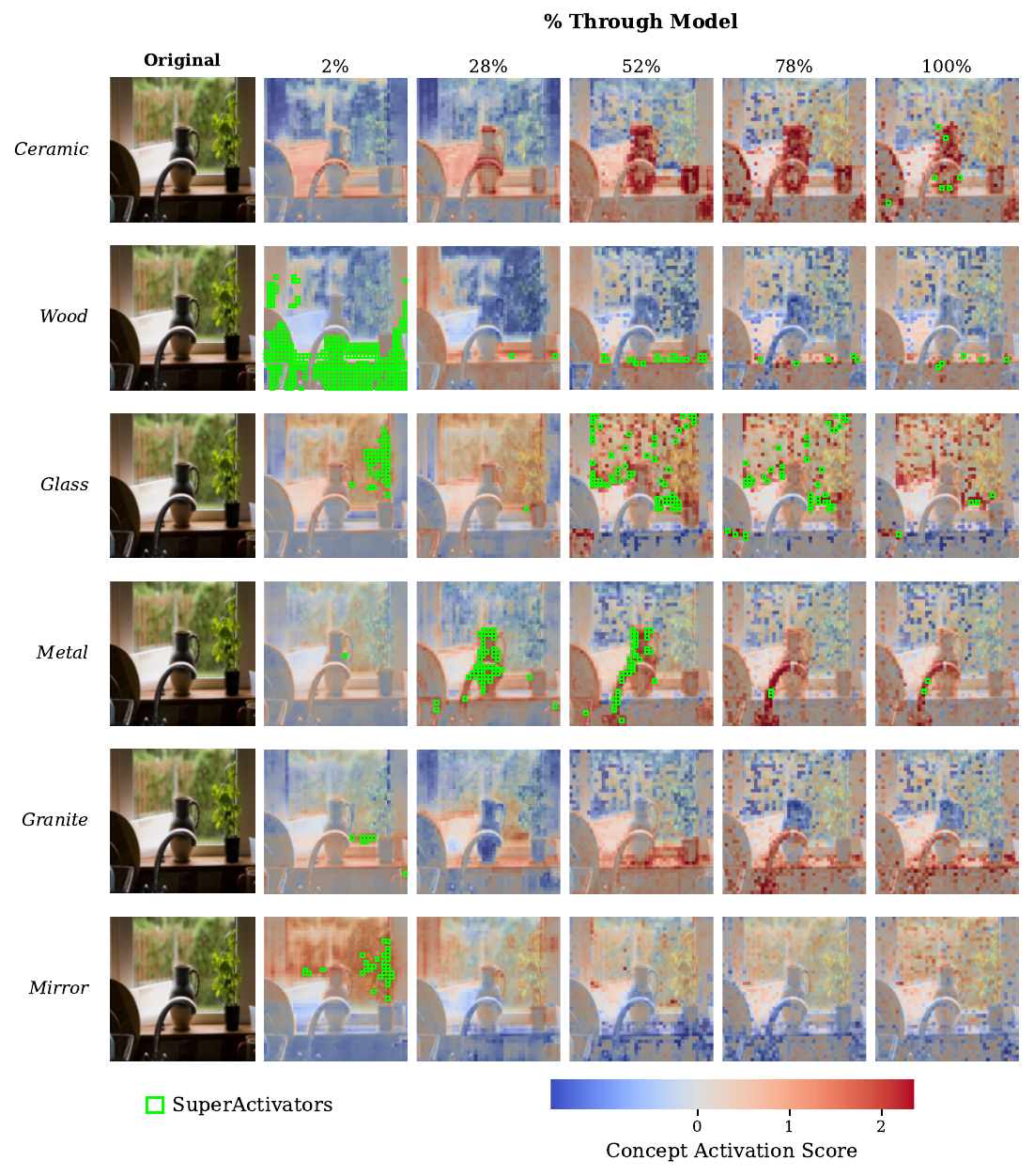}
    \caption{\emph{Broden-OpenSurfaces} -- \supers{} Across \emph{Llama-3.2-11B-Vision-Instruct} Layers}
\end{figure}

\begin{figure}[h]
    \centering
    \includegraphics[width=\textwidth]{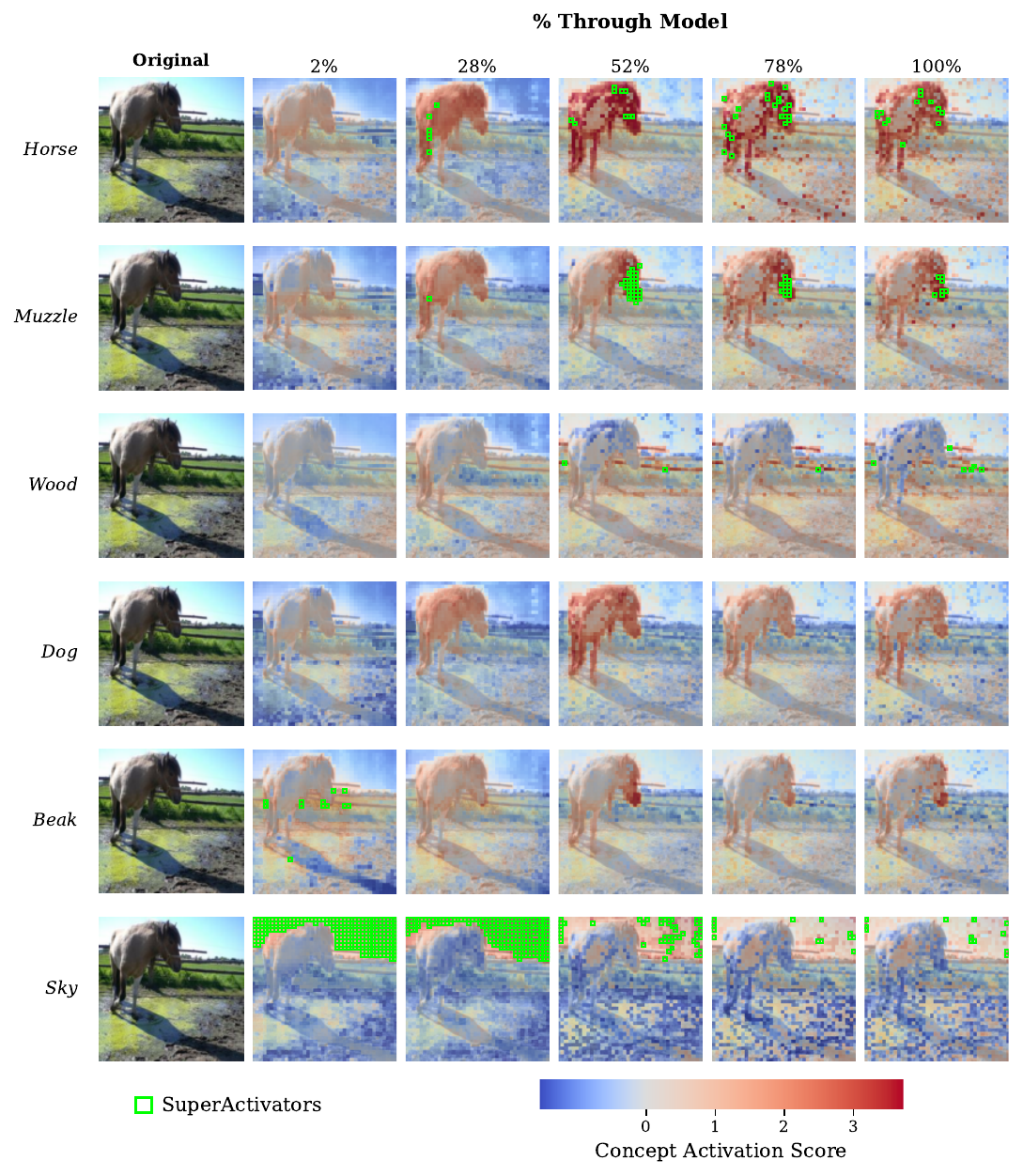}
    \caption{\emph{Broden-Pascal} -- \supers{} Across \emph{Llama-3.2-11B-Vision-Instruct} Layers}
\end{figure}

\begin{figure}[h]
    \centering
    \includegraphics[width=\textwidth]{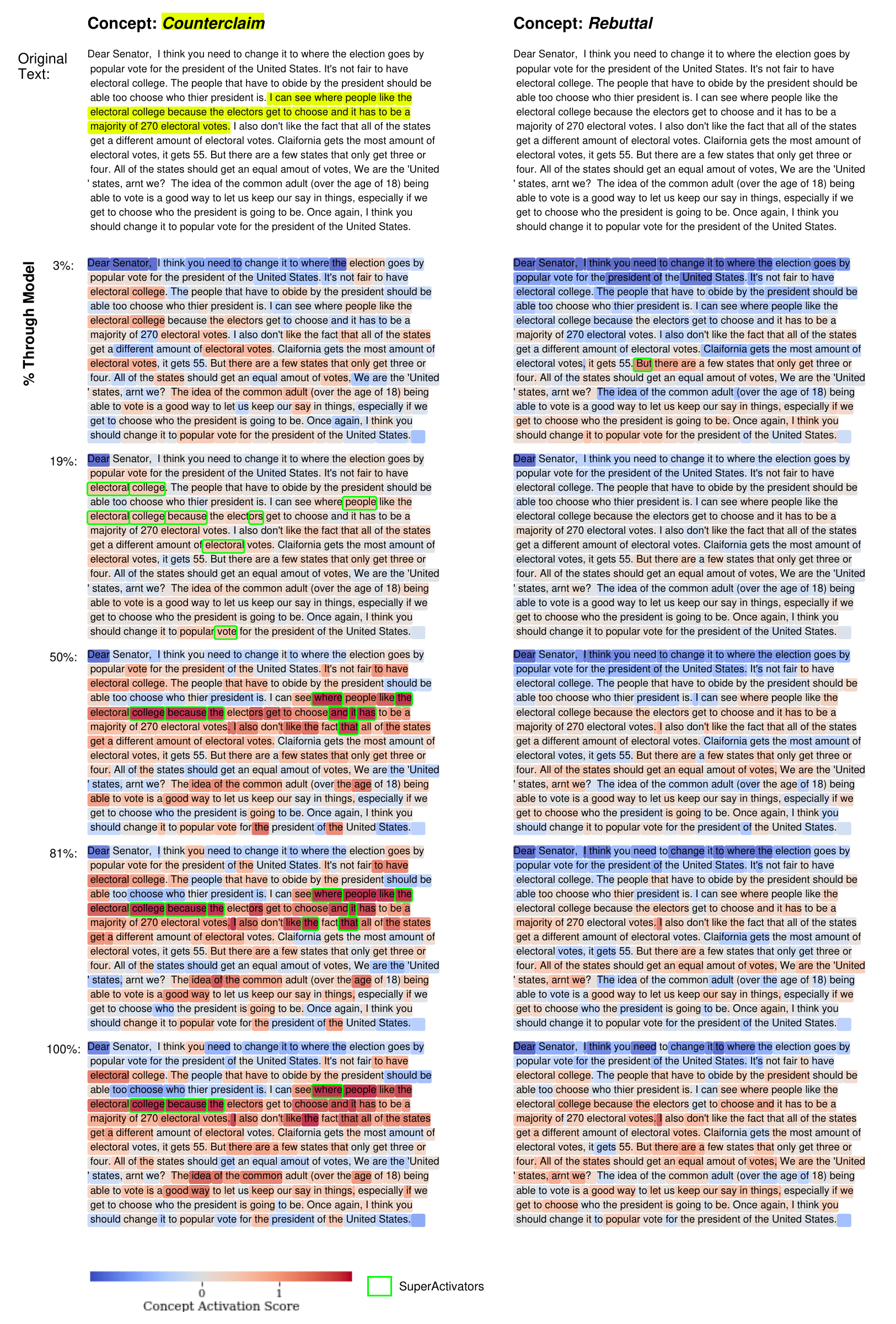}
    \caption{\emph{PERSUADE} -- \supers{} Across \emph{Llama-3.2-11B-Vision-Instruct} Layers}
\end{figure}

\begin{figure}[h]
    \centering
    \includegraphics[width=\textwidth]{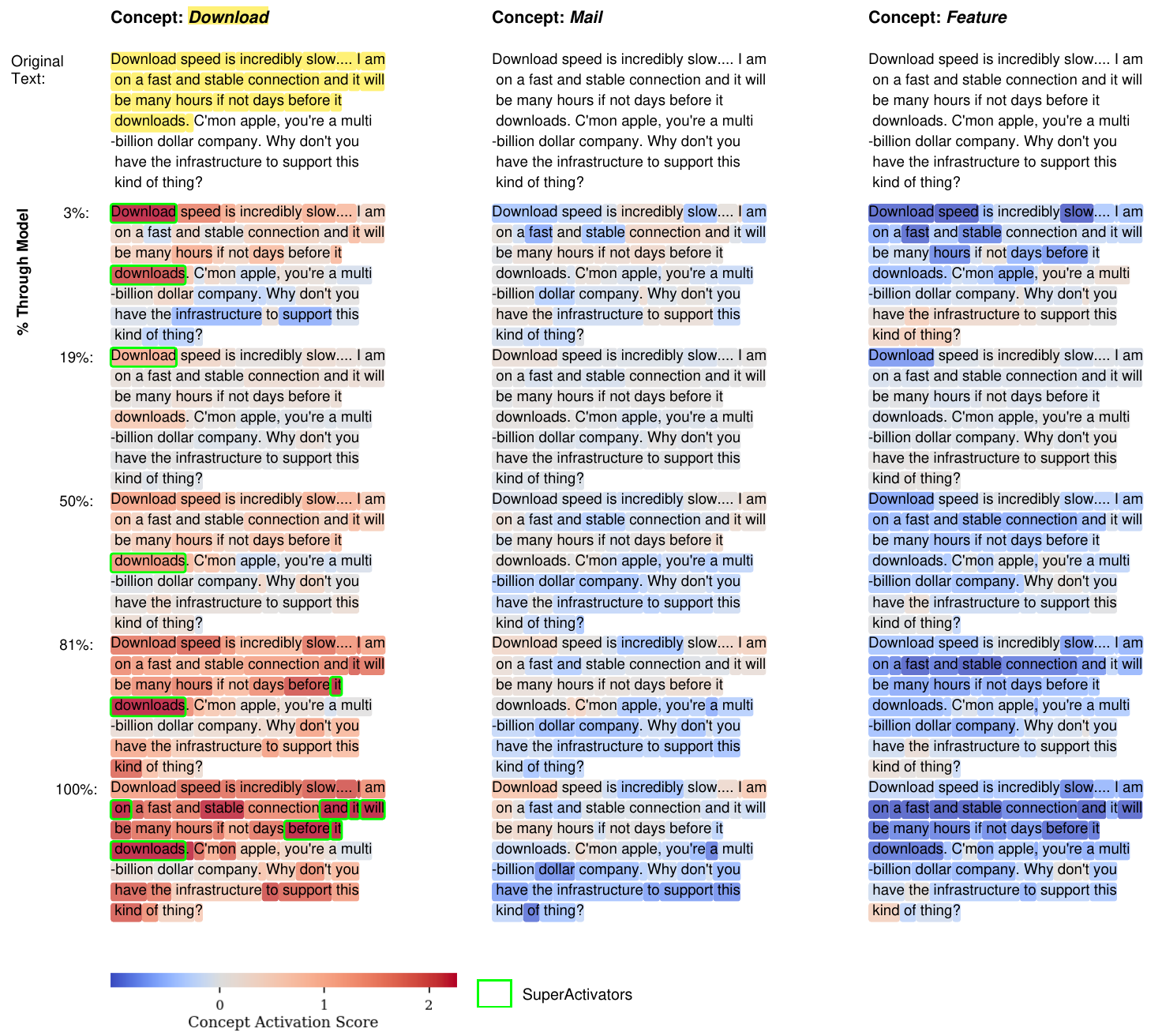}
    \caption{\emph{El Capitan} -- \supers{} Across \emph{Llama-3.2-11B-Vision-Instruct} Layers}
\end{figure}

\begin{figure}[h]
    \centering
    \includegraphics[width=\textwidth]{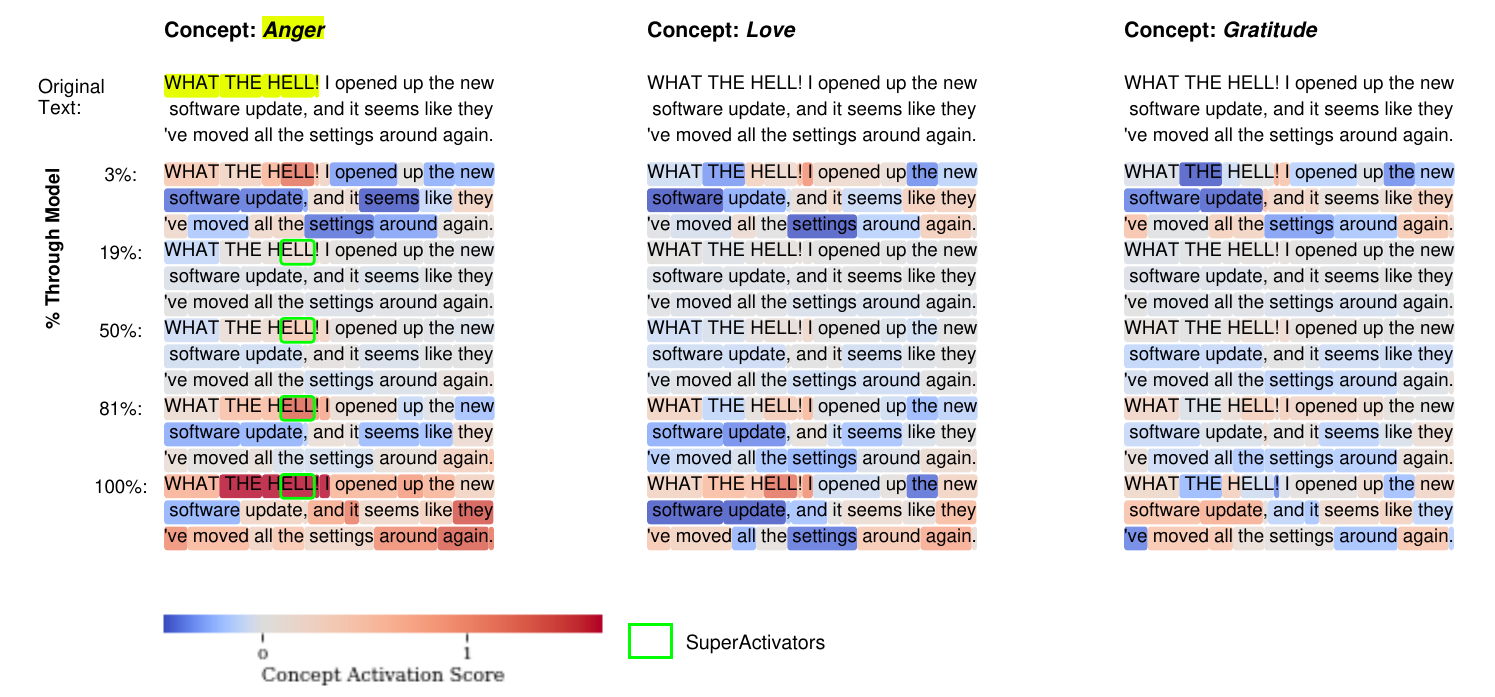}
    \caption{\emph{GoEmotions} -- \supers{} Across \emph{Llama-3.2-11B-Vision-Instruct} Layers}
\end{figure}

\clearpage
\section{Concept Attribution}
\label{app:concept_attribution}
\subsection{Attribution Methods}
\label{app:inversion_methods}
This section provides a brief overview of several attribution methods in which the objective is defined either by a global concept vector \(v_c\) or by the average embedding of local \supers{} $\mu_c(x)$.

\begin{itemize}
    \item \textbf{Direct Alignment (Raw Concept Activation)} As a baseline, we consider the raw per-token concept activations with respect to the target embedding as attribution scores.
    
    \item \textbf{LIME (Local Interpretable Model-agnostic Explanations)} \citep{ribeiro2016should} explains an individual prediction by approximating the complex model with a simpler, interpretable model (e.g., a linear model) in the local vicinity of the prediction. It achieves this by generating a new dataset of perturbed samples around the instance being explained and learning the simpler model on this new dataset, weighted by proximity to the original instance.

     \item \textbf{SHAP (SHapley Additive exPlanations)} \citep{lundberg2017unified} assigns each feature an importance value for a particular prediction based on the Shapley value from cooperative game theory. This value is the feature's marginal contribution averaged over all orderings of the features, and the attributions sum to the difference between the model's prediction and the expected model output.

    \item \textbf{RISE (Randomized Input Sampling for Explanation)} \citep{petsiuk2018rise} generates a visual explanation by probing the model with numerous randomly masked versions of an input image. The final importance map is a weighted average of these random masks, where weights are determined by the model's output confidence for each corresponding masked image.

    \item \textbf{SHAP IQ (SHAPley Interaction Quantification)} \citep{fel2023shap} estimates Shapley interaction indices, which extend the Shapley value from individual features to feature interactions of any order. It uses a unified sampling-based approximation that applies to any interaction index satisfying the linearity, symmetry, and dummy axioms.

    \item \textbf{IntGrad (Integrated Gradients)} \citep{sundararajan2017axiomatic} integrates the gradients of the model output along a straight-line path from a baseline input (e.g., a black image) to the actual input. Each feature's attribution is this integral scaled by the difference between the input and the baseline in that feature.

    \item \textbf{Grad-CAM (Gradient-weighted Class Activation Mapping)} \citep{selvaraju2017grad} produces a coarse localization map for CNNs using the gradients of the target class score with respect to the feature maps of the final convolutional layer. It averages these gradients spatially to weight each feature map and applies a ReLU to the weighted sum, highlighting regions that support the target class.

    \item \textbf{FullGrad} \citep{srinivas2019full} represents the network output as the sum of the input-gradient times the input and the bias-gradients, where a bias-gradient is the gradient with respect to a bias multiplied by that bias. For convolutional networks, it aggregates these terms from the input and every convolutional layer into a single saliency map.

    \item \textbf{CALM (Class Activation Latent Mapping)} \citep{kim2021calm} extends CAM by adding a latent variable that encodes the location of the cue for recognition, which makes the attribution map part of the training computational graph. The model is trained with expectation-maximization and identifies discriminative attributes more accurately than CAM.

    \item \textbf{MFABA (More Faithful and Accelerated Boundary-based Attribution)} \citep{zhu2024mfaba} runs gradient ascent on the loss to push the input across the decision boundary. It attributes the resulting loss change to each feature through a second-order Taylor expansion along this path, approximating the Hessian with differences of consecutive gradients.

\end{itemize}

\subsection{Additional Results for Concept Attribution}
\label{app:full-inversion}

This section presents the full concept attribution results across all experimental configurations, summarized in Table~\ref{tab:filtered} in the main text. These tables show that the main finding holds broadly across datasets, models, attribution methods, and concept types: using the average embedding of local \supers{} as the explanation
target matches or improves on the global concept vector in nearly all of the configurations.

We present our results across eight tables, each corresponding to one dataset and jointly evaluating both supervised and unsupervised concept representations. For each dataset—four image tasks (Tables~\ref{tab:inversion_clevr}-\ref{tab:inversion_opensurfaces}) and four text tasks (Tables~\ref{tab:inversion_persuade}-\ref{tab:inversion_isarcasm})—we report the concept-frequency–weighted average $F_{1}$ score (Appendix~\ref{app:concepts}). Error bars indicate the standard error of the mean $F_{1}$ score over multiple localization runs. Each table compares the \super{}-based attribution and global concept vector-based attribution across attribution methods, concept types, and models.

Our \super{}-based approach produces attribution maps that more closely align with ground-truth localization annotations than those derived from global concept vectors. Across the six primary datasets, using local \supers{} as the explanation target matches or improves $F_1$ in 96\% of configurations for dedicated attribution methods, with the largest and most consistent gains for linear separator concepts. Moreover, \super{}-based attribution methods outperform direct alignment in most dataset--model--concept-type configurations, and using raw cosine similarity to the average of local \supers{} as attribution scores rarely helps.

\begin{table}[ht] \caption{Average Localization F1 for the CLEVR Dataset.} 
\label{tab:inversion_clevr} \centering \small
\begin{tabular}{@{}llcccc@{}} 
\toprule
\multirow{2}{*}{Attribution Method} & \multirow{2}{*}{Concept Type} & \multicolumn{2}{c}{CLIP} & \multicolumn{2}{c}{Llama} \\
\cmidrule(lr){3-4} \cmidrule(lr){5-6}
& & Concept & SuperActs & Concept & SuperActs \\
\midrule
\multirow{4}{*}{Direct Alignment}
& Avg & \textbf{0.75 {\scriptsize $\pm$ 0.01}} & 0.74 {\scriptsize $\pm$ 0.01} & \textbf{0.90 {\scriptsize $\pm$ 0.00}} & 0.87 {\scriptsize $\pm$ 0.00} \\
& Linsep & \textbf{0.76 {\scriptsize $\pm$ 0.01}} & 0.75 {\scriptsize $\pm$ 0.01} & \textbf{0.97 {\scriptsize $\pm$ 0.00}} & 0.94 {\scriptsize $\pm$ 0.00} \\
& K-Means & 0.74 {\scriptsize $\pm$ 0.01} & \textbf{0.74 {\scriptsize $\pm$ 0.01}} & \textbf{0.52 {\scriptsize $\pm$ 0.00}} & 0.52 {\scriptsize $\pm$ 0.00} \\
& K-Linsep & \textbf{0.73 {\scriptsize $\pm$ 0.01}} & 0.71 {\scriptsize $\pm$ 0.01} & 0.51 {\scriptsize $\pm$ 0.00} & \textbf{0.51 {\scriptsize $\pm$ 0.00}} \\
\midrule 
\multirow{2}{*}{LIME} 
& Avg & 0.49 {\scriptsize $\pm$ 0.02} & \textbf{0.55 {\scriptsize $\pm$ 0.04}} & 0.76 {\scriptsize $\pm$ 0.03} & \textbf{0.81 {\scriptsize $\pm$ 0.02}} \\ 
& Linsep & 0.49 {\scriptsize $\pm$ 0.00} & \textbf{0.68 {\scriptsize $\pm$ 0.01}} & 0.70 {\scriptsize $\pm$ 0.01} & \textbf{0.85 {\scriptsize $\pm$ 0.01}} \\
\bottomrule \end{tabular} 
\end{table} 

\begin{table}[ht]
\caption{Average Localization F1 for the COCO Dataset.}
\label{tab:inversion_coco}
\centering
\small
\begin{tabular}{@{}llcccc@{}}
\toprule
\multirow{2}{*}{Attribution Method} & \multirow{2}{*}{Concept Type} & \multicolumn{2}{c}{CLIP} & \multicolumn{2}{c}{Llama} \\
\cmidrule(lr){3-4} \cmidrule(lr){5-6}
& & Concept & SuperActs & Concept & SuperActs \\
\midrule
\multirow{4}{*}{Direct Alignment}
& Avg & \textbf{0.56 {\scriptsize $\pm$ 0.00}} & 0.51 {\scriptsize $\pm$ 0.00} & \textbf{0.59 {\scriptsize $\pm$ 0.00}} & 0.51 {\scriptsize $\pm$ 0.00} \\
& Linsep & \textbf{0.63 {\scriptsize $\pm$ 0.00}} & 0.53 {\scriptsize $\pm$ 0.00} & \textbf{0.65 {\scriptsize $\pm$ 0.00}} & 0.50 {\scriptsize $\pm$ 0.00} \\
& K-Means & 0.47 {\scriptsize $\pm$ 0.00} & \textbf{0.48 {\scriptsize $\pm$ 0.00}} & \textbf{0.42 {\scriptsize $\pm$ 0.00}} & \textbf{0.42 {\scriptsize $\pm$ 0.00}} \\
& K-Linsep & 0.47 {\scriptsize $\pm$ 0.00} & \textbf{0.49 {\scriptsize $\pm$ 0.00}} & \textbf{0.41 {\scriptsize $\pm$ 0.00}} & 0.40 {\scriptsize $\pm$ 0.00} \\
\midrule
\multirow{4}{*}{LIME} 
& Avg & 0.32 {\scriptsize $\pm$ 0.01} & \textbf{0.38 {\scriptsize $\pm$ 0.02}} & 0.47 {\scriptsize $\pm$ 0.01} & \textbf{0.51 {\scriptsize $\pm$ 0.02}} \\
& Linsep & 0.29 {\scriptsize $\pm$ 0.02} & \textbf{0.40 {\scriptsize $\pm$ 0.03}} & 0.49 {\scriptsize $\pm$ 0.02} & \textbf{0.50 {\scriptsize $\pm$ 0.03}} \\
\bottomrule
\end{tabular}
\end{table}

\begin{table}[ht]
\caption{Average Localization F1 for the Pascal Dataset.}
\label{tab:inversion_pascal}
\centering
\small
\begin{tabular}{@{}llcccc@{}}
\toprule
\multirow{2}{*}{Attribution Method} & \multirow{2}{*}{Concept Type} & \multicolumn{2}{c}{CLIP} & \multicolumn{2}{c}{Llama} \\
\cmidrule(lr){3-4} \cmidrule(lr){5-6}
& & Concept & SuperActs & Concept & SuperActs \\
\midrule
\multirow{4}{*}{Direct Alignment}
& Avg & \textbf{0.61 {\scriptsize $\pm$ 0.00}} & 0.55 {\scriptsize $\pm$ 0.00} & \textbf{0.51 {\scriptsize $\pm$ 0.00}} & 0.44 {\scriptsize $\pm$ 0.00} \\
& Linsep & \textbf{0.67 {\scriptsize $\pm$ 0.00}} & 0.54 {\scriptsize $\pm$ 0.00} & \textbf{0.59 {\scriptsize $\pm$ 0.00}} & 0.44 {\scriptsize $\pm$ 0.00} \\
& K-Means & \textbf{0.50 {\scriptsize $\pm$ 0.00}} & 0.50 {\scriptsize $\pm$ 0.00} & \textbf{0.39 {\scriptsize $\pm$ 0.00}} & 0.39 {\scriptsize $\pm$ 0.00} \\
& K-Linsep & 0.50 {\scriptsize $\pm$ 0.00} & \textbf{0.51 {\scriptsize $\pm$ 0.00}} & \textbf{0.40 {\scriptsize $\pm$ 0.00}} & 0.38 {\scriptsize $\pm$ 0.00} \\
\midrule
\multirow{4}{*}{LIME}
& Avg      & 0.50 {\scriptsize $\pm$ 0.02} & \textbf{0.52 {\scriptsize $\pm$ 0.02}} & 0.69 {\scriptsize $\pm$ 0.02} & \textbf{0.71 {\scriptsize $\pm$ 0.03}} \\
& Linsep   & 0.51 {\scriptsize $\pm$ 0.03} & \textbf{0.55 {\scriptsize $\pm$ 0.01}} & 0.71 {\scriptsize $\pm$ 0.03} & \textbf{0.72 {\scriptsize $\pm$ 0.01}} \\
\bottomrule
\end{tabular}
\end{table}

\begin{table}[ht]
\caption{Average Localization F1 for the OpenSurfaces Dataset.}
\label{tab:inversion_opensurfaces}
\centering
\small
\begin{tabular}{@{}llcccc@{}}
\toprule
\multirow{2}{*}{Attribution Method} & \multirow{2}{*}{Concept Type} & \multicolumn{2}{c}{CLIP} & \multicolumn{2}{c}{Llama} \\
\cmidrule(lr){3-4} \cmidrule(lr){5-6}
& & Concept & SuperActs & Concept & SuperActs \\
\midrule
\multirow{4}{*}{Direct Alignment}
& Avg & \textbf{0.32 {\scriptsize $\pm$ 0.00}} & 0.30 {\scriptsize $\pm$ 0.00} & \textbf{0.26 {\scriptsize $\pm$ 0.00}} & 0.24 {\scriptsize $\pm$ 0.00} \\
& Linsep & \textbf{0.35 {\scriptsize $\pm$ 0.00}} & 0.29 {\scriptsize $\pm$ 0.00} & \textbf{0.31 {\scriptsize $\pm$ 0.00}} & 0.25 {\scriptsize $\pm$ 0.00} \\
& K-Means & \textbf{0.27 {\scriptsize $\pm$ 0.00}} & 0.26 {\scriptsize $\pm$ 0.00} & 0.19 {\scriptsize $\pm$ 0.00} & \textbf{0.19 {\scriptsize $\pm$ 0.00}} \\
& K-Linsep & \textbf{0.25 {\scriptsize $\pm$ 0.00}} & 0.24 {\scriptsize $\pm$ 0.00} & 0.18 {\scriptsize $\pm$ 0.00} & \textbf{0.20 {\scriptsize $\pm$ 0.00}} \\
\midrule
\multirow{4}{*}{LIME}
& Avg      & 0.42 {\scriptsize $\pm$ 0.03} & \textbf{0.50 {\scriptsize $\pm$ 0.01}} & 0.55 {\scriptsize $\pm$ 0.03} & \textbf{0.62 {\scriptsize $\pm$ 0.01}} \\
& Linsep   & 0.46 {\scriptsize $\pm$ 0.01} & \textbf{0.50 {\scriptsize $\pm$ 0.03}} & 0.60 {\scriptsize $\pm$ 0.01} & \textbf{0.68 {\scriptsize $\pm$ 0.02}} \\
\bottomrule
\end{tabular}
\end{table}

\begin{table}[ht]
\caption{Average Localization F1 for the PERSUADE Dataset.}
\label{tab:inversion_persuade}
\centering
\small
\begin{tabular}{@{}llcccc@{}}
\toprule
\multirow{2}{*}{Attribution Method} & \multirow{2}{*}{Concept Type} & \multicolumn{2}{c}{Llama} & \multicolumn{2}{c}{Qwen} \\
\cmidrule(lr){3-4} \cmidrule(lr){5-6}
& & Concept & SuperActs & Concept & SuperActs \\
\midrule
\multirow{4}{*}{Direct Alignment}
& Avg & \textbf{0.25 {\scriptsize $\pm$ 0.00}} & 0.19 {\scriptsize $\pm$ 0.00} & \textbf{0.10 {\scriptsize $\pm$ 0.00}} & 0.07 {\scriptsize $\pm$ 0.00} \\
& Linsep & \textbf{0.35 {\scriptsize $\pm$ 0.01}} & 0.21 {\scriptsize $\pm$ 0.00} & \textbf{0.27 {\scriptsize $\pm$ 0.00}} & 0.16 {\scriptsize $\pm$ 0.00} \\
& K-Means & 0.10 {\scriptsize $\pm$ 0.00} & \textbf{0.15 {\scriptsize $\pm$ 0.00}} & \textbf{0.08 {\scriptsize $\pm$ 0.00}} & \textbf{0.08 {\scriptsize $\pm$ 0.00}} \\
& K-Linsep & 0.10 {\scriptsize $\pm$ 0.00} & \textbf{0.17 {\scriptsize $\pm$ 0.00}} & 0.07 {\scriptsize $\pm$ 0.00} & \textbf{0.08 {\scriptsize $\pm$ 0.00}} \\
\midrule
\multirow{4}{*}{LIME} & Avg      & 0.28 {\scriptsize $\pm$ 0.05} & \textbf{0.33 {\scriptsize $\pm$ 0.05}} & 0.18 {\scriptsize $\pm$ 0.03} & \textbf{0.22 {\scriptsize $\pm$ 0.04}} \\
 & Linsep   & 0.32 {\scriptsize $\pm$ 0.05} & \textbf{0.40 {\scriptsize $\pm$ 0.06}} & 0.17 {\scriptsize $\pm$ 0.04} & \textbf{0.22 {\scriptsize $\pm$ 0.05}} \\
\bottomrule
\end{tabular}
\end{table}

\begin{table}[ht]
\caption{Average Localization F1 for the El Capitan Dataset.}
\label{tab:inversion_el_capitan}
\centering
\small
\begin{tabular}{@{}llcccc@{}}
\toprule
\multirow{2}{*}{Attribution Method} & \multirow{2}{*}{Concept Type} & \multicolumn{2}{c}{Llama} & \multicolumn{2}{c}{Qwen} \\
\cmidrule(lr){3-4} \cmidrule(lr){5-6}
& & Concept & SuperActs & Concept & SuperActs \\
\midrule
\multirow{4}{*}{Direct Alignment}
& Avg & \textbf{0.29 {\scriptsize $\pm$ 0.00}} & 0.25 {\scriptsize $\pm$ 0.00} & \textbf{0.33 {\scriptsize $\pm$ 0.00}} & 0.30 {\scriptsize $\pm$ 0.00} \\
& Linsep & \textbf{0.44 {\scriptsize $\pm$ 0.00}} & 0.36 {\scriptsize $\pm$ 0.00} & \textbf{0.42 {\scriptsize $\pm$ 0.00}} & 0.33 {\scriptsize $\pm$ 0.00} \\
& K-Means & \textbf{0.31 {\scriptsize $\pm$ 0.00}} & \textbf{0.31 {\scriptsize $\pm$ 0.00}} & \textbf{0.30 {\scriptsize $\pm$ 0.00}} & \textbf{0.30 {\scriptsize $\pm$ 0.00}} \\
& K-Linsep & 0.30 {\scriptsize $\pm$ 0.00} & \textbf{0.31 {\scriptsize $\pm$ 0.00}} & \textbf{0.29 {\scriptsize $\pm$ 0.00}} & \textbf{0.29 {\scriptsize $\pm$ 0.00}} \\
\midrule
\multirow{4}{*}{LIME}
& Avg      & 0.46 {\scriptsize $\pm$ 0.08} & \textbf{0.48 {\scriptsize $\pm$ 0.09}} & 0.41 {\scriptsize $\pm$ 0.10} & \textbf{0.44 {\scriptsize $\pm$ 0.10}} \\
& Linsep   & 0.50 {\scriptsize $\pm$ 0.10} & \textbf{0.52 {\scriptsize $\pm$ 0.10}} & 0.48 {\scriptsize $\pm$ 0.10} & \textbf{0.49 {\scriptsize $\pm$ 0.11}} \\
\bottomrule
\end{tabular}
\end{table}

\begin{table}[ht]
\caption{Average Localization F1 for the GoEmotions Dataset.}
\label{tab:inversion_goemotions}
\centering
\small
\begin{tabular}{@{}llcccc@{}}
\toprule
\multirow{2}{*}{Attribution Method} & \multirow{2}{*}{Concept Type} & \multicolumn{2}{c}{Llama} & \multicolumn{2}{c}{Qwen} \\
\cmidrule(lr){3-4} \cmidrule(lr){5-6}
& & Concept & SuperActs & Concept & SuperActs \\
\midrule
\multirow{4}{*}{Direct Alignment}
& Avg & \textbf{0.21 {\scriptsize $\pm$ 0.01}} & 0.15 {\scriptsize $\pm$ 0.00} & \textbf{0.46 {\scriptsize $\pm$ 0.01}} & 0.39 {\scriptsize $\pm$ 0.01} \\
& Linsep & \textbf{0.52 {\scriptsize $\pm$ 0.01}} & 0.42 {\scriptsize $\pm$ 0.01} & \textbf{0.48 {\scriptsize $\pm$ 0.01}} & 0.40 {\scriptsize $\pm$ 0.01} \\
& K-Means & \textbf{0.19 {\scriptsize $\pm$ 0.01}} & 0.17 {\scriptsize $\pm$ 0.00} & \textbf{0.42 {\scriptsize $\pm$ 0.01}} & \textbf{0.42 {\scriptsize $\pm$ 0.01}} \\
& K-Linsep & \textbf{0.43 {\scriptsize $\pm$ 0.01}} & 0.42 {\scriptsize $\pm$ 0.01} & \textbf{0.42 {\scriptsize $\pm$ 0.01}} & 0.41 {\scriptsize $\pm$ 0.01} \\
\midrule
\multirow{4}{*}{LIME}
& Avg      & 0.20 {\scriptsize $\pm$ 0.03} & \textbf{0.25 {\scriptsize $\pm$ 0.01}} & 0.27 {\scriptsize $\pm$ 0.01} & \textbf{0.31 {\scriptsize $\pm$ 0.02}} \\
& Linsep   & 0.29 {\scriptsize $\pm$ 0.02} & \textbf{0.34 {\scriptsize $\pm$ 0.03}} & 0.33 {\scriptsize $\pm$ 0.03} & \textbf{0.37 {\scriptsize $\pm$ 0.01}} \\
\bottomrule
\end{tabular}
\end{table}

\begin{table}[ht]
\caption{Average Localization F1 for the iSarcasm Dataset.}
\label{tab:inversion_isarcasm}
\centering
\small
\begin{tabular}{@{}llcccc@{}}
\toprule
\multirow{2}{*}{Attribution Method} & \multirow{2}{*}{Concept Type} & \multicolumn{2}{c}{Llama} & \multicolumn{2}{c}{Qwen} \\
\cmidrule(lr){3-4} \cmidrule(lr){5-6}
& & Concept & SuperActs & Concept & SuperActs \\
\midrule
\multirow{4}{*}{Direct Alignment}
& Avg & \textbf{0.72 {\scriptsize $\pm$ 0.01}} & 0.65 {\scriptsize $\pm$ 0.01} & \textbf{0.66 {\scriptsize $\pm$ 0.01}} & 0.62 {\scriptsize $\pm$ 0.01} \\
& Linsep & \textbf{0.83 {\scriptsize $\pm$ 0.01}} & 0.72 {\scriptsize $\pm$ 0.01} & \textbf{0.85 {\scriptsize $\pm$ 0.01}} & 0.74 {\scriptsize $\pm$ 0.01} \\
& K-Means & \textbf{0.68 {\scriptsize $\pm$ 0.01}} & \textbf{0.68 {\scriptsize $\pm$ 0.01}} & \textbf{0.61 {\scriptsize $\pm$ 0.01}} & \textbf{0.61 {\scriptsize $\pm$ 0.01}} \\
& K-Linsep & \textbf{0.74 {\scriptsize $\pm$ 0.01}} & 0.69 {\scriptsize $\pm$ 0.01} & \textbf{0.76 {\scriptsize $\pm$ 0.01}} & 0.73 {\scriptsize $\pm$ 0.01} \\
\midrule
\multirow{4}{*}{LIME}
& Avg      & 0.71 {\scriptsize $\pm$ 0.02} & \textbf{0.78 {\scriptsize $\pm$ 0.01}} & 0.63 {\scriptsize $\pm$ 0.02} & \textbf{0.67 {\scriptsize $\pm$ 0.03}} \\
& Linsep   & 0.79 {\scriptsize $\pm$ 0.01} & \textbf{0.87 {\scriptsize $\pm$ 0.02}} & 0.71 {\scriptsize $\pm$ 0.01} & \textbf{0.80 {\scriptsize $\pm$ 0.02}} \\
\bottomrule
\end{tabular}
\end{table}

Table~\ref{tab:inversion_llama_insdel} reports insertion and deletion scores for Llama-3.2-11B-Vision-Instruct linear separators on all eight datasets, using four attribution methods: LIME, SHAP, IntGrad, and MFABA. Higher insertion and lower deletion indicate that an attribution map ranks the input features the model relies on more faithfully. Explaining alignment to local \supers{} yields higher insertion and lower deletion scores than explaining alignment to the global concept vector across all datasets and attribution methods.

\begin{table}[t]
\centering
\caption{We report average insertion ($\uparrow$) and deletion ($\downarrow$) scores on all eight datasets for four attribution methods. Bold indicates the better mean for each Concept/SuperActs pair.}
\label{tab:inversion_llama_insdel}
\small
\begin{tabular}{llcccc}
\toprule
\multirow{3}{*}{\begin{tabular}[c]{@{}l@{}}Attribution Method\end{tabular}} & \multirow{3}{*}{Dataset}
 & \multicolumn{2}{c}{Insertion Score} & \multicolumn{2}{c}{Deletion Score} \\
 & & \multicolumn{2}{c}{\scriptsize ($\uparrow$ is better)} & \multicolumn{2}{c}{\scriptsize ($\downarrow$ is better)} \\
\cmidrule(lr){3-4} \cmidrule(lr){5-6}
 & & Concept & SuperActs & Concept & SuperActs \\
\midrule
\multirow{8}{*}{LIME}
 & COCO & 0.397 {\scriptsize $\pm$ 0.008} & \textbf{0.438} {\scriptsize $\pm$ 0.009} & 0.007 {\scriptsize $\pm$ 0.001} & \textbf{0.005} {\scriptsize $\pm$ 0.001} \\
 & Pascal & 0.379 {\scriptsize $\pm$ 0.006} & \textbf{0.405} {\scriptsize $\pm$ 0.005} & 0.007 {\scriptsize $\pm$ 0.001} & \textbf{0.006} {\scriptsize $\pm$ 0.001} \\
 & OpenSurfaces & 0.197 {\scriptsize $\pm$ 0.003} & \textbf{0.213} {\scriptsize $\pm$ 0.004} & 0.013 {\scriptsize $\pm$ 0.003} & \textbf{0.010} {\scriptsize $\pm$ 0.002} \\
 & CLEVR & 0.521 {\scriptsize $\pm$ 0.010} & \textbf{0.556} {\scriptsize $\pm$ 0.012} & 0.005 {\scriptsize $\pm$ 0.001} & \textbf{0.004} {\scriptsize $\pm$ 0.001} \\
 & PERSUADE & 0.128 {\scriptsize $\pm$ 0.003} & \textbf{0.131} {\scriptsize $\pm$ 0.003} & 0.018 {\scriptsize $\pm$ 0.003} & \textbf{0.012} {\scriptsize $\pm$ 0.003} \\
 & El Capitan & 0.194 {\scriptsize $\pm$ 0.004} & \textbf{0.206} {\scriptsize $\pm$ 0.004} & 0.015 {\scriptsize $\pm$ 0.002} & \textbf{0.013} {\scriptsize $\pm$ 0.002} \\
 & GoEmotions & 0.147 {\scriptsize $\pm$ 0.003} & \textbf{0.161} {\scriptsize $\pm$ 0.003} & 0.021 {\scriptsize $\pm$ 0.004} & \textbf{0.017} {\scriptsize $\pm$ 0.003} \\
 & iSarcasm & 0.418 {\scriptsize $\pm$ 0.009} & \textbf{0.469} {\scriptsize $\pm$ 0.008} & 0.007 {\scriptsize $\pm$ 0.000} & \textbf{0.004} {\scriptsize $\pm$ 0.001} \\
\midrule
\multirow{8}{*}{SHAP}
 & COCO & 0.403 {\scriptsize $\pm$ 0.008} & \textbf{0.439} {\scriptsize $\pm$ 0.003} & 0.006 {\scriptsize $\pm$ 0.001} & \textbf{0.005} {\scriptsize $\pm$ 0.002} \\
 & Pascal & 0.400 {\scriptsize $\pm$ 0.006} & \textbf{0.421} {\scriptsize $\pm$ 0.005} & \textbf{0.006} {\scriptsize $\pm$ 0.001} & \textbf{0.006} {\scriptsize $\pm$ 0.002} \\
 & OpenSurfaces & 0.181 {\scriptsize $\pm$ 0.004} & \textbf{0.222} {\scriptsize $\pm$ 0.003} & 0.012 {\scriptsize $\pm$ 0.003} & \textbf{0.007} {\scriptsize $\pm$ 0.002} \\
 & CLEVR & 0.538 {\scriptsize $\pm$ 0.007} & \textbf{0.562} {\scriptsize $\pm$ 0.007} & \textbf{0.005} {\scriptsize $\pm$ 0.001} & \textbf{0.005} {\scriptsize $\pm$ 0.001} \\
 & PERSUADE & 0.109 {\scriptsize $\pm$ 0.002} & \textbf{0.122} {\scriptsize $\pm$ 0.001} & 0.020 {\scriptsize $\pm$ 0.003} & \textbf{0.017} {\scriptsize $\pm$ 0.002} \\
 & El Capitan & 0.206 {\scriptsize $\pm$ 0.003} & \textbf{0.209} {\scriptsize $\pm$ 0.004} & \textbf{0.016} {\scriptsize $\pm$ 0.002} & \textbf{0.016} {\scriptsize $\pm$ 0.002} \\
 & GoEmotions & 0.132 {\scriptsize $\pm$ 0.002} & \textbf{0.144} {\scriptsize $\pm$ 0.002} & 0.027 {\scriptsize $\pm$ 0.003} & \textbf{0.019} {\scriptsize $\pm$ 0.002} \\
 & iSarcasm & 0.424 {\scriptsize $\pm$ 0.005} & \textbf{0.448} {\scriptsize $\pm$ 0.004} & 0.006 {\scriptsize $\pm$ 0.001} & \textbf{0.004} {\scriptsize $\pm$ 0.001} \\
\midrule
\multirow{8}{*}{IntGrad}
 & COCO & 0.397 {\scriptsize $\pm$ 0.004} & \textbf{0.437} {\scriptsize $\pm$ 0.009} & 0.009 {\scriptsize $\pm$ 0.003} & \textbf{0.006} {\scriptsize $\pm$ 0.003} \\
 & Pascal & 0.371 {\scriptsize $\pm$ 0.008} & \textbf{0.410} {\scriptsize $\pm$ 0.009} & 0.009 {\scriptsize $\pm$ 0.003} & \textbf{0.008} {\scriptsize $\pm$ 0.003} \\
 & OpenSurfaces & 0.185 {\scriptsize $\pm$ 0.004} & \textbf{0.204} {\scriptsize $\pm$ 0.004} & 0.017 {\scriptsize $\pm$ 0.004} & \textbf{0.015} {\scriptsize $\pm$ 0.004} \\
 & CLEVR & 0.496 {\scriptsize $\pm$ 0.007} & \textbf{0.527} {\scriptsize $\pm$ 0.006} & 0.007 {\scriptsize $\pm$ 0.001} & \textbf{0.005} {\scriptsize $\pm$ 0.001} \\
 & PERSUADE & 0.112 {\scriptsize $\pm$ 0.003} & \textbf{0.124} {\scriptsize $\pm$ 0.002} & 0.022 {\scriptsize $\pm$ 0.001} & \textbf{0.019} {\scriptsize $\pm$ 0.001} \\
 & El Capitan & 0.171 {\scriptsize $\pm$ 0.003} & \textbf{0.188} {\scriptsize $\pm$ 0.003} & 0.018 {\scriptsize $\pm$ 0.003} & \textbf{0.015} {\scriptsize $\pm$ 0.004} \\
 & GoEmotions & 0.129 {\scriptsize $\pm$ 0.002} & \textbf{0.146} {\scriptsize $\pm$ 0.002} & 0.020 {\scriptsize $\pm$ 0.002} & \textbf{0.016} {\scriptsize $\pm$ 0.002} \\
 & iSarcasm & 0.378 {\scriptsize $\pm$ 0.003} & \textbf{0.445} {\scriptsize $\pm$ 0.005} & 0.007 {\scriptsize $\pm$ 0.001} & \textbf{0.005} {\scriptsize $\pm$ 0.002} \\
\midrule
\multirow{8}{*}{MFABA}
 & COCO & 0.401 {\scriptsize $\pm$ 0.002} & \textbf{0.443} {\scriptsize $\pm$ 0.009} & 0.004 {\scriptsize $\pm$ 0.001} & \textbf{0.003} {\scriptsize $\pm$ 0.001} \\
 & Pascal & 0.395 {\scriptsize $\pm$ 0.005} & \textbf{0.436} {\scriptsize $\pm$ 0.005} & 0.006 {\scriptsize $\pm$ 0.002} & \textbf{0.004} {\scriptsize $\pm$ 0.001} \\
 & OpenSurfaces & 0.205 {\scriptsize $\pm$ 0.003} & \textbf{0.217} {\scriptsize $\pm$ 0.004} & 0.012 {\scriptsize $\pm$ 0.002} & \textbf{0.011} {\scriptsize $\pm$ 0.002} \\
 & CLEVR & 0.502 {\scriptsize $\pm$ 0.007} & \textbf{0.545} {\scriptsize $\pm$ 0.007} & 0.003 {\scriptsize $\pm$ 0.001} & \textbf{0.002} {\scriptsize $\pm$ 0.001} \\
 & PERSUADE & 0.132 {\scriptsize $\pm$ 0.002} & \textbf{0.143} {\scriptsize $\pm$ 0.002} & 0.016 {\scriptsize $\pm$ 0.002} & \textbf{0.013} {\scriptsize $\pm$ 0.001} \\
 & El Capitan & 0.194 {\scriptsize $\pm$ 0.003} & \textbf{0.210} {\scriptsize $\pm$ 0.004} & 0.009 {\scriptsize $\pm$ 0.002} & \textbf{0.008} {\scriptsize $\pm$ 0.002} \\
 & GoEmotions & 0.147 {\scriptsize $\pm$ 0.002} & \textbf{0.156} {\scriptsize $\pm$ 0.002} & 0.012 {\scriptsize $\pm$ 0.002} & \textbf{0.008} {\scriptsize $\pm$ 0.002} \\
 & iSarcasm & 0.429 {\scriptsize $\pm$ 0.005} & \textbf{0.486} {\scriptsize $\pm$ 0.006} & 0.004 {\scriptsize $\pm$ 0.001} & \textbf{0.002} {\scriptsize $\pm$ 0.001} \\
\bottomrule
\end{tabular}
\end{table}

\clearpage

\subsection{Qualitative Example Showing \supers{} for Improved Concept Attribution}
\label{app:inversion-qual-examples}

We present additional qualitative examples to demonstrate the benefit of computing attributions for concepts using local \supers{} instead of global concept vectors.

\begin{figure*}[!h]
    \centering
    \setlength{\fboxsep}{0pt}
    \begin{subfigure}[t]{0.31\textwidth}
    \includegraphics[width=\linewidth]{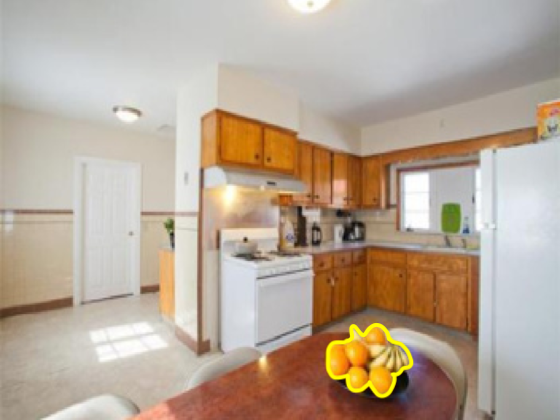}
        \caption{Image with \textit{Food} labeled}
    \end{subfigure}% 
    \hspace{0.01\textwidth}
    \begin{subfigure}[t]{0.31\textwidth}
    \includegraphics[width=\linewidth]{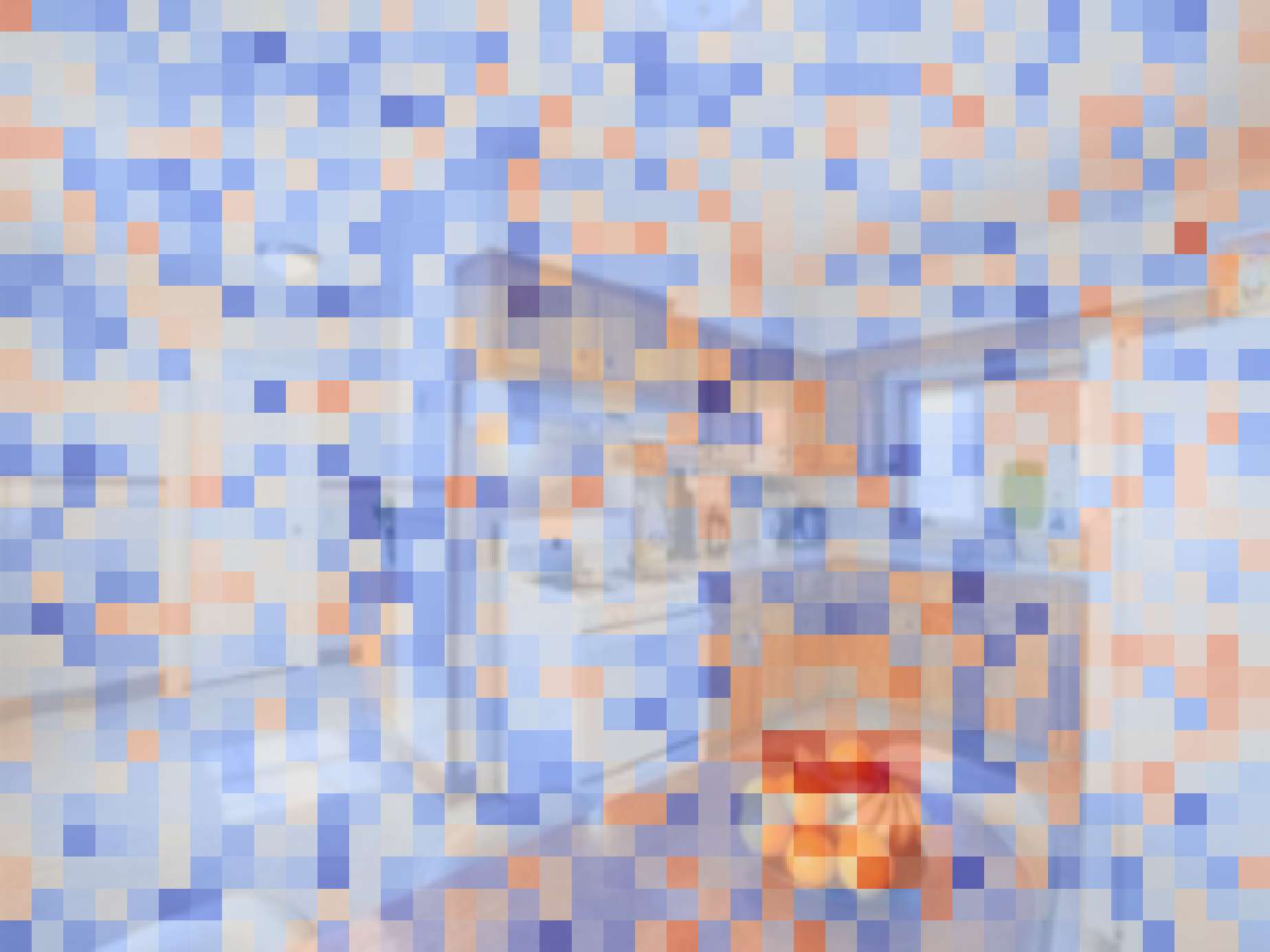}
        \caption{Attr. target: \textit{Food} concept vector}
    \end{subfigure}% 
    \hspace{0.01\textwidth}
    \begin{subfigure}[t]{0.31\textwidth}
        \includegraphics[width=\linewidth]{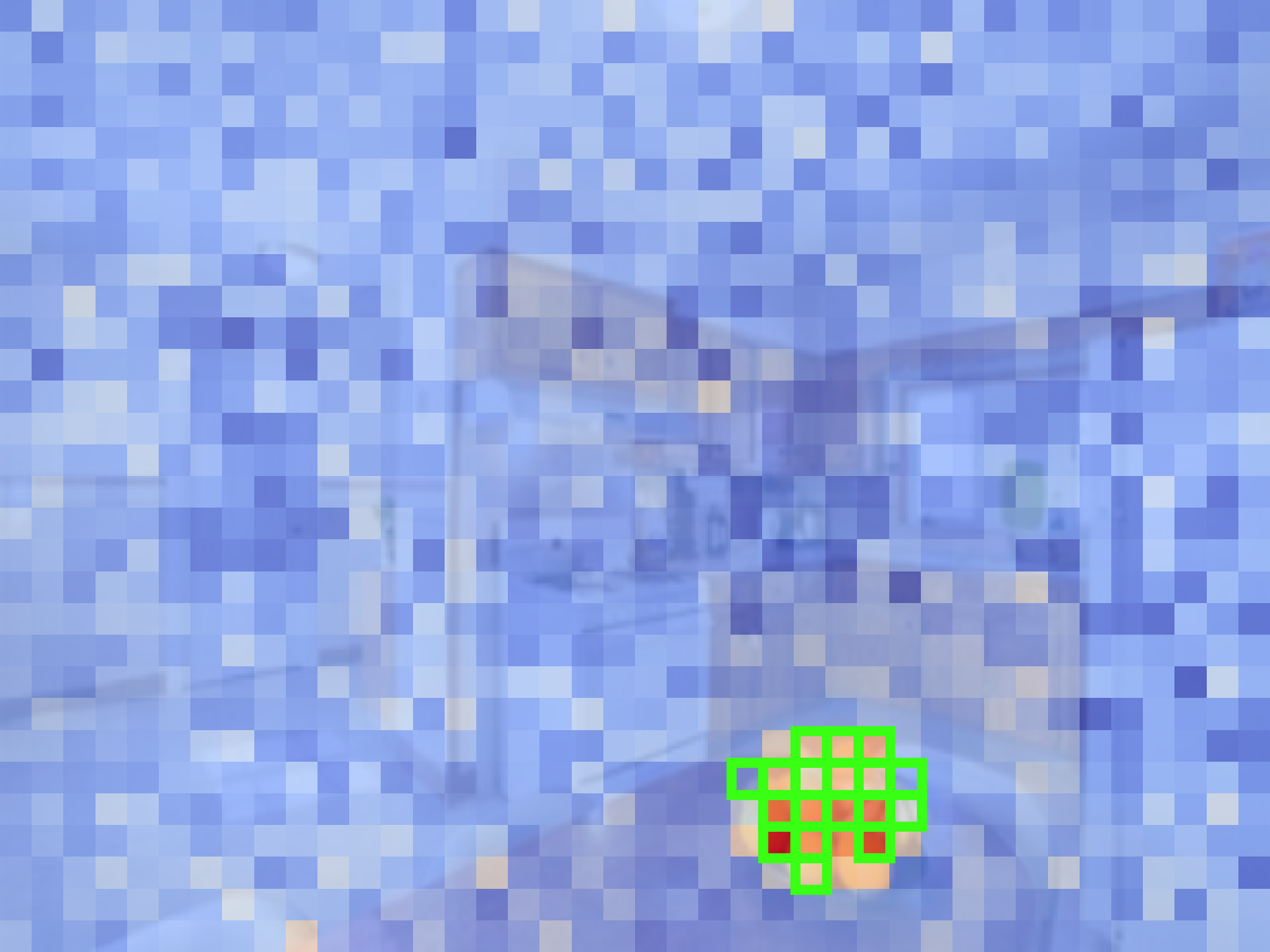}
        \caption{Attr. target: \textit{Food} \supers{}}
    \end{subfigure}% 
    \caption{In this \emph{COCO} example, \supers{} produce attribution masks that precisely localize ground-truth \textit{Food} regions, concentrating high scores on the bananas and oranges. The global concept vector-based attribution is more diffuse, scoring background regions highly as well.}
    \label{fig:lime-example-1}
\end{figure*}

\begin{figure*}[!h]
    \centering
    \setlength{\fboxsep}{0pt}
    \begin{subfigure}[t]{0.31\textwidth}
    \includegraphics[width=\linewidth]{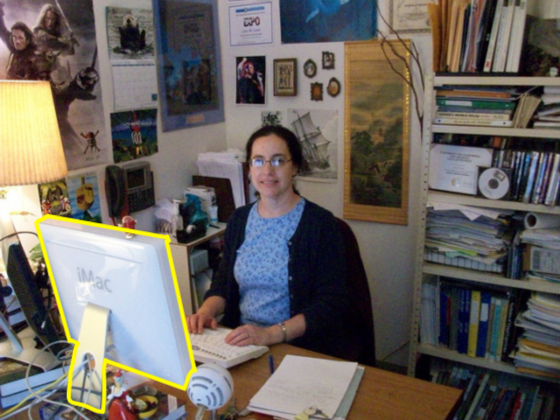}
        \caption{Image with \textit{TV} labeled}
    \end{subfigure}% 
    \hspace{0.01\textwidth}
    \begin{subfigure}[t]{0.31\textwidth}
    \includegraphics[width=\linewidth]{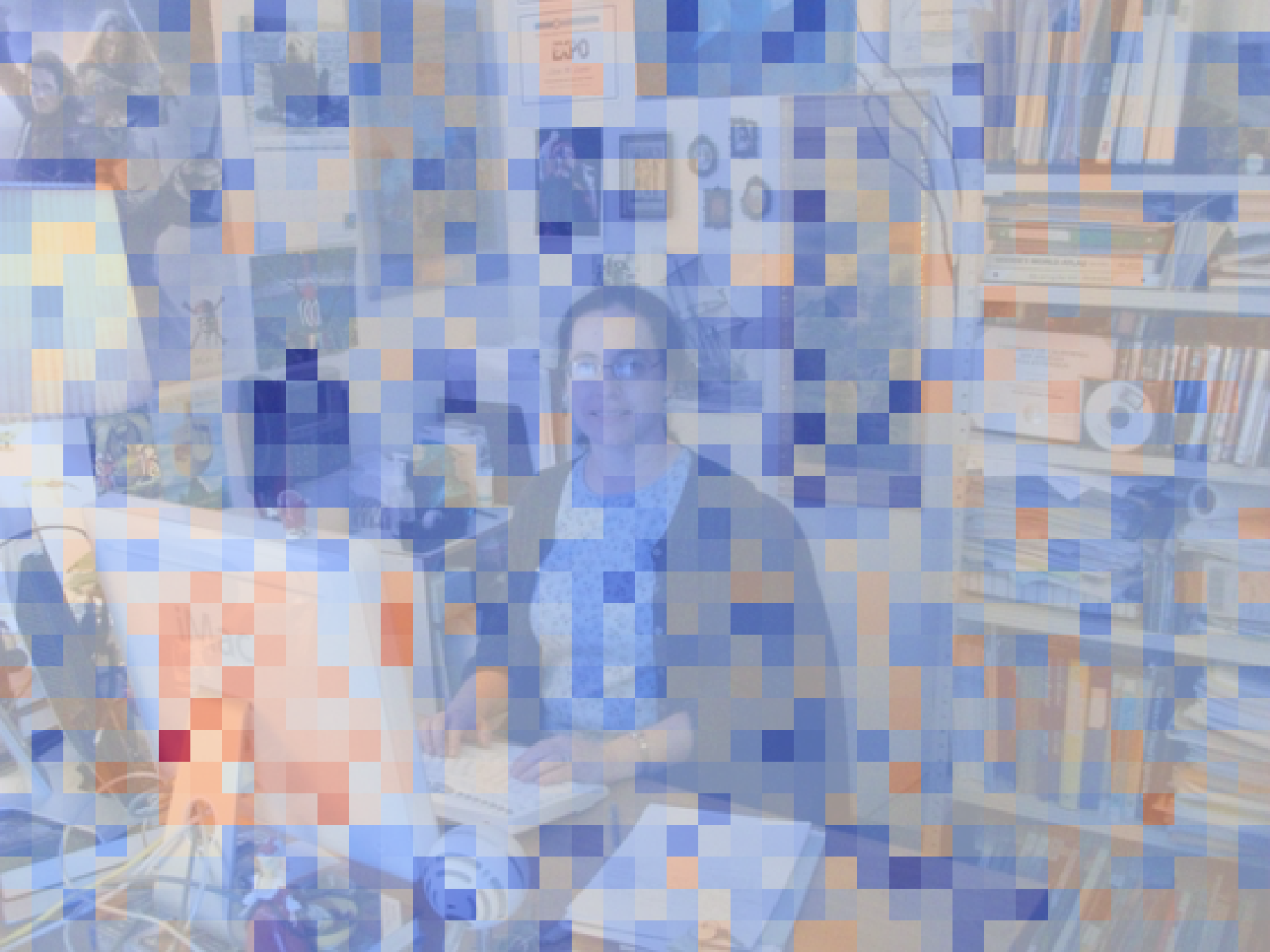}
        \caption{Attr. target: \textit{TV} concept vector}
    \end{subfigure}% 
    \hspace{0.01\textwidth}
    \begin{subfigure}[t]{0.31\textwidth}
        \includegraphics[width=\linewidth]{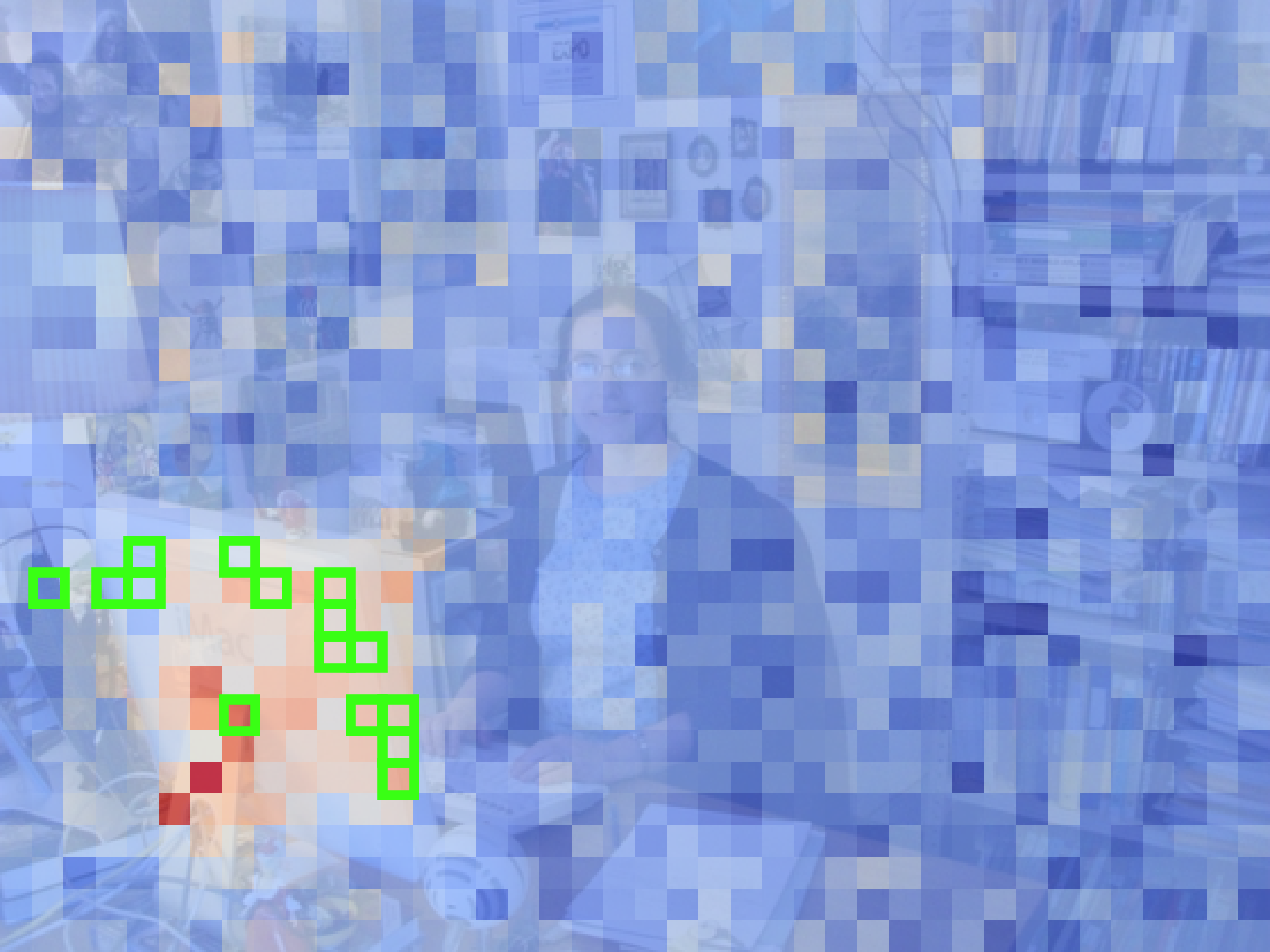}
        \caption{Attr. target: \textit{TV} \supers{}}
    \end{subfigure}% 
    \caption{In this \emph{COCO} example, \supers{} produce attribution masks that precisely localize ground-truth \textit{TV} regions, while the global concept vector-based attribution highlights correlated objects like books.}
    \label{fig:lime-example-2}
\end{figure*}

\begin{figure}[!h]
    \centering
    \setlength{\fboxsep}{0pt}
    \begin{subfigure}[t]{1\textwidth}
    \includegraphics[width=\linewidth]{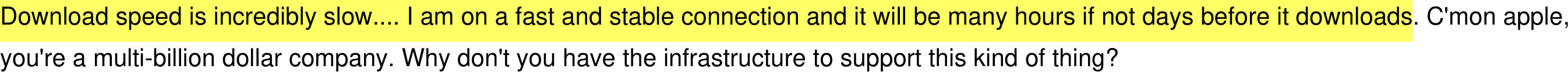}
        \caption{Original text (\textit{Download} highlighted)}
        \label{fig:lime-a}
    \end{subfigure}% 
    \hspace{0.01\textwidth}
    \begin{subfigure}[t]{1\textwidth}
    \includegraphics[width=\linewidth]{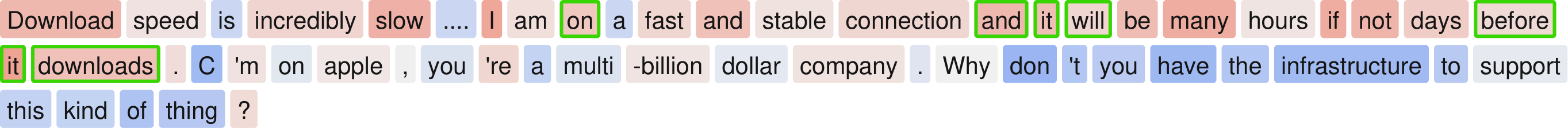}
        \caption{Attr. target: \textit{Download} global concept vector}
    \label{fig:lime-b}
    \end{subfigure}%
    \hspace{0.01\textwidth}
    \begin{subfigure}[t]{1\textwidth}
    \includegraphics[width=\linewidth]{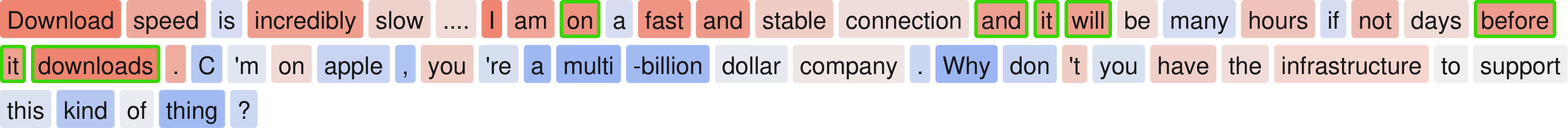}
        \caption{Attr. target: \textit{Download} \supers{}}
    \label{fig:lime-c}
    \end{subfigure}%
    \caption{In this example from the \emph{El Capitan} dataset, the most highly attributed tokens in the \super{}-based scheme align with the labeled download span. The global concept vector-based attribution map is much more diffuse across the text sample.
    }

    \label{fig:sarcasm-example}
\end{figure}
\clearpage
\section{Sparse Autoencoders}
\label{app:saes}
\subsection{SAEs for Concept Detection}
\label{app:sae-results}
Sparse autoencoders~\citep{bricken2023monosemanticity} (SAEs) are a mechanism for uncovering latent concepts in large models. By training an encoder–decoder architecture with sparsity constraints, SAEs aim to discover a set of basis features that are both interpretable and disentangled. This approach is attractive for concept analysis because sparsity encourages individual hidden units to capture relatively specific and semantically meaningful directions in representation space. In principle, such units could act as natural “concept detectors” without additional supervision.

Despite these benefits, SAEs come with notable limitations. Training them at scale is extremely resource-intensive, and thus only a small number of pretrained SAEs have been made publicly available. These models are typically trained on very specific layers of particular architectures and cannot be easily transferred to other checkpoints or layers. For this reason, we restrict our comparisons to what is currently feasible: an SAE trained on a late residual stream of CLIP~\citep{Radford2021LearningTV, ewingtonpitsos2024clipscope_pypi} (covering ~92\% of the model depth for images) and SAEs trained on intermediate layers of Gemma~\citep{team2024gemma, Lieberum2024GemmaSO} (covering ~81\% of the depth for text). A second practical issue is that SAEs output thousands of candidate units, which makes automatic labeling more difficult. To address this, we filtered out units that activated on nearly all samples or no samples~\citep{Cywinski2025SAeUronIC}, or with insufficient activation strength~\citep{Gao2024ScalingAE}.

After filtering, we evaluated the retained SAE units as potential unsupervised concept detectors. We apply the same \super{} paradigm for detection, treating [CLS] and token-alignment with the retained SAE units as concept activation scores. Table~\ref{tab:sae_superact_detection} shows the $F_1$ concept detection performance for the best-performing SAE units for each ground truth concept. 

\begin{figure}[h!]
    \centering
    \includegraphics[width=0.7\textwidth]{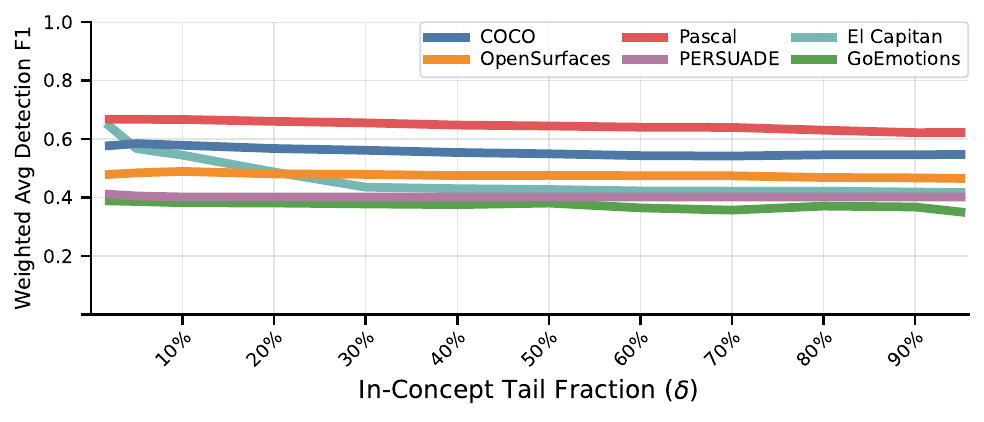}
    \caption{For SAEs, detection performance is similar across tail fraction $\delta$s.}
    \label{fig:detection-sparsity-sae}
\end{figure}

With SAE concepts, \supers{} remain competitive with the strongest aggregation baseline on every dataset, outperforming baselines on the text datasets and matching MeanTok within error on the image datasets (Table~\ref{tab:sae_superact_detection}). However, Figure~\ref{fig:detection-sparsity-sae} shows that performance is similar across tail fraction $\delta$s. One plausible explanation is the sparsity penalty in the SAE training
objective: by suppressing weak responses and shrinking the scale of the surviving ones, it may reduce the contrast between the strongest and moderate activations, spreading concept evidence across more tokens rather than concentrating it in the tail.

\begingroup
\centering
\small
\setlength{\tabcolsep}{8pt}
\captionof{table}{\textbf{Concept detection performance using SAE activations across image and text datasets.} Results compare RandTok, LastTok, MeanTok, CLSTok, and SuperAct. \textbf{Bold} indicates the highest mean; \underline{underline} indicates the second-highest distinct mean.}
\label{tab:sae_superact_detection}
\begin{tabular}{lccccc}
    \toprule
    & \multicolumn{5}{c}{Concept Detection Methods} \\
    \cmidrule(lr){2-6}
    & RandTok & LastTok {\scriptsize\cite{Chen2025PersonaVM}} & MeanTok {\scriptsize\citep{McKenzie2025DetectingHI}} & CLSTok {\scriptsize\citep{Yu2024LatentCE}} & SuperAct \\
    \midrule
    COCO          & 0.313 {\scriptsize$\pm 0.010$} & 0.309 {\scriptsize$\pm 0.011$} & \textbf{0.579} {\scriptsize$\pm 0.014$} & 0.420 {\scriptsize$\pm 0.014$} & \underline{0.578} {\scriptsize$\pm 0.014$} \\
    OpenSurfaces  & 0.333 {\scriptsize$\pm 0.010$} & 0.321 {\scriptsize$\pm 0.010$} & \textbf{0.485} {\scriptsize$\pm 0.016$} & 0.398 {\scriptsize$\pm 0.014$} & \underline{0.479} {\scriptsize$\pm 0.016$} \\
    Pascal        & 0.389 {\scriptsize$\pm 0.006$} & 0.408 {\scriptsize$\pm 0.006$} & \textbf{0.670} {\scriptsize$\pm 0.007$} & 0.567 {\scriptsize$\pm 0.007$} & \underline{0.668} {\scriptsize$\pm 0.007$} \\
    \midrule
    PERSUADE      & 0.402 {\scriptsize$\pm 0.019$} & 0.402 {\scriptsize$\pm 0.019$} & 0.417 {\scriptsize$\pm 0.021$} & \underline{0.421} {\scriptsize$\pm 0.021$} & \textbf{0.428} {\scriptsize$\pm 0.022$} \\
    El Capitan    & 0.390 {\scriptsize$\pm 0.020$} & 0.427 {\scriptsize$\pm 0.022$} & \textbf{0.631} {\scriptsize$\pm 0.023$} & \underline{0.432} {\scriptsize$\pm 0.023$} & \textbf{0.631} {\scriptsize$\pm 0.024$} \\
    GoEmotions    & 0.070 {\scriptsize$\pm 0.009$} & 0.091 {\scriptsize$\pm 0.015$} & \underline{0.386} {\scriptsize$\pm 0.029$} & 0.119 {\scriptsize$\pm 0.020$} & \textbf{0.389} {\scriptsize$\pm 0.029$} \\
    \bottomrule
\end{tabular}
\par
\vspace{-1em}
\endgroup

\clearpage
\subsection{SAEs for Concept Attribution}
Having established that SAEs can act as competitive unsupervised detectors, we next evaluate whether they can also support concept attribution. Tables~\ref{tab:sae_inversion_image_clip} and \ref{tab:sae_inversion_text_qwen} report average attribution $F_1$ across both image and text datasets.  

Across all methods, we observe a consistent pattern: using the average of local \supers{} derived from SAE concepts produces attribution maps that align better with ground truth labels. On image datasets, \supers{} improve scores in nearly every setting, often by non-trivial margins. Similar trends appear in text, where \supers{} again provide the strongest performance in most cases. While the average $F_1$ across all concepts remains modest relative to supervised baselines, the results highlight a consistent trend: even for SAEs, \supers{} consistently provide a more accurate signal for both concept detection and attribution than global CLS-based pooling.

\begin{table}[ht]
\caption{Average Attribution F1 for SAEs on Image Datasets with CLIP model.}
\label{tab:sae_inversion_image_clip}
\centering
\small
\begin{subtable}{0.9\textwidth}
\centering
\caption{CLEVR and COCO Dataset}
\begin{tabular}{@{}lcccc@{}}
\toprule
Attribution Method
& \multicolumn{2}{c}{CLEVR}
& \multicolumn{2}{c}{COCO} \\
\cmidrule(lr){2-3} \cmidrule(lr){4-5}
& CLS & \supers{} & CLS & \supers{} \\
\midrule
LIME & 0.45 {\scriptsize $\pm$ 0.04} & \textbf{0.49} {\scriptsize $\pm$ 0.01} & 0.32 {\scriptsize $\pm$ 0.03} & \textbf{0.33} {\scriptsize $\pm$ 0.04} \\
SHAP & 0.47 {\scriptsize $\pm$ 0.05} & \textbf{0.51} {\scriptsize $\pm$ 0.03} & 0.31 {\scriptsize $\pm$ 0.03} & \textbf{0.34} {\scriptsize $\pm$ 0.02} \\
RISE & 0.44 {\scriptsize $\pm$ 0.03} & \textbf{0.48} {\scriptsize $\pm$ 0.03} & 0.30 {\scriptsize $\pm$ 0.02} & \textbf{0.33} {\scriptsize $\pm$ 0.01} \\
SHAP IQ & \textbf{0.46} {\scriptsize $\pm$ 0.04} & \textbf{0.46} {\scriptsize $\pm$ 0.02} & 0.28 {\scriptsize $\pm$ 0.05} & \textbf{0.33} {\scriptsize $\pm$ 0.04} \\
IntGrad & 0.40 {\scriptsize $\pm$ 0.05} & \textbf{0.44} {\scriptsize $\pm$ 0.04} & 0.27 {\scriptsize $\pm$ 0.04} & \textbf{0.31} {\scriptsize $\pm$ 0.03} \\
GradCAM & 0.36 {\scriptsize $\pm$ 0.05} & \textbf{0.40} {\scriptsize $\pm$ 0.05} & 0.26 {\scriptsize $\pm$ 0.05} & \textbf{0.30} {\scriptsize $\pm$ 0.04} \\
FullGrad & 0.37 {\scriptsize $\pm$ 0.04} & \textbf{0.41} {\scriptsize $\pm$ 0.02} & \textbf{0.32} {\scriptsize $\pm$ 0.03} & 0.31 {\scriptsize $\pm$ 0.04} \\
CALM & 0.44 {\scriptsize $\pm$ 0.02} & \textbf{0.47} {\scriptsize $\pm$ 0.04} & 0.27 {\scriptsize $\pm$ 0.05} & \textbf{0.32} {\scriptsize $\pm$ 0.03} \\
MFABA & 0.44 {\scriptsize $\pm$ 0.03} & \textbf{0.45} {\scriptsize $\pm$ 0.02} & 0.28 {\scriptsize $\pm$ 0.04} & \textbf{0.30} {\scriptsize $\pm$ 0.03} \\
\bottomrule
\end{tabular}
\end{subtable}
\vspace{0.5em} % small vertical space between subtables
\begin{subtable}{0.9\textwidth}
\centering
\caption{OpenSurfaces and Pascal Dataset}
\begin{tabular}{@{}lcccc@{}}
\toprule
Attribution Method
& \multicolumn{2}{c}{OpenSurfaces}
& \multicolumn{2}{c}{Pascal} \\
\cmidrule(lr){2-3} \cmidrule(lr){4-5}
& CLS & \supers{} & CLS & \supers{} \\
\midrule
LIME & 0.41 {\scriptsize $\pm$ 0.04} & \textbf{0.43} {\scriptsize $\pm$ 0.04} & 0.40 {\scriptsize $\pm$ 0.05} & \textbf{0.45} {\scriptsize $\pm$ 0.04} \\
SHAP & 0.31 {\scriptsize $\pm$ 0.03} & \textbf{0.37} {\scriptsize $\pm$ 0.02} & 0.41 {\scriptsize $\pm$ 0.04} & \textbf{0.45} {\scriptsize $\pm$ 0.03} \\
RISE & 0.36 {\scriptsize $\pm$ 0.05} & \textbf{0.41} {\scriptsize $\pm$ 0.02} & 0.40 {\scriptsize $\pm$ 0.05} & \textbf{0.43} {\scriptsize $\pm$ 0.05} \\
SHAP IQ & 0.37 {\scriptsize $\pm$ 0.04} & \textbf{0.40} {\scriptsize $\pm$ 0.05} & 0.42 {\scriptsize $\pm$ 0.05} & \textbf{0.45} {\scriptsize $\pm$ 0.01} \\
IntGrad & 0.39 {\scriptsize $\pm$ 0.02} & \textbf{0.43} {\scriptsize $\pm$ 0.02} & 0.46 {\scriptsize $\pm$ 0.05} & \textbf{0.50} {\scriptsize $\pm$ 0.02} \\
GradCAM & 0.32 {\scriptsize $\pm$ 0.05} & \textbf{0.36} {\scriptsize $\pm$ 0.02} & 0.34 {\scriptsize $\pm$ 0.03} & \textbf{0.39} {\scriptsize $\pm$ 0.04} \\
FullGrad & 0.34 {\scriptsize $\pm$ 0.03} & \textbf{0.37} {\scriptsize $\pm$ 0.03} & 0.36 {\scriptsize $\pm$ 0.05} & \textbf{0.40} {\scriptsize $\pm$ 0.02} \\
CALM & 0.26 {\scriptsize $\pm$ 0.05} & \textbf{0.31} {\scriptsize $\pm$ 0.02} & 0.35 {\scriptsize $\pm$ 0.04} & \textbf{0.39} {\scriptsize $\pm$ 0.03} \\
MFABA & \textbf{0.39} {\scriptsize $\pm$ 0.04} & \textbf{0.39} {\scriptsize $\pm$ 0.02} & 0.41 {\scriptsize $\pm$ 0.03} & \textbf{0.44} {\scriptsize $\pm$ 0.02} \\
\bottomrule
\end{tabular}
\end{subtable}
\end{table}

\begin{table}[ht]
\caption{Average Attribution F1 for SAEs on Text Datasets with Gemma Model.}
\label{tab:sae_inversion_text_qwen}
\centering
\small
\begin{subtable}{0.9\textwidth}
\centering
\caption{PERSUADE and El Capitan Dataset}
\begin{tabular}{@{}lcccc@{}}
\toprule
Attribution Method
& \multicolumn{2}{c}{PERSUADE}
& \multicolumn{2}{c}{El Capitan} \\
\cmidrule(lr){2-3} \cmidrule(lr){4-5}
& CLS & \supers{} & CLS & \supers{} \\
\midrule
LIME      & 0.16 {\scriptsize $\pm$ 0.01} & \textbf{0.18} {\scriptsize $\pm$ 0.01} & 0.32 {\scriptsize $\pm$ 0.02} & \textbf{0.33} {\scriptsize $\pm$ 0.01} \\
SHAP      & \textbf{0.20} {\scriptsize $\pm$ 0.01} & \textbf{0.20} {\scriptsize $\pm$ 0.01} & 0.35 {\scriptsize $\pm$ 0.01} & \textbf{0.37} {\scriptsize $\pm$ 0.02} \\
RISE      & 0.20 {\scriptsize $\pm$ 0.01} & \textbf{0.19} {\scriptsize $\pm$ 0.01} & 0.36 {\scriptsize $\pm$ 0.02} & \textbf{0.39} {\scriptsize $\pm$ 0.01} \\
SHAP IQ   & 0.21 {\scriptsize $\pm$ 0.01} & \textbf{0.22} {\scriptsize $\pm$ 0.01} & 0.35 {\scriptsize $\pm$ 0.01} & \textbf{0.36} {\scriptsize $\pm$ 0.02} \\
IntGrad   & 0.17 {\scriptsize $\pm$ 0.01} & \textbf{0.19} {\scriptsize $\pm$ 0.01} & 0.32 {\scriptsize $\pm$ 0.02} & \textbf{0.35} {\scriptsize $\pm$ 0.01} \\
GradCAM   & 0.19 {\scriptsize $\pm$ 0.02} & \textbf{0.20} {\scriptsize $\pm$ 0.01} & 0.36 {\scriptsize $\pm$ 0.02} & \textbf{0.40} {\scriptsize $\pm$ 0.02} \\
FullGrad  & \textbf{0.17} {\scriptsize $\pm$ 0.01} & \textbf{0.17} {\scriptsize $\pm$ 0.01} & 0.34 {\scriptsize $\pm$ 0.01} & \textbf{0.37} {\scriptsize $\pm$ 0.02} \\
CALM      & 0.18 {\scriptsize $\pm$ 0.02} & \textbf{0.19} {\scriptsize $\pm$ 0.01} & \textbf{0.37} {\scriptsize $\pm$ 0.02} & \textbf{0.37} {\scriptsize $\pm$ 0.01} \\
MFABA     & \textbf{0.19} {\scriptsize $\pm$ 0.01} & \textbf{0.19} {\scriptsize $\pm$ 0.01} & 0.36 {\scriptsize $\pm$ 0.01} & \textbf{0.37} {\scriptsize $\pm$ 0.02} \\
\bottomrule
\end{tabular}
\end{subtable}
\vspace{0.5em}
\begin{subtable}{0.9\textwidth}
\centering
\caption{GoEmotions and iSarcasm Dataset}
\begin{tabular}{@{}lcccc@{}}
\toprule
Attribution Method
& \multicolumn{2}{c}{GoEmotions}
& \multicolumn{2}{c}{iSarcasm} \\
\cmidrule(lr){2-3} \cmidrule(lr){4-5}
& CLS & \supers{} & CLS & \supers{} \\
\midrule
LIME      & 0.17 {\scriptsize $\pm$ 0.01} & \textbf{0.18} {\scriptsize $\pm$ 0.02} & 0.41 {\scriptsize $\pm$ 0.03} & \textbf{0.45} {\scriptsize $\pm$ 0.02} \\
SHAP      & 0.19 {\scriptsize $\pm$ 0.01} & \textbf{0.20} {\scriptsize $\pm$ 0.01} & 0.46 {\scriptsize $\pm$ 0.02} & \textbf{0.47} {\scriptsize $\pm$ 0.03} \\
RISE      & 0.15 {\scriptsize $\pm$ 0.01} & \textbf{0.16} {\scriptsize $\pm$ 0.01} & 0.47 {\scriptsize $\pm$ 0.03} & \textbf{0.50} {\scriptsize $\pm$ 0.03} \\
SHAP IQ   & 0.19 {\scriptsize $\pm$ 0.01} & \textbf{0.20} {\scriptsize $\pm$ 0.02} & 0.49 {\scriptsize $\pm$ 0.02} & \textbf{0.50} {\scriptsize $\pm$ 0.02} \\
IntGrad   & 0.16 {\scriptsize $\pm$ 0.01} & \textbf{0.17} {\scriptsize $\pm$ 0.01} & 0.45 {\scriptsize $\pm$ 0.03} & \textbf{0.48} {\scriptsize $\pm$ 0.02} \\
GradCAM   & \textbf{0.16} {\scriptsize $\pm$ 0.02} & \textbf{0.16} {\scriptsize $\pm$ 0.01} & 0.47 {\scriptsize $\pm$ 0.02} & \textbf{0.49} {\scriptsize $\pm$ 0.02} \\
FullGrad  & 0.15 {\scriptsize $\pm$ 0.01} & \textbf{0.18} {\scriptsize $\pm$ 0.02} & 0.44 {\scriptsize $\pm$ 0.03} & \textbf{0.45} {\scriptsize $\pm$ 0.02} \\
CALM      & 0.19 {\scriptsize $\pm$ 0.02} & \textbf{0.20} {\scriptsize $\pm$ 0.02} & \textbf{0.45} {\scriptsize $\pm$ 0.02} & \textbf{0.45} {\scriptsize $\pm$ 0.03} \\
MFABA     & 0.17 {\scriptsize $\pm$ 0.01} & \textbf{0.16} {\scriptsize $\pm$ 0.01} & 0.46 {\scriptsize $\pm$ 0.03} & \textbf{0.49} {\scriptsize $\pm$ 0.02} \\
\bottomrule
\end{tabular}
\end{subtable}
\end{table}
\clearpage

\end{document}